\documentclass[fleqn, twocolumn,10pt]{wlscirep}
\usepackage[utf8]{inputenc}

\usepackage{cite}
\makeatletter
\let\NAT@parse\undefined
\makeatother
\usepackage{hyperref}
\usepackage{geometry}
\usepackage{graphicx}
\usepackage{algpseudocode}
\usepackage{epsfig,graphics,subfigure}
\usepackage{color}
\usepackage{amsmath}
\usepackage{amssymb}
\usepackage{algorithm}
\usepackage{jabbrv}
\usepackage{bbding}
\usepackage{amsbsy}
\usepackage{multirow}
\usepackage{tabularx}
\usepackage{mathtools}
\usepackage{array, booktabs}
\usepackage{graphicx}
\usepackage{framed}
\usepackage{float} 
\usepackage{caption}
\usepackage{tcolorbox}
\usepackage{capt-of}

\makeatletter
\renewcommand{\@biblabel}[1]{#1.}
\makeatother

\usepackage{marginnote}
\usepackage{hyperref}
\usepackage{url}

\usepackage[utf8]{inputenc}
\usepackage{todonotes}

\usepackage{longtable}
\usepackage{graphicx}

\title{Piecewise Linear Neural Networks and Deep Learning\#}

\author[1,*]{Qinghua Tao}
\author[2,*]{Li Li}
\author[3,*]{Xiaolin Huang}
\author[4]{Xiangming Xi}
\author[2]{Shuning Wang}
\author[1]{\\ Johan A.K. Suykens}
\affil[1]{STADIUS, ESAT, KU Leuven, Heverlee, 3001, Belgium}
\affil[2]{Department of Automation, Tsinghua University, Beijing, 100084, China}
\affil[3]{Department of Automation,Shanghai Jiao Tong Univeristy, Shanghai, 200240,  China}
\affil[4]{Zhejiang Lab, Hangzhou, 311121, China}

\affil[*]{corresponding.author: qinghua.tao@esat.kuleuven.be, li-li@mail.tsinghua.edu.cn, xiaolinhuang@sjtu.edu.cn
\vspace{0.2cm}

\# This work has been published in \emph{Nature Reviews Methods Primers}. For citations, please cite ``Tao, Q., Li, L., Huang, X. et al. Piecewise linear neural networks and deep learning. Nature Reviews Methods Primers 2, 42 (2022). \url{https://doi.org/10.1038/s43586-022-00125-7}''. 

For the published version, please access \url{https://rdcu.be/cPIGw} for online-reading,
 download the manuscript from \url{https://www.nature.com/articles/s43586-022-00125-7\#citeas}, 
and check the supplementary information via \url{https://www.nature.com/articles/s43586-022-00125-7\#Sec38}.}


\begin{abstract}
As a powerful modelling method, PieceWise Linear Neural Networks (PWLNNs) have proven successful in various fields, most recently in deep learning.  To apply PWLNN methods, both the representation and the learning have long been studied. In 1977, the canonical representation pioneered the works of shallow PWLNNs learned by incremental designs, but the applications to large-scale data were prohibited. In 2010, the Rectified Linear Unit (ReLU) advocated the prevalence of PWLNNs in deep learning. Ever since, {PWLNNs have been successfully applied to extensive tasks and achieved  advantageous performances.} In this Primer, we systematically introduce the methodology of PWLNNs    by grouping the  works into shallow and deep networks. Firstly, different PWLNN representation models are constructed with elaborated examples. With PWLNNs, the evolution of  learning algorithms for data  is presented  and fundamental theoretical analysis  follows up for  in-depth understandings. Then, representative applications are introduced together with discussions and outlooks.
\end{abstract}

\begin{document}

\flushbottom
\maketitle

\thispagestyle{empty}

\section{ Introduction}\label{sec:introduction}
\renewcommand{\thefootnote}{\fnsymbol{footnote}} 
PieceWise Linear (PWL) Neural Networks (NN) have been  studied particularly since the 1970s, and are now a successful mainstream method in deep learning. PWLNNs  partition the domain into numerous sub-regions, each of which has localized linearity, whilst maintaining nonlinearity throughout the whole domain with great modelling flexibility \cite{leenaerts2013piecewise}. In PWLNNs, the PWL nonlinearity is realized through network architectures, which do not require  explicit descriptions on  sub-regions of localized linearity.
PWLNNs are explicit in geometrical interpretation and  flexibile in approximation,  such as an induced  conclusion by the {Stone-Weierstrass  approximation theorem\cite{1999Real}[G]\footnotetext[4]{The glossary items are marked with [G] and each of them is given with a succinct explanation in Section \ref{sec:g}.}}. Through PWLNN methods, complex nonlinear systems can be modelled by a finite number of linear functions localized over different sub-regions. This characteristic naturally bridges linearity and nonlinearity, and is more amenable to the modelling, learning, and analysis than other nonlinear  methods.

\renewcommand{\thefootnote}{\arabic{footnote}} 

PWLNNs are neural network-structured versions of  {PWL functions [G]}. 
The conventional representation for  PWL functions
is given by a  region-by-region manner, which explicitly lists each  sub-region tessellating  the  domain and its localized  linear function \cite{chien1977solving}.  The conventional  representation  demands an excessive number of parameters and  is intractable  in  practical applications involving  numerous sub-regions and complex domain configurations \cite{chua1988}. In contrast, PWLNNs efficiently represent PWL functions with  compact expressions organized as network architectures, through which versatile functionalities can be facilitated. PWLNNs have now become a mainstream method for utilizing PWL functions in data science.
To realize the power of PWLNNs in practice, two challenges need to be resolved: how to construct proper models to represent  PWLNNs (representation) and how to effectively optimize the PWLNNs to well model the data or systems (learning). To this end, great efforts have been made by many researchers from different fields. These works can be categorized into two main groups: shallow  (e.g., {Canonical Piecewise Linear Representation (CPLR)\cite{chua1977section} [G]}) and deep  (e.g., networks with {Rectified Linear Units (ReLU)\cite{Nair2010Rectified} [G]}) PWLNNs. During the evolution from shallow to deep, the representation, learning and analysis of PWLNNs are closely related.

CPLR pioneered the compact expressions and analytical studies for  PWL functions   \cite{Kang1978Chua,1993Canonical}. {Hinging hyperplanes\cite{Breiman1993} [G]}  is another important  representation model, constructed with  geometrical explanations of hinges. These two compact representations result in  PWLNNs with one hidden layer, namely shallow PWLNNs, where  absolute-value operators and  maximization operators are adopted  to induce PWL nonlinearity, as in ReLU  for deep learning. 
Subsequent research has focused on variant representation models \cite{1995Explicitlin,Terela1999Lattice,Julin2003The,2005wen,Wang2005GHH,DBLP:conf/isnn/SunW05,Xu2009AHH,yu2017incremental}, theoretical analysis \cite{Chua1985Canonical,Wang2004General,2010AWangsmooth,2010Stability,2011Dynamic,2012Exact,Xu2016Irredundant,Xu2016Minimal},  learning algorithms \cite{Pucar95smoothhinging,Hush1998Efficient,2002Nonlinear,wangsun2007,2008Configuration,2010Identification}, to name a few.
 Most of the resulting  PWLNNs were  shallow-architectured, and yet their performances were limited in  high-dimensional and large-scale problems.
 In 2010,  ReLU  successfully embedded  PWL nonlinearity in the mainstream Deep neural networks (PWL-DNNs)  and has achieved  record-breaking performances  in various benchmarks \cite{krizhevsky2012imagenet,DBLP:conf/cvpr/HeZRS16,DBLP:conf/cvpr/HuangLMW17}. For example, given the same number of neurons, PWL-DNNs have exponentially greater capacity than that of their shallow counterparts \cite{DBLP:conf/iclr/AroraBMM18}.

For learning algorithms, shallow PWLNNs commonly use incremental designs, which  iteratively grow wider networks. PWL-DNNs inherit the regular learning of generic DNNs, in which network structures are predefined and parameters are optimized by the {backpropagation  strategy [G]} and  {stochastic gradient descent (SGD) [G]} algorithm. With the support of powerful graphical and tensor processing units and efficient implementation platforms such as PyTorch \cite{paszke2019pytorch}, the powerful learning ability of PWL-DNNs is fundamentally realized to tackle complex tasks, and PWL-DNNs have since become the mainstream deep learning method after  decades of rigorous developments.

In this primer, the PWLNN method is systematically introduced and versatile perspectives  are provided,  to give a more insightful understanding  towards representations, learning, and analysis. 
Some preliminaries are given, and details of how to construct  representation models for PWLNNs (Experimentation) followed by learning algorithms for data analysis and their theoretical properties (Results). We introduce  several representative applications of PWLNNs (Applications) and the use of PWLNNs for data under standard cases (Reproducibility and data deposition). To conclude, the ongoing issues (Limitations and optimizations) and some potential future directions (Outlook) are discussed. 

\section{ Experimentation}\label{sec:representation}
In this section, some preliminaries are firstly given, and then the representations of PWLNNs are introduced in detail. 

\subsection{ Preliminaries}\label{sec:background}
\subsubsection{  PWL functions}\label{sec:background:pwlfunction}
Linear functions are basic mathematical models, but lack flexibility in  practical scenarios which commonly pertain nonlinear natures. PWL  functions are powerful remedies bridging linearity to nonlinearity for great model flexibility. PWL functions are not necessarily continuous; in practice, continuity  pervasively and sometimes  naturally exists. For the purpose of this Primer, we will discuss continuous PWL functions.

In the conventional representation, PWL functions are given by a region-by-region manner \cite{chien1977solving}. Let $f(\pmb x): \Omega \mapsto \mathbb R$ be a function defined in domain $\Omega \subseteq \mathbb R^n $.  $f(\pmb x)$ is a PWL function, if it satisfies equations \eqref{eq:convention:1} and \eqref{eq:convention:2} (or \eqref{eq:convention:3}). 

To be specific, the domain $\Omega$  is divided into a finite number of polyhedral sub-regions $\Omega_i$ with interior $\mathring{\Omega}_{i}$ satisfying
\begin{equation}\label{eq:convention:1}
\bigcup \Omega_i = \Omega, \ \mathring{\Omega}_{i_1} \bigcap \mathring{\Omega}_{i_2} = \emptyset, \ \forall {i_1} \neq {i_2}, \ {i_1}, {i_2}=1, \ldots, d,
\end{equation}
by a finite set of boundaries $\mathcal B = \{\pi_j(\pmb x)\}_{j=1}^h$, such that each boundary is  an $(n-1)$-dimensional hyperplane characterized by $\pi_j(\pmb x)\coloneqq \pmb {\alpha}_j^T\pmb x - \beta_j=0$ with $\pmb {\alpha}_j \in \mathbb R^n$, $\beta_j\in \mathbb R$, and cannot be covered by any $(n-2)$-dimensional hyperplane.
There exists a finite number of linear functions $l_1(\pmb x), \ldots, l_d(\pmb x)$ which constitute the complete expression of  $f(\pmb x)$, such that 
\begin{equation}\label{eq:convention:2}
f(\pmb x)\in \{l_1(\pmb x), \ldots, l_d(\pmb x)\},  \ \forall \pmb x \in \Omega,
\end{equation}
or equivalently 
\begin{equation}\label{eq:convention:3}
f(\pmb x) = l_i(\pmb x)=\pmb J^T_i\pmb x + b_i, \ \forall \pmb x\in \Omega_i, \ i=1, \ldots, d,
\end{equation}
where  $\pmb J_i \in \mathbb R^n$ is called the Jacobian vector of  the sub-region $\Omega_i$ with the bias $b_i\in \mathbb R$.

For example, denoting  $f: \mathbb R^3 \mapsto \mathbb R$ as a  PWL function, by the  conventional representation, $f(\pmb x)$  is expressed as 
\begin{equation}\label{eqn:ex:1}
f(\pmb x)=\left\{
\begin{array}{ll}
l_1(\pmb x) = x_1 - x_2 + x_3 +1, &  \pi(\pmb x)\geq 0, \\
l_2(\pmb x) = -x_1 + x_2 -x_3 -1, &   \pi(\pmb x) < 0,\\
\end{array} \right.
\end{equation}
where the biases are $b_1=1$ and $b_2=-1$, the boundary set $\mathcal H $ contains  $\pi(\pmb x)\coloneqq  x_1 - x_2 + x_3 + 1=0$, the sub-regions are $\Omega_1=\{\pmb x \in \mathbb R^3| x_1 - x_2 + x_3 + 1 \geq 0 \}$ and $\Omega_2 =\{\pmb x \in \mathbb R^3| x_1 - x_2 + x_3 + 1 < 0 \}$, and the  Jacobian vectors  of $\Omega_1$ and $\Omega_2$ are  $\pmb J_1 = [1, -1, 1]^T$ and  $\pmb J_2 = [-1, 1, -1]^T$, respectively.

Note that with $d = 1$, the PWL function $f(\pmb x)$ is reduced to be linear, thus, we regard linear functions as a special case of PWL functions throughout the Primer. Compared to other nonlinear models,  PWL functions possess explicit geometric interpretation, and  many practical systems can be easily transformed into PWL nonlinear functions \cite{Julian1999phd}, such as {PWL memristors [G]} \cite{Ohnishi1994A,itoh2008memristor},  specialized cost functions \cite{Bradley1997Clustering,Kim2000A,2010Acost,Liu2016Sparse,2017Aliu}, and part mathematical programmings \cite{Zhang2006Separable,Zhang2008Nonseparable,Guisewite1991Minimum,2001Linearconace,DBLP:journals/npl/XiHSW16,xu2015tunneling}. As powerful nonlinear models, PWL functions are proven universal approximators\cite{Goodfellow2013MaxoutN}: let $\Omega\subset \mathbb R^n$ be a compact domain,  and  $p(\pmb x):\Omega \mapsto \mathbb R$ be a continuous function. When $\forall \epsilon>0$,  there exists a PWL function $f(\pmb x)$ (depending upon $\epsilon$), such that $\forall \pmb x\in \Omega$, $|f(\pmb x) - p(\pmb x)| < \epsilon$.

 \subsubsection{  PWLNNs}\label{sec:background:pwlnnn}
Before applying the PWL nonlinearity, it is important to construct proper mathematical formulas to represent such PWL functions. Neural networks have been widely recognized as one of the most powerful nonlinear approximators in data science.  Neural networks refer to mathematical layered models composed of artificial neurons  and their connections \cite{hopfield1982neural}. Mapping through neurons introduces nonlinearity  by activation functions, leading to powerful flexible models.  In neural networks, such mappings are realized via the mapping functions of neurons. In PWLNNs, such nonlinear mapping functions are specified as PWL functions, bringing PWL nonlinearity to the network. Box 1 defines PWLNNs from the perspective of network architectures. This definition has been generalised to include the interconnection weight of an edge $(v^{\prime}, v)$ between neuron $v^{\prime}$ and neuron $v$, 
\begin{tcolorbox}[title=Box 1$|$ \textbf{PieceWise Linear Neural Networks (PWLNNs)},sidebyside align=top,lower separated=false, colback = white, float=ht!]
    \begin{center}
    \includegraphics[width=1\linewidth]{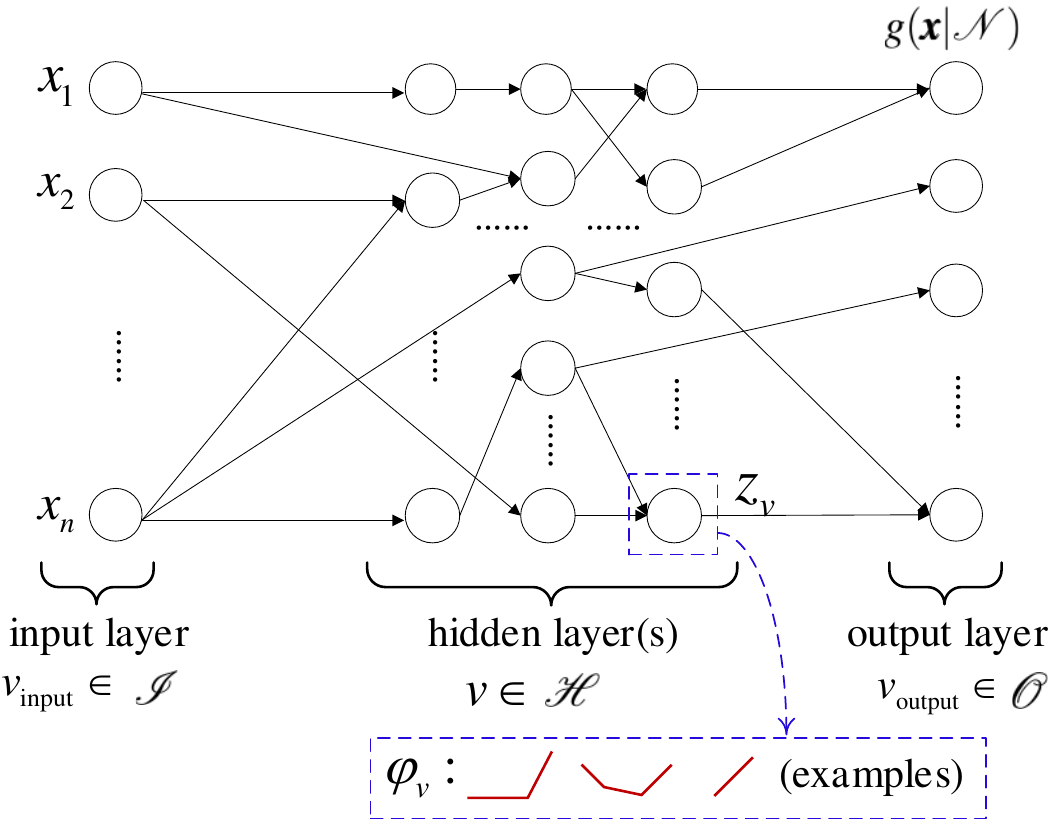}
    \captionof*{figure}{An illustration of PWLNNs  defined below.}
    \end{center}
Let $\pmb g(\pmb x| \mathcal{N})$ be a PWLNN with the architecture $\mathcal{N}=(\mathcal{V}, \mathcal{E})$, determined by a finite set of neurons, $\mathcal{V}$, and a finite set of edges, $\mathcal{E}$, depicting the directed connections between paired neurons in $\mathcal{V}$.
Given an edge $e = (v^{\prime}, v) \in \mathcal{E}$, it is an incoming edge for $v$ and
outgoing edge for $v^{\prime}$. Neurons in $\mathcal{V}$ can be grouped into non-overlapping and non-empty subsets consisting of input neurons ($\mathcal{I}$), output neurons ($\mathcal{O}$), and hidden neurons ($\mathcal{H}$), such that $\mathcal{V} = \mathcal{I} \cup \mathcal{O} \cup \mathcal{H}$. 

The architecture of such a network can be structured into layers. Neurons in  $\mathcal{I}$ constitute the input layer, or the $0$-th layer.
Neurons in $\mathcal{H}$ can be  organized into hidden layers $L$ ($L \geq 1$) connected sequentially, such that a neuron in the $l$-th $(1\leq l \leq L)$ layer  only has incoming edges from neurons in precedent layers (including the input layer) and has outgoing edges to neurons in subsequent layers (including the output layer).  The output layer is constructed by neurons in $\mathcal{O}$, and is also called the $(L+1)$-th layer.

For any neuron $v \in \mathcal{V} \setminus \mathcal{I}$, there is either a linear or PWL mapping function $\phi_v : \mathbb{R}^{n_v} \to \mathbb{R}$, so that 
$$
z_v = \phi_v(z_{v_1}, \cdots,  z_{v_{n_v}};\pmb \theta_v )
$$
where $z_{v}$ is the neuron output, 
$n_v$ is the number of incoming edges for $v$, which are directed from $v_{i}, i =1, \cdots, n_v$ in precedent layers to $v$, and $\pmb \theta_v$ denotes the learnable parameters related to neuron $v$. With $L=1$, it is  a shallow PWLNN, when $L\geq 2$ it is a PWL-DNN.
\end{tcolorbox}
as generally used in neural networks,  into the PWL mapping function of the neuron $v$. This makes PWLNNs more expressible for complex mapping functions in neurons, such as those used in the Maxout neural networks \cite{Goodfellow2013MaxoutN}. Furthermore, all the parameters $\pmb \Theta = \{\pmb \theta_v\}_{v \in \mathcal{V}\setminus\mathcal{I}}$ constitute the parameter space of the network, and   correspond to the commonly-used interconnection weights and bias terms in  other descriptions for neural networks.

Two main aspects should be considered before applying PWLNNs in practice. One is the determination of the network structures of $\mathcal{N}$, and the other is the determination of the learnable parameters.

PWLNNs are the network-structured versions of PWL functions. However, given a specific representation model of PWLNNs, where $\mathcal{N}$ is specified to a certain class of  network structures, this representation model does not necessarily have universal representation ability for arbitrary PWL functions in $\mathbb R^n$. This leads to the definition of  
universal representation ability as follows: when $\mathcal F$ is the set of all PWL functions as defined by the conventional representation,
a PWLNN representation model  $g(\pmb x| \mathcal{N})$ with a single output neuron in $\mathcal{O}$ is said to have  universal representation ability for $\mathcal F$, if the following condition is satisfied
\begin{equation}\label{eq:represent}
\forall f \in \mathcal F, \ \exists  {\mathcal{N}} \ \text{such that} \ f(\pmb x) = g(\pmb x|  \mathcal{N}).
\end{equation}

\begin{figure}[ht!]
\begin{center}
\includegraphics[height=7.6cm]{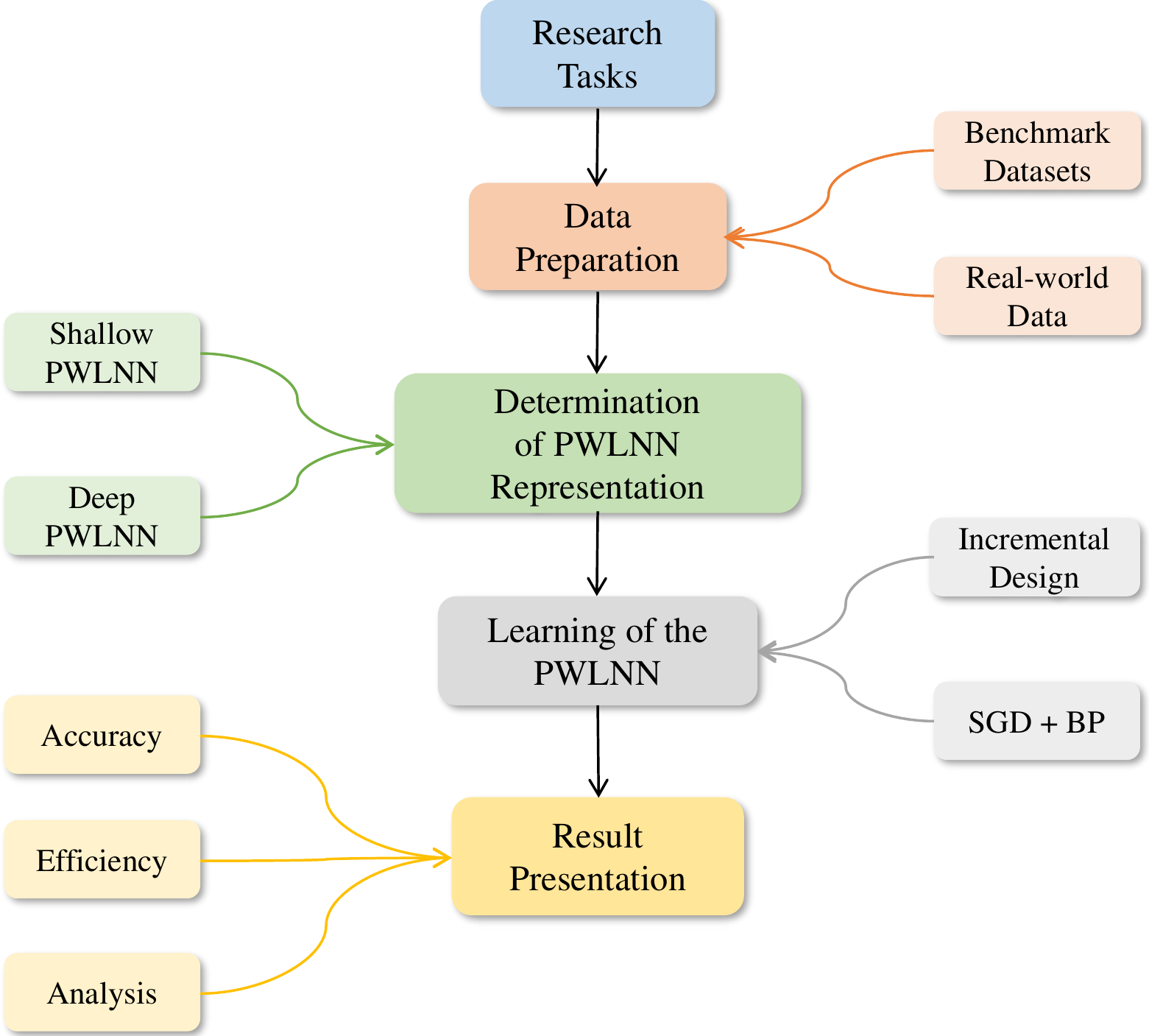}
\caption{A general workflow of applying the PWLNN method. {Given a task of prediction or analysis, standard procedures start from data preparation, where pre-processing or data-augmentation techniques can be involved. Then, an appropriate PWLNN representation  is selected to perform the modelling together with a learning algorithm for fitting the data, where the selection can be determined by practitioners' particular interest or through trail-and-error strategy. With the learned PWLNN, prediction outputs of the given task can be computed, where empirical and theoretical investigations can both be conducted to present the final results.}}\label{fig:workflow}
\end{center}
\end{figure}

Besides the determination of the network  architecture of a PWLNN,  how to effectively identify  model parameters  is also critical to attain good performance in practice.  Figure \ref{fig:workflow} gives the workflow when applying PWLNNs to specific tasks.

\subsection{ Shallow  PWLNNs}

In shallow  PWLNNs, there are two main types of representations: the models consisting of  basis functions  and  the Lattice representation.  An overall sketch on the chronicle of all surveyed PWLNN representations can be found in the supplementary information.

 \subsubsection{  Representations based on basis functions -- the canonical family}
\begin{figure}[ht!]
\centering
\subfigure[a] { \label{fig:pwl:basis}
\includegraphics[height=0.42\columnwidth]{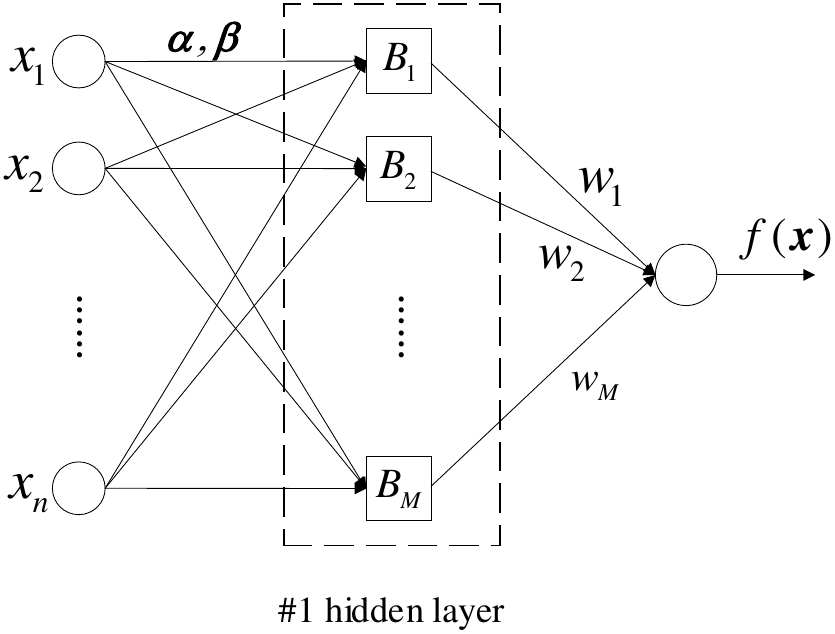}
}
\subfigure[b] { \label{fig:pwl:lattice}
\includegraphics[height=0.42\columnwidth]{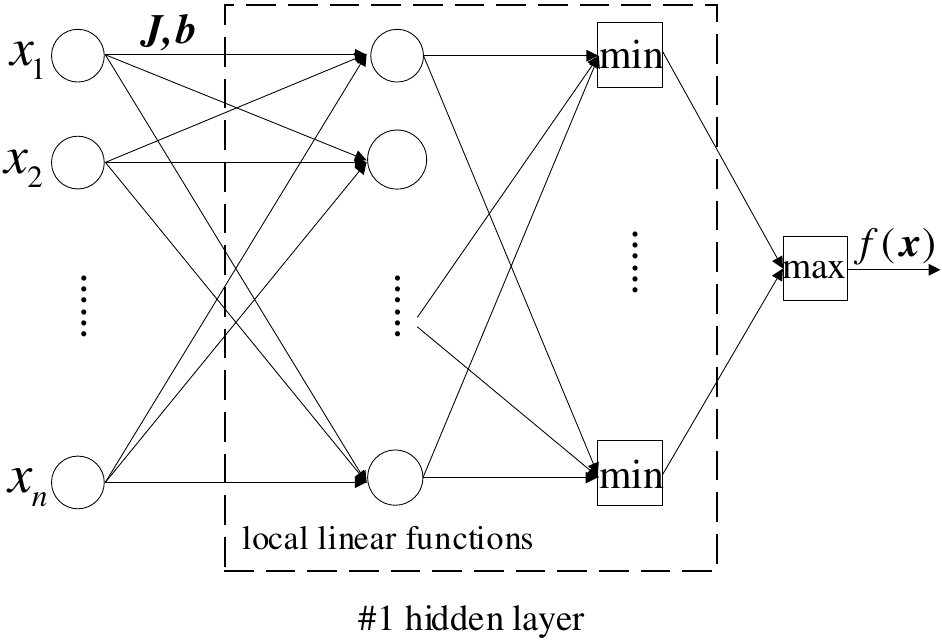}
}
\subfigure[c, d] { \label{fig:pwl:nn}
\includegraphics[height=0.42\columnwidth]{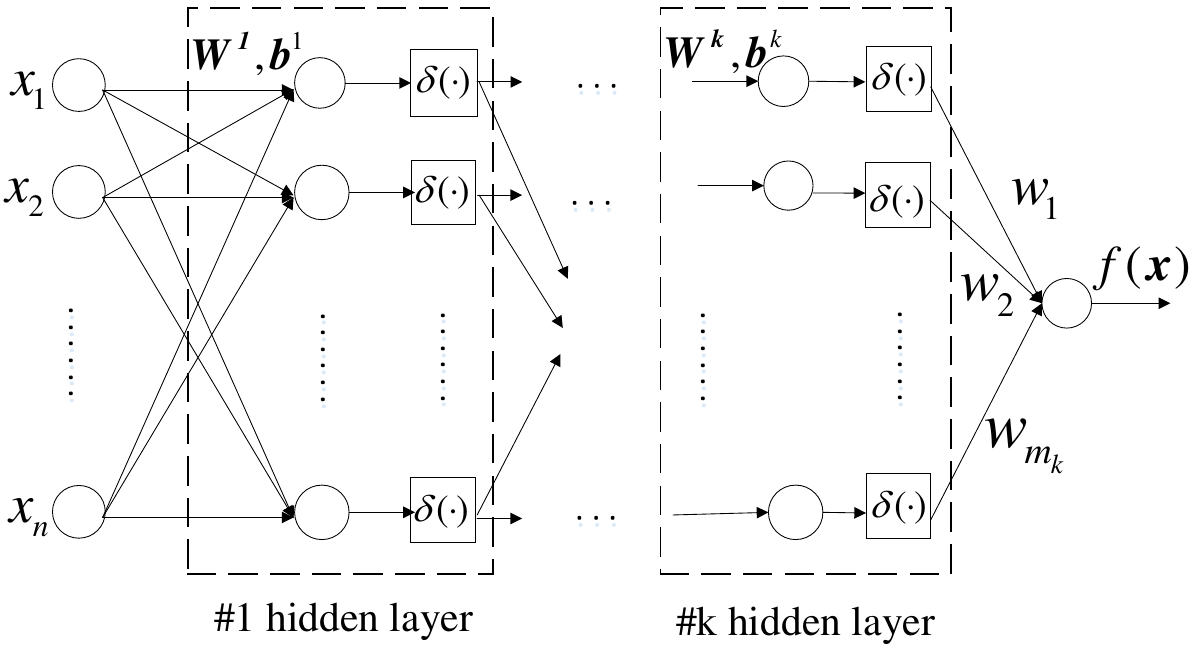}
}
\caption{An illustration on the  topology of  PieceWise Linear Neural Network (PWLNN) representations, where the outputs of  squared nodes denote the PWL  mappings. a) Representation of basis functions in equation \eqref{eq:basis}; b) Lattice representation of equation \eqref{eq:lattice}; c) PWL-DNNs of equation  \eqref{eq:nn:neuron:output} with 1 hidden layer; d) PWL-DNNs of equation  \eqref{eq:nn:neuron:output} with $k$ hidden layers.}
\label{fig:pwl:architecture}
\end{figure}

 The representation models consisting of basis functions are formulated in the form of
  \begin{equation}\label{eq:basis}
f(\pmb x) = \sum_{m=1}^M w_mB_m(\pmb x; \pmb{\theta}_{v_m}),
\end{equation}
where $B_m(\pmb x; \pmb{\theta}_{v_m})$ is the $m$-th  basis function    {introducing PWL nonlinearity,}  $w_m$ is  the  coefficient, and $M$ is the number of basis functions.   {The general network architecture $\mathcal{N}$ resulting from equation \eqref{eq:basis}  is shown in Figure \ref{fig:pwl:basis}, where  one hidden layer is utilized. In such  PWLNNs,  the PWL mapping function $\phi_{v_m}$ of each hidden neuron $v_m, m=1,\cdots, M$, corresponds to each basis function $B_m(\pmb x;\pmb{\theta}_{v_m})$, which can be with learnable parameters $\pmb{\theta}_{v_m}$, where the incoming edges of each hidden neuron are 
from the $n$ neurons in the input layer. The PWL mapping function of the neuron ${v_{\rm output}}$ in the output layer is the weighted sum of the values from its incoming edges, where the weights correspond to $\pmb \theta_{v_{\rm output}} = \{w_m\}_{m = 1, \cdots, M}$.  In this type of PWLNN representations,} the canonical  and the hinge-based families are mainly involved.

\begin{figure*}[ht!]
\begin{center}
\subfigure[a] { \label{fig:eg:2}
\includegraphics[height=3.8cm]{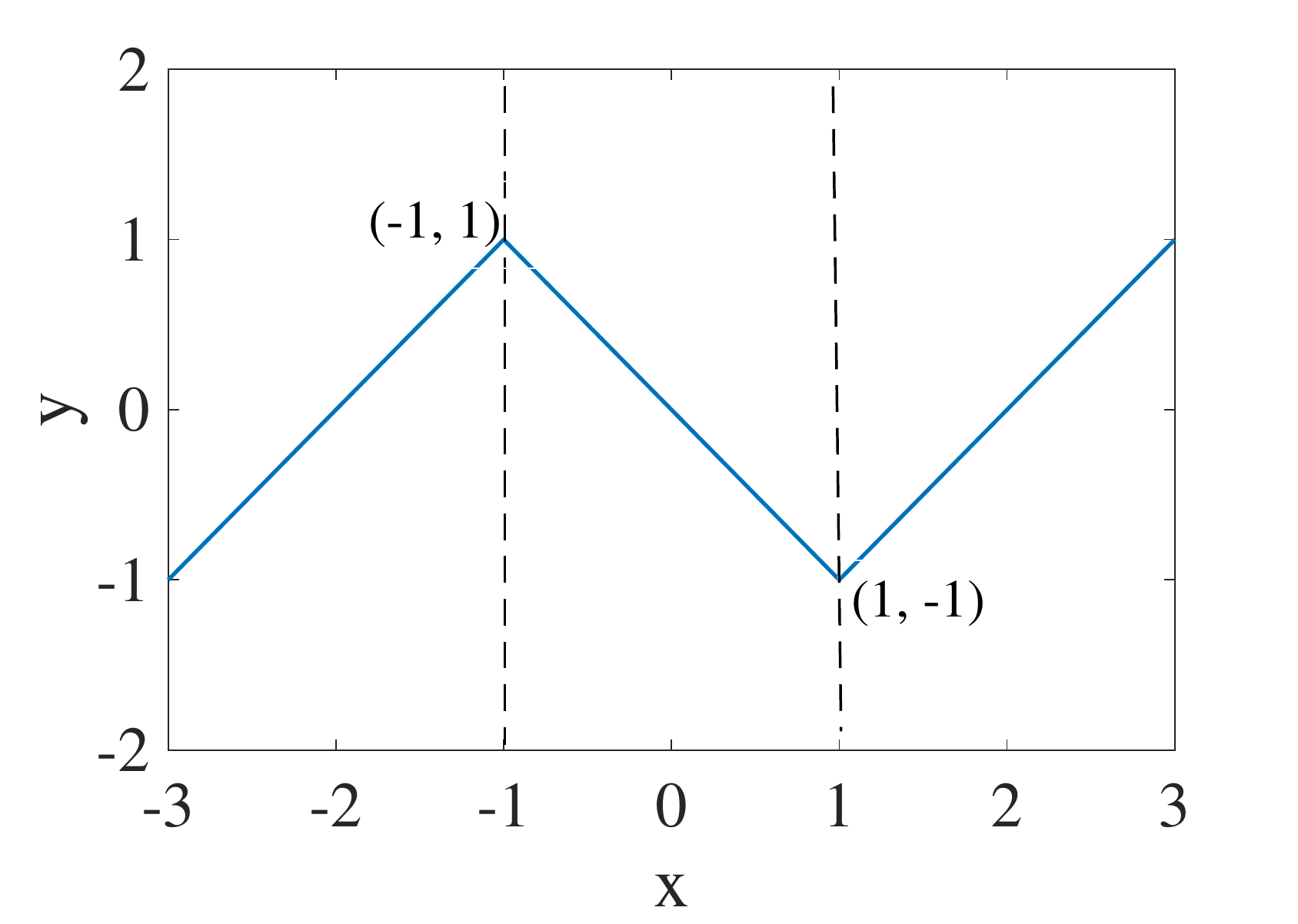}
}
\subfigure[b] { \label{2d-counter}
\includegraphics[width=0.7\columnwidth]{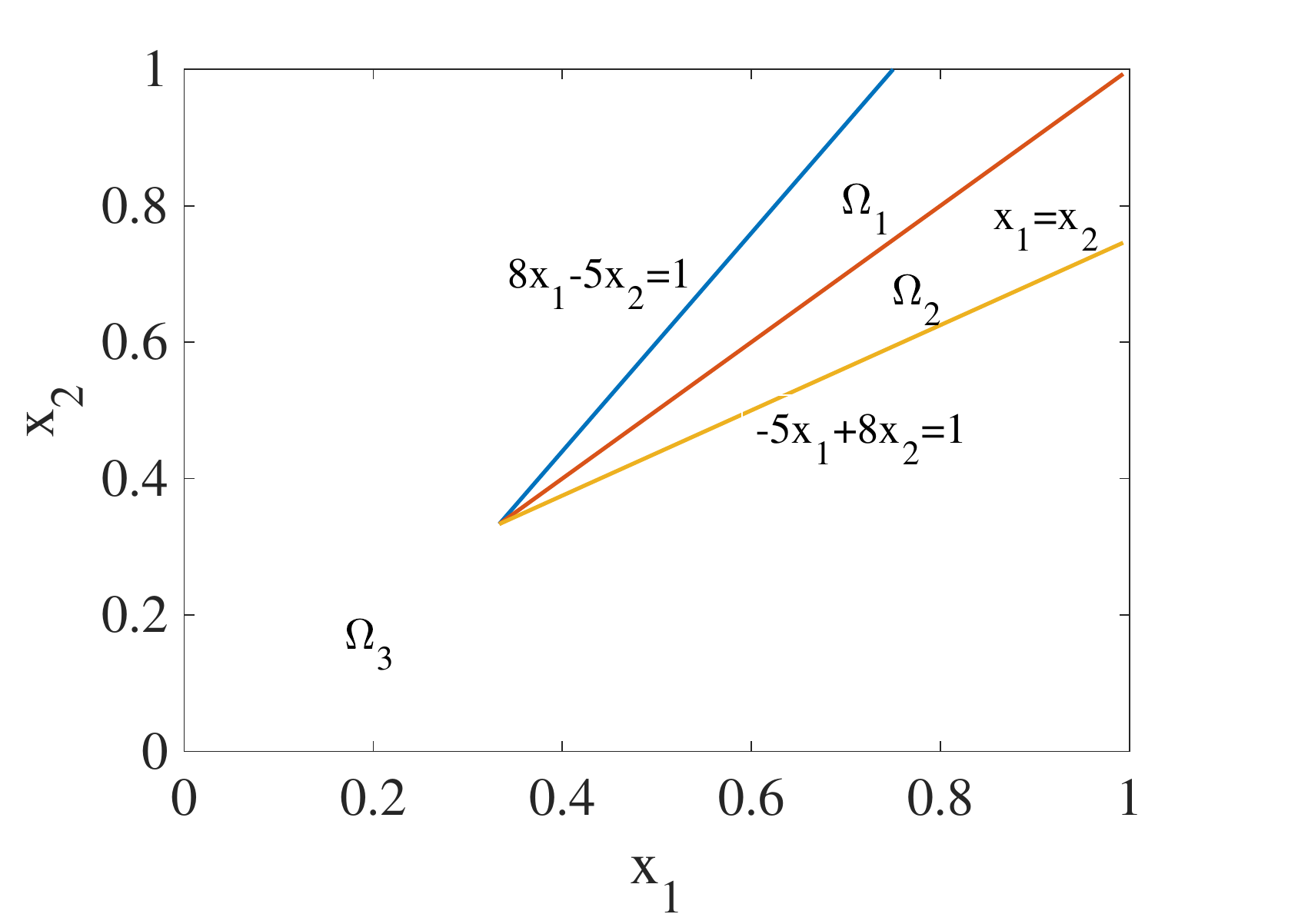}
}
\\
\subfigure[c] { \label{fig:ex2:shallow:pwlnn:1}
\includegraphics[width=0.55\columnwidth]{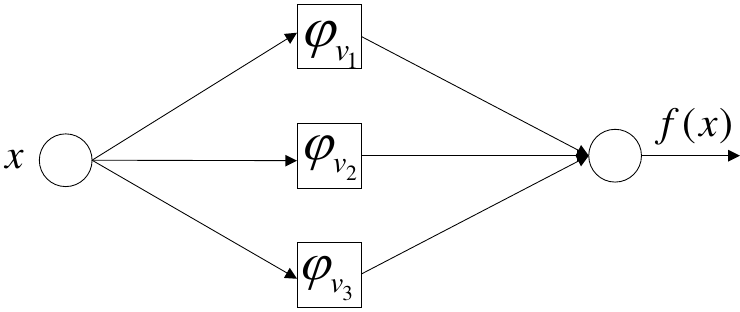}
}
\subfigure[d] { \label{fig:ex2:shallow:pwlnn:2}
\includegraphics[width=0.55\columnwidth]{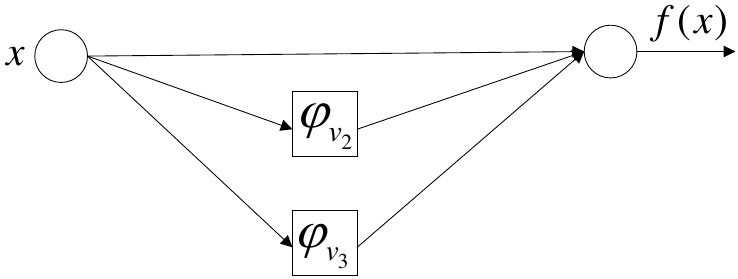}
}\\
\subfigure[e] { \label{fig:ex3:shallow:pwlnn}
\includegraphics[width=0.55\columnwidth]{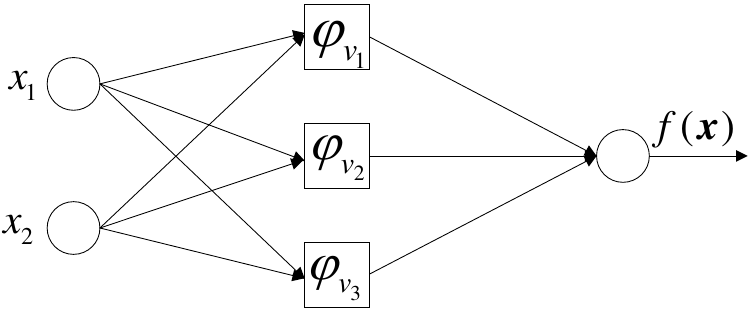}
}
\subfigure[f] { \label{fig:ex3:shallow:deep:pwlnn}
\includegraphics[width=0.55\columnwidth]{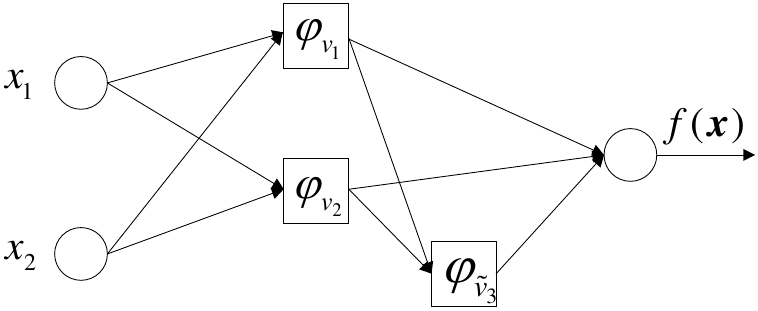}
}\\
\subfigure[g] { \label{fig:intro:pwl}
\includegraphics[width=0.55\columnwidth]{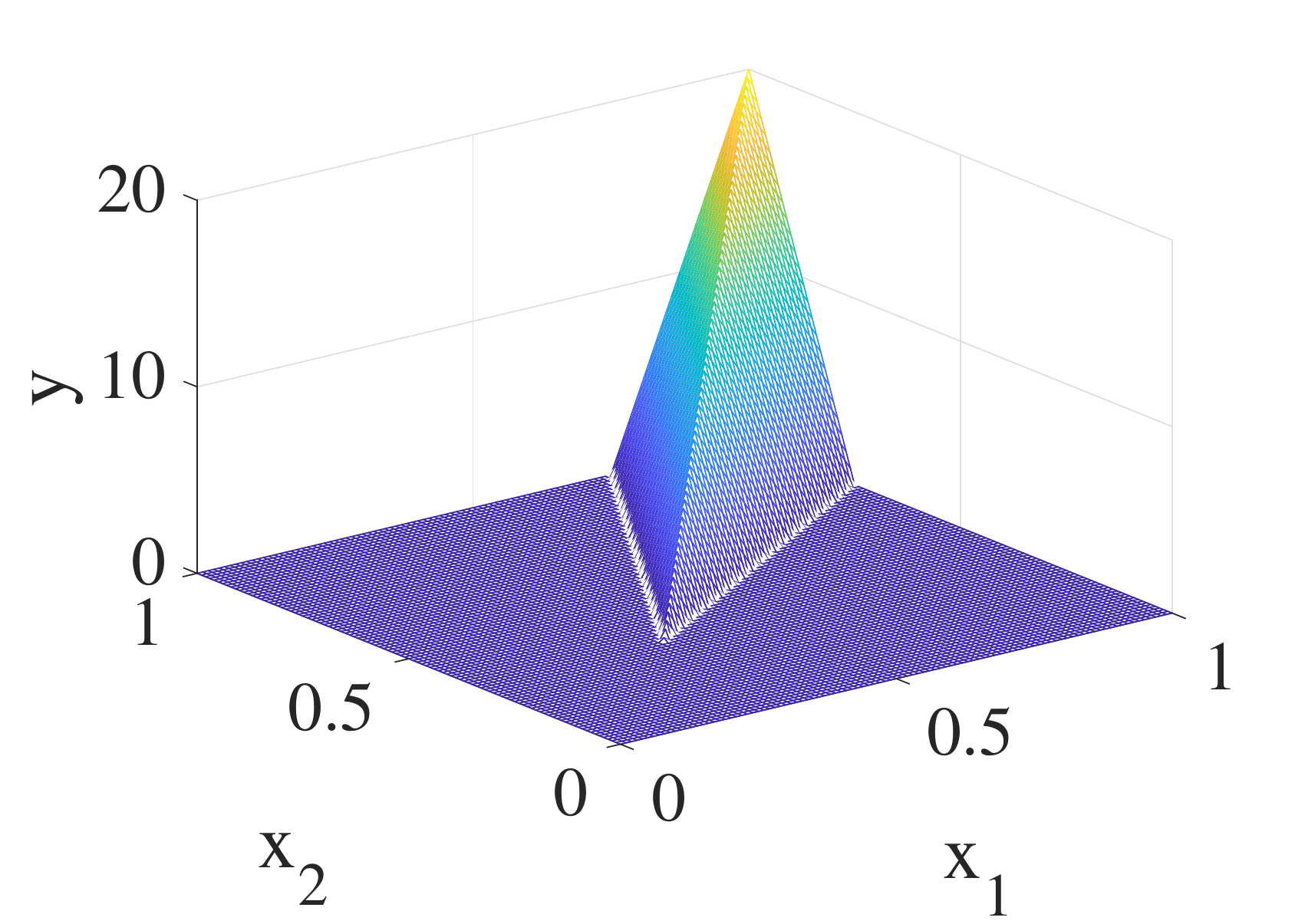}
}
\subfigure[h] { \label{fig:intro:fun}
\includegraphics[width=0.55\columnwidth]{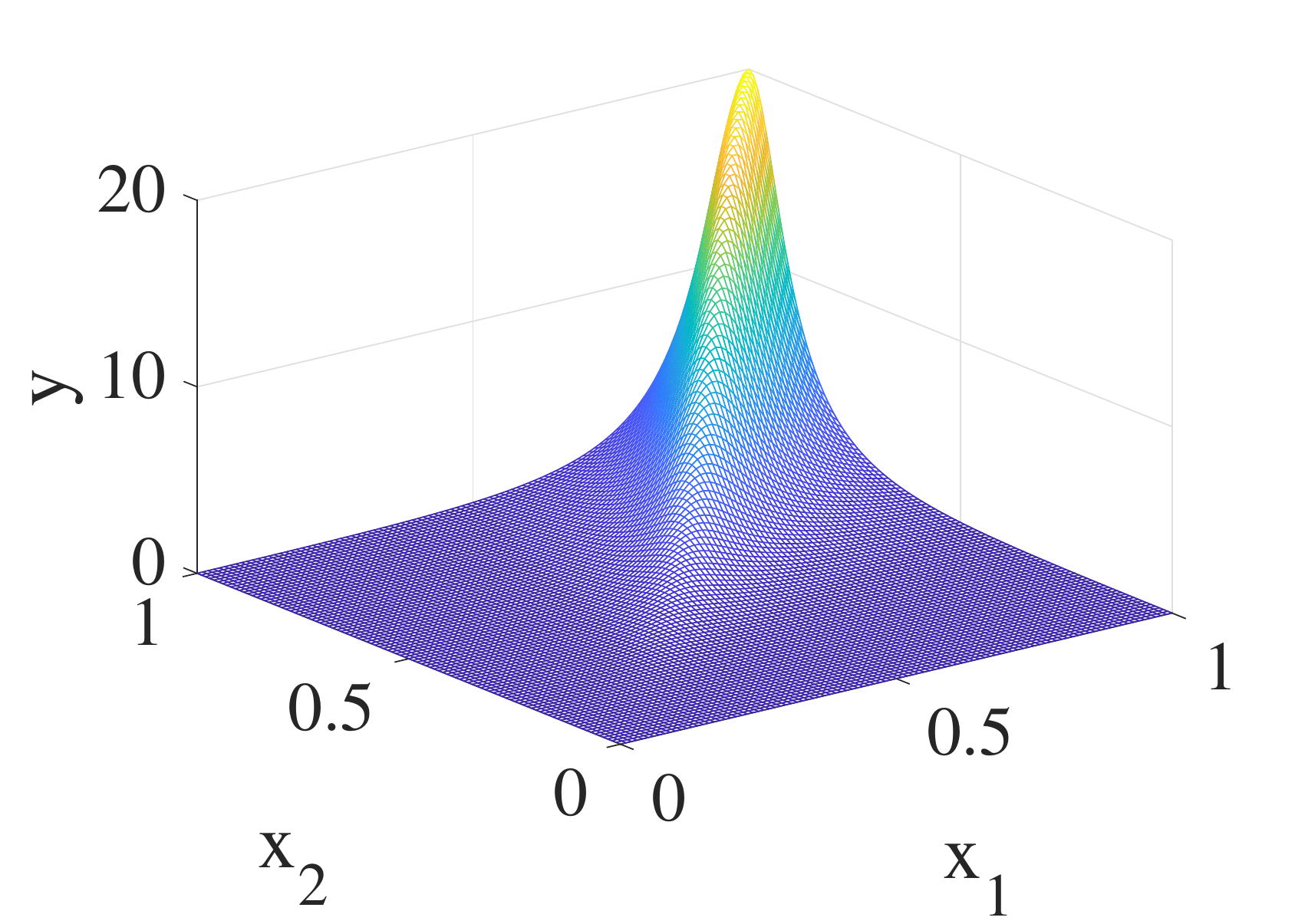}
}

\caption{Simple illustrations to visulise the resulting PWLNNs in equations \eqref{eqn:eg:2:cplr} and \eqref{eqn:ex:3:g-cplr}. a) Plot of equation \eqref{eqn:eg:2:cplr} to illustrate the canonical piecewise linear representation (CPLR) representation; b) Domain configuration of equation \eqref{eqn:ex:3:g-cplr}; c, d) shallow PWLNNs for \eqref{eqn:eg:2:cplr}; e) a shallow PWLNN for equation \eqref{eqn:ex:3:g-cplr}; f) a two-hidden-layered PWLNN for equation \eqref{eqn:ex:3:g-cplr}; g) Plot of equation \eqref{eqn:ex:3:g-cplr}; h) An exemplified function to approximate when considering using equation \eqref{eqn:ex:3:g-cplr}.}\label{fig:total:examples}
\end{center}
\end{figure*}

 CPLR  was originally proposed  in the univariate  formulation \cite{chua1977section} and  has been extended  to higher dimensions \cite{Kang1978Chua} by
\begin{equation}\label{eq:cplr:2}
f(\pmb x)= \pmb{\alpha}_0^T\pmb {x}+\beta_0 +  \sum_{m=1}^M \eta_m  | \pmb{\alpha}_m^T\pmb  x+\beta_m|,
\end{equation}
where $\pmb x\in \mathbb R^n$ is the input vector, $\eta_m =\pm 1$, $ \pmb{\alpha}_0,  \pmb{\alpha}_m, \in \mathbb{R}^n$ and $\beta_0, \beta_m \in \mathbb{R}$ are the parameters. Figure \ref{fig:eg:2} gives the plot of a simple PWLNN  for an illustration on the CPLR representation. For example, given a univariate PWL function $f(x)$ 
\begin{equation}\label{eqn:eg:2}
f( x)=\left\{
\begin{array}{lll}
x+2, & x\in (-\infty, -1], \\
-x, & x\in (-1, 1], \\
x-2, & x\in (1, \infty],
\end{array}\right.
\end{equation}
it can be represented by CPLR as follows
\begin{equation}\label{eqn:eg:2:cplr}
f(x) =  x -|x+1| +|x-1|,
\end{equation} 
with three basis functions, each of which corresponds to the PWL mapping function in each of the resulting hidden neurons.
As indicated in  Figure  \ref{fig:pwl:basis},  the output neuron $v_{\rm output}$  has incoming edges from three  hidden neurons in $\mathcal{H}=\{v_1, v_2, v_3\}$, and the output neuron's output is the weighted sum of the hidden neurons'  outputs from  its three incoming edges, where the weights are  $\pmb{\theta}_{v_{\rm output}} =\{1, -1, 1\}$.
Based on Box 1,  the network structure of this PWLNN is thereby  built upon the neurons $\mathcal{V} =\{v_{\rm input}, v_1, v_2, v_3, v_{\rm output}\}$ and the edges $\mathcal{E}=\{(v_{\rm input}, v_1)$, $(v_{\rm input}, v_2)$,  $(v_{\rm input}, v_3)$, $(v_1, v_{\rm output})$, $(v_2, v_{\rm output})$, $(v_3, v_{\rm output})\}$, as illustrated in Figure \ref{fig:ex2:shallow:pwlnn:1}.

 We can also remove $v_1$ from $\mathcal H$ and build a PWLNN with only two hidden neurons if we replace the edges with $(v_{\rm input}, v_{\rm output})$ in a  skip-layer manner (Figure  \ref{fig:ex2:shallow:pwlnn:2}). { It shows that a PWL function can be formulated
into varied PWLNNs with different network structures.}

 However, a  crucial problem exists: CPLR  can represent arbitrary PWL functions in  $\mathbb R$ but this representation ability is harmed  in  $\mathbb R^n (n\geq 2)$ \cite{chua1988}.  For example\cite{Wang2005GHH}, given a PWL function  $f: \mathbb R^2 \mapsto \mathbb R$ with $\pmb x=[x_1, x_2]^T$ as follows 
\begin{equation}\label{eqn:ex:3}
f(\pmb x)=\left\{
\begin{array}{ll}
l_1(\pmb x) = 80x_1 - 50x_2 -10 &  \pmb x\in \Omega_1,\\
l_2(\pmb x) = -50x_1 + 80x_2 -10 &   \pmb x\in \Omega_2,\\
l_3(\pmb x) = 0 &  \pmb x\in \Omega_3,\\
\end{array} \right.
\end{equation}
its  boundaries contain the ridges $x_1=x_2, 8x_1-5x_2=1$ and $-5x_1 + 8x_2=1$. However, these  ridges  vanish
 in the sub-region $\Omega_3=\{\pmb x| \ 3x_1\leq1, 2x_2\leq 1\}$, and such domain configurations cannot be realized by any CPLR. This incomplete representation ability can  essentially hurt the approximation performance when applying  PWLNNs to  data in the CPLR representation. For example,  Figure \ref{fig:intro:pwl}  can be an approximator for the function in Figure \ref{fig:intro:fun}, but it cannot be represented in the form of CPLR.  {Note that although the simple PWL approximator in Figure \ref{fig:intro:pwl} cannot be represented by CPLR, there exist other PWLNNs in the form of CPLR that can approximate Figure \ref{fig:intro:fun} to arbitrary accuracy, due to its universal approximation ability \cite{chua1988}.}

To overcome  the incomplete representation ability of CPLR in $\mathbb R^2$, the two-level nesting of CPLR is constructed \cite{Kahlert1990CS}:
\begin{equation}\label{eq:gcplr}
f(\pmb x)=\pmb{\alpha}_0^T\pmb {x}+\beta_0 +  \sum_{m=1}^M \eta_m |  {\pmb{\alpha}_{m,1}}^T\pmb  x + \beta_{m,1} + |{\pmb{\alpha}_{m,2}}^T\pmb  x+\beta_{m,2}||.
\end{equation}
Then, equation \eqref{eqn:ex:3}  can be represented as
\begin{equation}\label{eqn:ex:3:g-cplr}
\begin{array}{ll}
f(\pmb x)=&  7.5(x_1+x_2) - 5 -32.5|x_1-x_2| + \\
&|32.5|x_1-x_2|-7.5(x_1+x_2)+5|.
\end{array}
\end{equation}
Similar to equation \eqref{eqn:eg:2:cplr}, the resulting PWLNN of equation \eqref{eqn:ex:3:g-cplr}  has three hidden neurons  with  PWL mapping functions corresponding to  basis functions. The  neurons $\mathcal{V} =\{v_{\rm input, 1}, v_{\rm input, 2}, v_1, v_2, v_3, v_{\rm output}\}$ and the edges $\mathcal{E}=\{(v_{\rm input, 1}, v_1)$,  $(v_{\rm input, 2}, v_1)$, $(v_{\rm input, 1}, v_2)$,  $(v_{\rm input, 2}, v_2)$, $(v_{\rm input, 1}, v_3)$,  $(v_{\rm input, 2}, v_3)$, $(v_1, v_{\rm output})$, $(v_2, v_{\rm output})$, $(v_3, v_{\rm output})\}$ are obtained, as shown in Figure \ref{fig:ex3:shallow:pwlnn}.

 It is worth mentioning that different network structures can also be built for equation \eqref{eqn:ex:3:g-cplr} and it is of particular interest  for  $v_3$  having a PWL mapping function based on the two-level nesting of absolute-value operators.  Figure \ref{fig:ex3:shallow:deep:pwlnn} gives an illustration of  another PWLNN variant by reformulating  $v_3$ into $\tilde{v}_3$. This means equation \eqref{eqn:ex:3:g-cplr} can also be interpreted as a deep architecture with two hidden layers, where the original PWL mapping  function $|32.5|x_1-x_2|-7.5(x_1+x_2)+5|$ is equivalently formulated into a simplified form of $|-z_{v_2} - z_{v_1}|$  by transforming the incoming edges $(v_{\rm input, 1}, v_3)$ and $(v_{\rm input, 2}, v_3)$ from input neurons into  $(v_{1}, \tilde{v}_3)$ and $(v_{2}, \tilde{v}_3)$ from the 1st hidden layer. $v_1$ can also be reformulated with skip-layer connections to the output neuron as done in Figure \ref{fig:ex2:shallow:pwlnn:2}, but the details are not exhaustively presented here.

Nonetheless, in $\mathbb R^n$ ($n> 2$), the two-level nesting of CPLR  cannot cover  all PWL functions  \cite{Kahlert1990CS}. Thus,   more flexible representations are needed. 
Similar to the  two-level nesting, Generalized CPLR (G-CPLR) can be used \cite{Lin1994CS}.
 G-CPLR  refers to any finite number of nestings of CPLR  (defined as $K$-level CPLR with $K\geq 1$), which takes the form
\begin{equation}\label{eq:hlcplr}
f(\pmb x)= f_0(\pmb x) + C|f_1(\pmb x)|,
\end{equation}
where both $f_0(\pmb x)$ and $f_1(\pmb x)$ are models of  up to $K-1$ levels of nested CPLR.  Theoretically,  G-CPLR  can express all PWL functions in $n$ dimensions with at most  {$n$-level CPLR}  \cite {Lin1994CS}. 

With G-CPLR, the canonical formulations in equations (\ref{eq:cplr:2}) and \eqref{eq:gcplr}  can be regarded as the {one-level CPLR} and the {two-level CPLR}, respectively. 
 More generally than Figure \ref{fig:ex3:shallow:deep:pwlnn},  G-CPLR  can be regarded as a trial for  deep-architectured PWLNNs. However, G-CPLR was mainly of theoretical significance for universal representation ability, and an  effective learning algorithm has yet to be constructed.


\subsubsection{  Representations based on basis functions -- the hinge family}\label{sec:hh}
\begin{figure}[ht!]
\begin{center}
\subfigure[a] { \label{fig:1d:hinge}
\includegraphics[width=0.45\columnwidth]{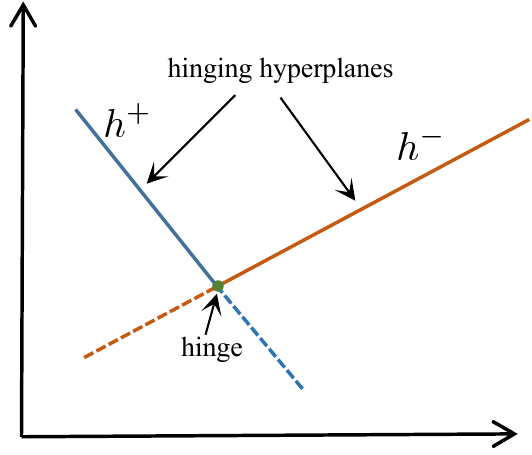}
}
\subfigure[b] { \label{fig:2d:hinge}
\includegraphics[width=0.45\columnwidth]{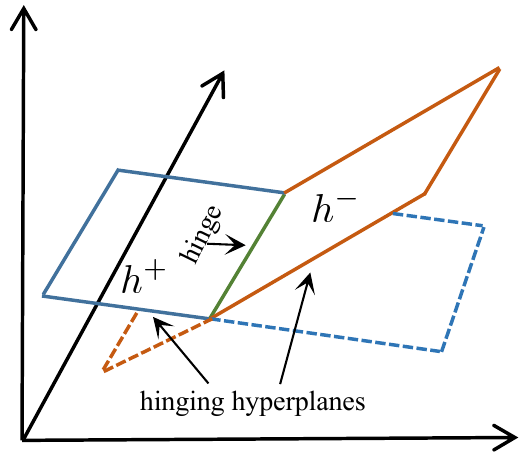}
}
\caption{{An illustration on the hinge function and its hinging hyperplanes. a) One-dimensional hinge; b) Two-dimensional hinge.}}\label{fig:example:hinge}
\end{center}
\end{figure}

The hinge family {refers to the  PWLNN representations based on} hinge functions where two hyperplanes continuously join together (Figure  \ref{fig:example:hinge}).

Given two hyperplanes $h^+ = {\pmb \alpha}_+^T\pmb x+\beta_+$ and $h^-={\pmb \alpha}_-^T\pmb x+\beta_-$ joining together at $\{\pmb x| ( {\pmb \alpha}_+ - {\pmb \alpha}_-)^T\pmb x = 0\}$,  their joint is defined as the hinge for  $h^+$ and $h^-$    {and is formulated as $\max\{h^+, h^-\}$. The  model of hinging hyperplanes  \cite{Breiman1993} is} given by
$
 f(\pmb x)=\sum_{m=1}^M w_m \max\{{\pmb \alpha}_{m_+}^T\pmb x+\beta_{m_+},{\pmb \alpha}_{m_-}^T\pmb x+\beta_{m_-}\},
$
commonly used as
\begin{equation}\label{eq:hh}
 f(\pmb x)=\pmb{\alpha}_0^T\pmb {x}+\beta_0  + \sum_{m=1}^M w_m \max\{{\pmb \alpha}_{m}^T\pmb x+\beta_{m},0\}.
 \end{equation}
Here, the hinges  have more distinguishable geometrical explanations than that of CPLR,  making it more amenable to design effective learning algorithms \cite{1998hingefinding,Pucar1998,Ramirez2004Implementation,huang2013hinging,DBLP:conf/smc/HuangXW10}. 

Similar to CPLR, the hinging hyperplanes still cannot represent all PWL functions in    {$\mathbb R^n (n\geq 2)$}.
To amend the incomplete representation ability, the Generalized hinging hyperplanes (GHH) \cite{Wang2005GHH} can be adopted by adding  a sufficient number of linear functions to the   hinges, such that
\begin{equation}\label{eq:ghh}
 f(\pmb x)= \sum_{m=1}^M  w_m \max\{{\pmb \alpha}_1^Tx+\beta_1, \ldots ,{\pmb \alpha}_{k_m+1}^Tx+\beta_{k_m+1}\}.
\end{equation}
Given that $k_m \leq n$, this is also referred to as the $n$-order Hinging Hyperplanes ($n$-HH); any $n$-dimensional PWL function can be represented by  an $n$-HH  \cite{Wang2005GHH}. 

Similarly,  equation \eqref{eqn:ex:3} can  be represented in the   GHH form:
\begin{equation}\label{eqn:ex:3:ghh}
\begin{array}{ll}
f(\pmb x)= &\max\{ 65(x_1 - x_2), 65(x_2 - x_1), 15(x_1 + x_2) -10\} \\
 & -\max\{65(x_1 - x_2), 65(x_2 - x_1)\},
\end{array}
\end{equation}
  which results in two hidden neurons with the PWL mappings based on $2$-HH.  The output neuron gives a weighted sum of the  neuron outputs from the hidden layer with weights $1$ and $-1$. In the form of GHH, equation \eqref{eqn:ex:3} can be represented by a PWLNN  with only two hidden neurons, resulting in  simpler network structures but requiring more flexible PWL mapping functions in the hidden layer.

 \subsubsection{  Representations based on basis functions -- others}\label{sec:other}
There are other basis functions, which are designed with joint considerations to  flexibility and learnability.
Based on simplicial  partitions and vertex interpolation,   High-Level CPLR (HL-CPLR)  \cite{Julian1999CS} can be constructed with the basis functions as
\begin{equation}\label{eq:hlcpwl}
B^R_{j_{k_{1}}, \ldots, j_{k_{R}}}(\pmb x)=  \max\{0, \min\{x_{k_{1}}-j_{k_{1}}d, \ldots,  x_{k_{r}}-j_{k_{r}}d,  \ldots, x_{k_{R}}-j_{k_{R}}d\}\},
\end{equation}
where $d$ is the partitioning interval, $k_{r}\in \{1, \ldots, n\}$  varies  with $r=1,\ldots, R$, and $j_{k_{r}}$ is selected from the  $\{1, \ldots, m_{k_{r}}\}$ partitions on the axis.    {Similar ideas of vertex-based modelling can also be found elsewhere\cite{padberg2000approximating,croxton2003comparison,Keha2006branch,vielma2010mixed}.}
 
By utilizing recursive domain partitions, the Adaptive Hinging Hyperplanes  (AHH) \cite{Xu2009AHH} is formulated as:
\begin{equation}\label{eq:ahh}
 f(\pmb x)=\sum_{m=1}^M w_m  \min_{j}\{ \max \{0, \delta_{j,m}({x}_{v_{j,m}} - \beta_{j,m})\}\},
\end{equation}
where  $x_{v_{j,m}}$ is the $v_{j,m}$-th input variable, and $\beta_{j,m}$ is the splitting knot, $j\in J_m$,    {$J_m \subseteq \{1, \ldots, n\}$ is the set containing the indices of  input variables involved in the $m$-th basis function}, and $\delta_{j,m} = \pm 1$.  
AHH can be regarded as a special case of GHH, and HL-CPLR  is a special case of AHH. 

The Simplex Basis Function 
(SBF)  \cite{yu2017incremental} model is  given by
\begin{equation}\label{eq:sbf}
f(\pmb x) = \sum_{m}^M w_m \max\{0, 1-\sum_{i=1}^n\gamma_{m, i}|x_i -\zeta_{m, i}|\},
\end{equation}
where $\gamma_{m, i}$ and $\zeta_{m, i}$ are the parameters controlling the shape of the $m$-th basis function. For any SBF,  equivalent transformations to HH and CPLR also exist.

\subsubsection{  Lattice Representations}\label{sec:lattice}
The Lattice representation \cite{lattice1963} is constructed based on the Lattice theory \cite{Birkhoff1958Lattice},  where simply the  ``$\max{\text{-}}\min$'' composition of linear functions is sufficient \cite{Terela1999Lattice}. It is worth mentioning that   the   ``$\max{\text{-}}\min$'' and the absolute-value operators for PWL nonlinearity have been addressed in
mathematical programming and functional analysis. For example,  major attention is given to aspects
of Lipschitz continuity, nondifferentiability, nonsmoothness and their algorithmic aspects, rather than to PWLNNs and data driven
applications \cite{streubel2013representation,DBLP:journals/oms/Griewank13,fiege2019algorithm,griewank2020polyhedral}.

Letting $f(\pmb x)$ be an arbitrary PWLNN  with $d$  distinct  linear functions, the Lattice representation  is formulated as
\begin{equation}\label{eq:lattice}
f(\pmb x ) = \underset{i\in \mathbf Z(M)}{\max}\ \underset{j \in S_i}{\min} \{\pmb J_j^T \pmb x + b_j\},
\end{equation}
where $\mathbf Z(M)=\{1, 2, \ldots, M\}$, $ [ \pmb J_1, \ldots, \pmb J_d ]\in \mathbb{R}^{n \times d}$, $[ b_1, \ldots, b_d ]^T \in \mathbb{R}^{d}$ and $S_i \subseteq \mathbf Z(d)$.  Equation \eqref{eqn:ex:4}  gives a simple illustration of the Lattice representation model. Given a  PWL function  $f: [0, 5] \mapsto \mathbb R$ 
\begin{equation}\label{eqn:ex:4}
f( x)=\left\{
\begin{array}{ll}
l_1(x) =  0.5x + 0.5, &  x\in \Omega_1=[0, 1],\\
l_2( x) = 2x - 1 &   x\in \Omega_2 = [1, 1.8 ],\\
l_3( x) = 2 &  x\in \Omega_3 = [1.8, 3.2], \\
l_4(x) = -2x + 9 &  x\in \Omega_4=[3.2, 4],\\
l_5( x) = -0.5x + 3 &   x\in \Omega_5=[4, 5],\\
\end{array} \right.
\end{equation}  
 its Lattice representation model is formulated as $M=5$, $S_1 =\{1, 3,4, 5\}$, $S_2 =\{2, 3,4, 5\}$, $S_3 =\{2, 3, 4\}$, $S_4 =\{1, 2, 3, 4\}$ and $S_5 =\{1, 2, 3, 5\}$.
   The resulting PWLNN based on the Lattice representation has 5 hidden neurons with  PWL  mapping functions based on the  ``$\min$''  operators across a maximum of 5 linear functions (Figure \ref{fig:pwl:lattice}). The PWL mapping function of the output neuron is simply the  ``$\max$''  operator across a maximum of 5 outputs of hidden neurons ($M$).
Although the Lattice representation is formulated quite differently from those based on basis functions, it plays a significant role in the theoretical analysis of different PWLNN representations to reveal  their intrinsic relations and modelling evolution.

 Given a PWL function,  PWLNNs with different network structures $\mathcal{V}$ and edges $\mathcal{E}$ can be constructed,  when  different PWL mapping functions are chosen for  neurons  $\mathcal{V}\setminus \mathcal{I}$.  Such PWL mapping functions are built upon  either the ``min",  the ``max",  the absolute-value operators, or their multi-level nestings. Often,  more flexible PWL mapping functions  result in less neurons and edges, {while with simpler PWL mapping functions the network} can be possibly  reorganized into  more  layers, for example the cases shown in Figure \ref{fig:total:examples}.

\subsection{  PWL-DNNs}\label{sec:nn}
Mainstream DNNs are sequentially composed of an input layer, multiple hidden layers and an output layer (Figure \ref{fig:pwl:nn}). Through the incoming edges, the outputs of neurons in 
 the previous layer are linearly weighted and summed. Upon application of an activation function,  nonlinearity is induced. 
Generally, the outputs of neurons in the $k$-th ($k\geq1$) layer in DNNs is computed as 
\begin{equation}\label{eq:nn:neuron:output}
\pmb{f}^k(\pmb x)= \sigma^k(\Xi^k(\pmb x)),
\end{equation}
where  $\Xi^k(\pmb x) = \pmb{W}^k\pmb{f}^{k-1}(\pmb x) + \pmb b^k$ denotes the weighted sum. Commonly in deep learning,  $\pmb{W}^k \in \mathbb R^{m_k \times m_{k-1}}$ and $\pmb  b^k \in \mathbb R^{m_k}$ are the weight matrix and the bias term for this neuron, respectively, where $m_k$ and $m_{k-1}$ refer to the numbers of neurons in the $k$-th and the $(k-1)$-th layers, respectively.  $\sigma^k(\cdot)$ is the nonlinear activation function,  usually in  a very simple form. 
In DNNs, the idea is to  introduce PWL activation functions $\sigma^k(\cdot)$ to neurons in the hidden layers, so that  PWL feature mappings are composited  across layers for greater flexibility. Following notations in Box 1, the  network architecture $\mathcal{N}=(\mathcal{V}, \mathcal{E})$ of  equation \eqref{eq:nn:neuron:output} confines  neurons in a hidden layer to only have incoming edges from  those in the previous layer, and the  corresponding  PWL mapping functions take the form of $\phi(\pmb{f}^{k-1}; \pmb{W}^k, \pmb b^k ) = \sigma^k(\pmb{W}^k\pmb{f}^{k-1} + \pmb b^k)$. They take the parameters in $ \pmb W^k$ as weights assigned to incoming edges of neurons, and $ \pmb b^k$  as a bias term to the weighted sum of the outputs of neurons in the previous layer.

It is  demonstrated that ReLU can significantly alleviate the {gradient vanishing problem [G]} in DNNs \cite{glorot2011deep}. The recent prevalence of ReLU in deep learning\cite{krizhevsky2012imagenet,DBLP:conf/cvpr/HeZRS16}  showcases the great flexibility and power of PWL-DNNs in various complex  tasks, though  the very first PWL neuron  stemmed from the modified threshold logic unit early in 1940s \cite{mcculloch1943logical} and  multi-layer  PWLNNs  were discussed in 1990s \cite{batruni1991multilayer,Lin1994CS,Lin1995Canonical}.
PWL activation functions, such as ReLU, are now acknowledged as the first choice in deep learning, owing to their advantageous performances such as better generalization performance and faster computation  \cite{DBLP:journals/neco/RawatW17}. 
In order to enhance network flexibility  and diversity, multiple variants of PWL activations have been proposed in deep learning, mostly by  re-shaping the hinge of ReLU. In fact, the resulting PWL mapping functions of the neurons deploying variant activations resemble the basis functions in the shallow PWLNN architectures. For instance,
 a shallow  neural network  with ReLU in equation \eqref{eq:nn:neuron:output}  is equivalent to the model of hinging hyperplanes, or can be equivalently transformed into  CPLR and SBF. Another typical example is the neural networks with  Maxout  \cite{Goodfellow2013MaxoutN}, which  resemble  GHH  \cite{Wang2005GHH}. 
  Table \ref{tab:cpwl:activation} summarizes the PWL activation functions in PWL-DNNs and their shallow-architectured counterparts. 
  
  \begin{table*}[ht]
\caption{Descriptions on the surveyed PWL activation functions in DNNs and their relations to  shallow  PWLNN representations.}\label{tab:cpwl:activation}
\scriptsize
\begin{center}
\begin{tabular}{lllll}
\hline  
Activation & Expression $\sigma(\cdot)$ & Description &Related Representations  \\
\hline
ReLU \cite{Nair2010Rectified} &       $\max\{x, 0\}$   & retaining the positive part &  CPLR, HH, SBF   \\
Leaky ReLU  \cite{leakyrelu}& $ \max\{x, 0\} - \lambda\max\{-x, 0\}$ &  $\lambda$ as a constant nonzero slope&CPLR, HH, SBF   \\
Parametric ReLU   \cite{he2015delving}&   $\max\{x_c, 0\} - \lambda_c\max\{-x_c, 0\}$ & $\lambda_c$ as   trainable  slopes in different channels $x_c$ &  CPLR, HH, SBF   \\
Randomized ReLU   \cite{DBLP:journals/corr/XuWCL15}&  $\max\{x^{(n)}_c, 0\} - \lambda_c^{(n)}\max\{-x^{(n)}_c, 0\}$  & $\lambda^{(n)}$ as   randomized  slopes  of sample $x^{(n)}$ &  CPLR, HH, SBF  \\
Biased ReLU \cite{DBLP:journals/ijon/LiangX21}& $\max\{0, x_i - b_{i1}\}, \cdots, \max\{0, x_i - b_{iq_i}\}$ & several biases for each input variable $x_i$&   AHH, HL-CPLR   \\
Concatenated ReLU   \cite{DBLP:conf/icml/ShangSAL16} & $(\max\{x,0\}, \max\{-x, 0\})$ & concatenating both positive and negative parts &  CPLR, HH, SBF    \\
S-shaped ReLU \cite{2015Deep}& $ \alpha_{0} {x}+\beta_{0} +  \alpha_{1} |x - t^{\rm l}|+  \alpha_{2}|x- t^{\rm r}|$ & 3  linear intervals with break points $t^{\rm l}$ and $t^{\rm r}$ &  CPLR, HH, SBF \\
Flexible ReLU \cite{DBLP:conf/icpr/QiuXC18}& $\max\{x+a, 0\} +b$ & $a$ and $b$ as trainable parameters &  CPLR, HH, SBF  \\
APL  \cite{DBLP:journals/corr/AgostinelliHSB14} & $\max\{x, 0\} + \sum_{s=1}^Sa_i^s \max\{0, -x + b_i^s\}$ &  a sum of $S$ hinge-shaped functions &  CPLR, HH, SBF   \\
APRL  \cite{DBLP:conf/dcsmart/BodyanskiyDPS19}& $ \lambda_{\rm R}\max\{x, 0\} - \lambda_{\rm L}\max\{-x, 0\}$& $ \lambda_{\rm R}$ and $ \lambda_{\rm L}$ as trainable slopes  &  CPLR, HH, SBF  \\
Maxout   \cite{Goodfellow2013MaxoutN}& $\max_{i\in I}\{z_i\}$ & maximal of multiple inputs & GHH\\
\hline
\end{tabular}
\end{center}
\end{table*}

As indicated in Box 1, PWLNNs can be flexibly designed in many ways. Firstly, with the edges in $\mathcal{E}$, the connections between neurons in $\mathcal{V}$ can be formulated in different ways through the mappings (shallow PWLNN representations), rather than simply the linearly weighted sum on the previous layer (deep learning). Skip-layer connections are also naturally allowed for any paired neurons. Skip-layer connections have been widely applied to shallow PWLNN representation models, such as the PWLNN in Figure \ref{fig:ex3:shallow:deep:pwlnn} and the variant of AHH \cite{EHH}, which enables sparse and even decomposable network structures with great interpretability. In recent years,  the flexible skip-layer connections have also received wide attention in deep learning, such as the ResNet \cite{DBLP:conf/cvpr/HeZRS16} and its generalization, the DenseNet \cite{DBLP:conf/cvpr/HuangLMW17}. Secondly, the mapping function $\phi_v$ can  be multi-dimensional, meaning that each incoming edge of $v$ can be assigned weights in vectors, such as the PWL-DNNs with Maxout and  shallow PWLNNs with GHH and AHH. The activation functions in PWL-DNNs can also include learnable parameters for more flexible PWL mappings for each neuron, such as in S-shaped ReLU \cite{2015Deep} and APL  \cite{DBLP:journals/corr/AgostinelliHSB14}, where multiple breakpoints exist and can be made learnable. Similar ideas were realised  early in the 1990s \cite{suykens1997family}.

Figure  \ref{fig:pwl:architecture} compares the topology of PWLNNs. 
With limited computational resources,  shallow architectures are usually the better choice for  small-scale problems, owing to their flexibility and the alleviation of overfitting and model redundancy issues. For example, SBF is more efficient in its learning scheme and shows to be more robust against noises, whilst AHH and the Lattice have better interpretability. In particular, when  the contributions of input variables or the effects of variable connections need to be explored, AHH is a good predictor. When local explicitness is required, the Lattice is the preferred choice, as in the explicit Model Predictive Control (MPC) problems. When tremendous data are given and highly complicated tasks are involved, PWL-DNNs are the superior choice, owing to their optimization techniques and mature computational platforms. Though no single model is capable of resolving all problems, there are a number of well-designed techniques in PWLNNs available for different demands.
 PWLNNs are natural  embodiments of PWL functions exerting network structures, so that  the powerful learning ability and other merits of (D)NNs  can be implemented.  At the same time, the introduction of PWL nonlinearity helps resolve some learning barriers  in deep learning  with  significant  developments. PWLNNs and deep learning have boosted each other, together bringing unprecedented success in the big data era.

\section{ Results}\label{sec:analysis}
Learning algorithms are the key to applying PWLNN models to practical  tasks. This includes how to effectively determine the network parameters and structures to well resolve the tasks. A chronicle overview on learning algorithms can be found in the supplementary information.  In this section, we look into PWLNN models and their learning with theoretical analysis to better understand the mechanism.

\subsection{  Learning shallow PWLNNs}\label{sec:learning:shallow}

Shallow PWLNNs are commonly learned  incrementally. Computational  platforms and hardware were  limited in the 1970s-2000s, and so the incremental design was a good choice to balance efficiency and accuracy. These  incremental  learning algorithms differ depending on the representations; they can be categorized into  the hinge finding algorithm, the tree searching  algorithm, the structured  decision  algorithm, and others.

\subsubsection{  Hinge finding algorithm and Newton's algorithm}
To learn the hinging hyperplanes,  new basis functions are incrementally added in each ($M$-th) iteration by adopting the hinge finding algorithm on $y = f(\pmb x)-\sum_{m=1}^{M-1} w_mB(\pmb x)$ and adjusting the sum $\sum_{m=1}^{M} w_mB(\pmb x)$. The key step in hinge finding is 
to perform the {least squares method [G]} to determine the estimated parameters of the hyperplanes fitted to  $S_-$ and $S_+$; 
\begin{equation}\label{eq:ls:ha}
\begin{array}{ll}
{\pmb \alpha}_{m_+} = \left( \sum_{\pmb x^i \in S_+}  \pmb x^i{\pmb x^i}^T  \right )^{-1} \sum_{\pmb x^i \in S_+} \pmb x^iy^i, \\
{\pmb \alpha}_{m_-} = \left( \sum_{\pmb x^i \in S_-}  \pmb x^i{\pmb x^i}^T  \right )^{-1} \sum_{\pmb x^i \in S_-} \pmb x^iy^i,
\end{array}
\end{equation}
where $S_+ = \{\pmb x: \pmb x^T({\pmb \alpha}_{m_+} - {\pmb \alpha}_{m_-}) >0\}$ and $S_- = \{\pmb x: \pmb x^T({\pmb \alpha}_{m_+} - {\pmb \alpha}_{m_-}) \leq 0\}$.

In this way,  new basis functions are incrementally obtained by locating the hinge functions over the training data. 
 In fact, the estimation in the hinge finding is a special case of the  {Gauss-Newton algorithm [G]}  \cite{Pucar1998}. 
 Similarly, the learning  of GHH can inherit such algorithm, where multiple linear functions are involved in each hinge \cite{wangsun2007}.
Simultaneously identifying all basis functions is also possible  \cite{Pucar1998}, but whether the sequential or simultaneous update of basis functions is the better option, if the computational resources are  adequate, remains an open question.

\subsubsection{   Tree Searching Algorithm}
\begin{figure}[ht!]
\centering
\subfigure[a] { \label{fig:partition:1}
\includegraphics[width=0.22\columnwidth]{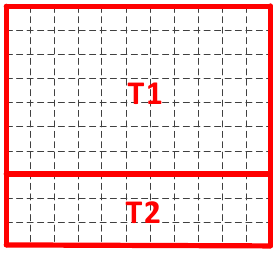}}
\subfigure[b] { \label{fig:partition:2}
\includegraphics[width=0.22\columnwidth]{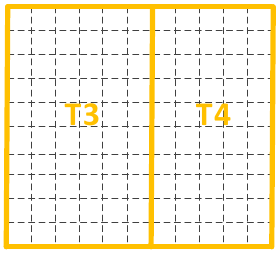}}
\subfigure[c] { \label{fig:partition:3}
\includegraphics[width=0.22\columnwidth]{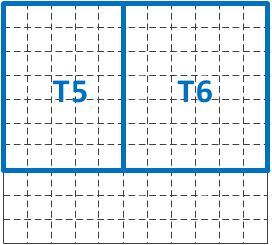}}
\subfigure[d] { \label{fig:partition:4}
\includegraphics[width=0.22\columnwidth]{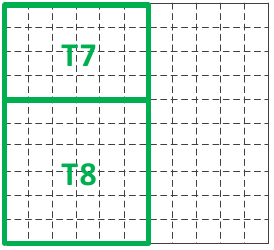}}
\subfigure[e] {\label{fig_ahh_ehh}
\includegraphics[width=0.5\columnwidth]{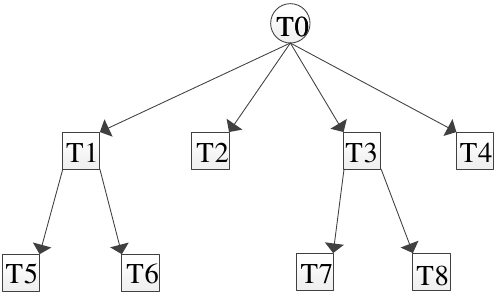}}
\caption{ A simple illustration on  the geometrical description and tree searching of learning an adaptive hinging hyperplane (AHH). 
a) Iteration 1 where $B_{1}(\pmb x) =\max\{0, x_2-0.3\}$, $B_{2}(\pmb x)=\max\{0, 0.3-x_2\}$; 
b) Iteration 2 where  $B_{3}(\pmb x)=\max\{0, 0.6-x_1\}$; 
c) Iteration 3 where $B_{4}(\pmb x)=\max\{0, x_1-0.6\}$, $B_{5}(\pmb x)=\min\{B_{1}, B_{3}\}$, $B_{6}(\pmb x)=\min\{B_{1},B_{4}\}$; 
d) Iteration 4 where $B_{7}(\pmb x)=\min\{B_{1}, B_{3}\}$, and $B_{8}(\pmb x)=\min\{B_{2},B_{3}\}$; 
e) the resulting generic tree topology.}
\label{fig_ahhtree_partition}
\end{figure}

AHH can be seen as a PWL analogy to the  {multivariate adaptive regression splines\cite{friedman1991multivariate} [G]}. 
Similarly,  the  learning  of AHH is based on the recursive domain partitions by deploying the tree searching algorithm; the learning process of AHH  can be interpreted as a generic tree with  basis functions $B_{m}(\pmb x)$ as leaf nodes (Figure  \ref{fig_ahhtree_partition}). Each basis function (neuron) $B_m(\pmb x)$  in Iterations 1-4  correspond to a sub-region ${\rm{T}}_m$, giving the resulting  tree topology of learning  AHH (Figure \ref{fig_ahh_ehh}).
The tree searching algorithm does not require computed gradients, and gives a novel interpretation of learning. In classic decision trees,  the concept of PWL approximation can be applied to fit linear models on  nodes \cite{wang1996induction}. However, these local linear regressions bring substantially higher computational burden. 
A novel PWL decision tree as a  flexible and efficient alternative has been previously constructed\cite{A2020Learningtao}, which can be regarded as the extension of ReLU to the learning framework of decision trees.

\subsubsection{  Structured Decision Algorithm}
\begin{figure}[ht!]
\centering
\subfigure[a] { \label{fig:sbf:1}
\includegraphics[width=0.38\columnwidth]{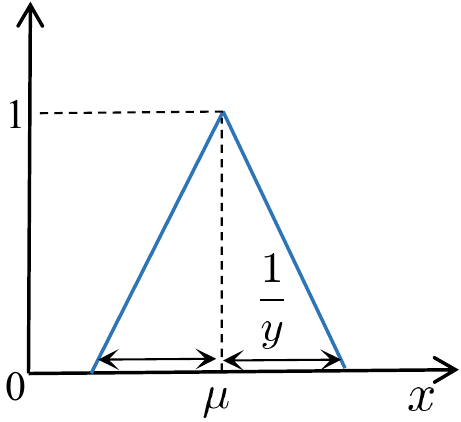}
}
\subfigure[b] { \label{fig:sbf:2}
\includegraphics[width=0.44\columnwidth]{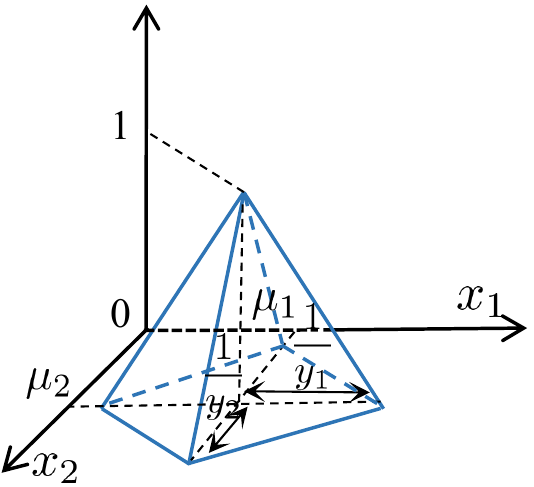}
}
\caption{{An illustration on the basis functions of simplex basis function (SBF) representation. a) One-dimensional SBF; b) Two-dimensional SBF.}}\label{fig:sbf}
\end{figure}

The basis functions in SBF are simplices and their shapes are controlled by   $\gamma_{m, i}$ and $\zeta_{m, i}$ (Figure \ref{fig:sbf}).

SBF takes the structures of basis functions as decision variables to design the learning algorithm rather than directly taking the coefficient $w_m$ as decision variables.  In each iteration, the current  highest peak point is selected as the center $\zeta_i$. It then seeks
the optimal structure of the simplices  to minimize the overall approximation error \cite{yu2017incremental}.

\subsubsection{  Other Algorithms}  
PWLNNs can be learned  by interpolating the vertices over simplices and performing local linear approximation. Following this idea,  HL-CPLR can be adopted. Without specifying  domain configurations, linear approximation in sub-regions has been successfully applied in other learning algorithms for PWLNNs. For example, bi-level algorithms can be used, but they are  only suitable  in  low dimensions with a few sampling  data \cite{ferrari2003clustering,nakada2005identification}.
For the Lattice, the structure could be identified by repeatedly  updating the parameters by  local fitting, where neither numerical simulations nor practical implementations were  given \cite{2002Nonlinear}.

\subsection{  Learning PWL-DNNs}\label{sec:learning:deep}
In PWL-DNNs, the optimization problem is highly nonconvex and  complicated, particularly for high-dimensional and large-scale  data. Fortunately, the stacking network topology has enabled the optimization problem to be well resolved by successful applications of backpropagation and SGD.

\subsubsection{   {  Optimization of Network Parameters}}\label{sec:learning:deep:parameters}
With backpropagation, the chain rule can be applied to compute the gradients of learnable parameters, and then  gradient-based algorithms can be  applied.  Specifically, denoting the weighted input of the $k$-th layer as $\pmb z_k$ and the output of the $k$-th layer as the activation $\pmb \sigma_k$ for $k=1, \ldots, K$, see equation \eqref{eq:nn:neuron:output}. The activation $\pmb \sigma_k$ and the derivatives  evaluated at $\pmb z_k$ in each layer are computed to be cached and then used to complete the backward propagation of gradients by the chain rule. In  terms of matrix multiplication, the derivative of the loss with respect to the inputs is given by
\begin{equation}\label{eq:bp}
\frac{\partial \mathcal{L}(g(\pmb x|\mathcal{N}))  }{\partial \pmb \sigma_K}\cdot \frac{\partial \pmb \sigma_{K}}{\partial \pmb z_K}
\cdot \frac{\partial \pmb z_{K}}{\partial \pmb \sigma_{K-1}} \cdots \cdot  \frac{\partial \pmb \sigma_{1}}{\partial \pmb z_{1}} \cdot \frac{\partial \pmb z_{1}}{\partial \pmb x},
\end{equation}
 where $\mathcal{L}(\cdot)$  is a general loss and  $g(\pmb x|\mathcal{N})$ is the  PWL-DNN.

In such learning schemes to compute the gradients in PWL-DNNs,  powerful graphical or tensor processing unit hardware can be equipped to greatly accelerate the computation in the learning process. 
The massive amounts of data and considerably deep architectures make the optimization problems much more complicated, hence the computational efficiency of gradient descent algorithms is limited. For this reason, the stochastic version, namely SGD \cite{bottou1991stochastic}, is the mainstream algorithm in deep learning, in which the batch-wise optimization accelerates the computation and improves the generalization ability. It also helps escape from the saddle points efficiently \cite{jin2021nonconvex}. After the success of SGD, numerous variants were developed with more sophisticated techniques {to further enhance the performance in deep learning, such as adaptive learning rates \cite{duchi2011adaptive}, gradients with momentum \cite{DBLP:journals/corr/KingmaB14}, preconditioned gradients \cite{gupta2018shampoo}, and second-order methods \cite{anil2020scalable}}.

In DNNs, the  gradient vanishing problem is devastating, when the computation of gradients is propagated over layers. Fortunately, this problem is greatly alleviated by using  ReLU \cite{glorot2011deep}, allowing PWL-DNNs to be designed fairly deep (more than hundreds of layers) \cite{krizhevsky2012imagenet}. Beside the gradient-based algorithms for parameters,  a robust initialization method is constructed by giving specified considerations to  PWL nonlinearity, which enables to  train
extremely deep networks  from scratch \cite{he2015delving}.  Other useful techniques for generic DNNs can be utilized to improve the learning performance of PWL-DNNs, such as  dropout \cite{srivastava2014dropout}, batch normalization  \cite{ioffe2015batch}, data augmentation \cite{shorten2019survey}, and pre-training \cite{erhan2010does}.

Concerning the PWL nonlinearity in deep learning, specialized algorithms have also been studied by utilizing the different local linear expressions of activated neurons to design  novel algorithms. An adaptation of path-SGD is proposed to learn plain recurrent neural networks with ReLU,  capturing long-term dependency structure and significantly improving the learnability\cite{neyshabur2016path}. By defining  a transformation space of the positive scaling operators from ReLU, a modified SGD can be constructed in such a space, outperforming the conventional SGD\cite{DBLP:conf/iclr/MengZZCYMYL19}. By assuming linearly separable data and utilizing local linearity of ReLU, a novel SGD is also constructed \cite{wang2019learning}. Further, by  partitioning node inputs of PWL-DNNs,  the convex hull over the partitions can be obtained, and thus mixed-integer  and even convex optimization can be performed \cite{tsay2021partition,ergen2021global}. However, these  algorithms  currently require stringent assumptions and simple network structures with only a few layers, and yet are inapplicable  to the learning of generic PWL-DNNs. Nevertheless, they innovate to exploit the benefits and utilize the special characteristics of  PWL nonlinearity,  promoting more in-depth understandings in learning deep architectures.

\subsubsection{{  Optimization of Network Structures}}\label{sec:learning:deep:parameters}
Currently, the regular learning of PWL-DNNs is inherited from generic DNNs, where network structures are predefined through trial and error procedures, based on the researchers’ prior knowledge and experience. The learnable parameters of interconnection weights and bias terms are then optimized with the aforementioned algorithms. In contrast, the learning algorithms in shallow  PWLNNs involve simultaneous determination of both network structures and parameters, and they are model dependant by incorporating specific  characteristics in each representation.

The greater flexibility of DNNs results in substantial redundancy in the network structures, leading to increasing efforts to optimize network structures in recent years.
There have been numerous studies on compressing DNNs while guaranteeing similar or even better accuracy, such as sparsity regularization \cite{wen2016learning}, neuron connections pruning \cite{han2015learning} and low rank approximation \cite{denton2014exploiting}. Recently, the lottery ticket hypothesis shows that randomly-initialized DNNs contain sub-networks (winning tickets) which are capable of reaching similar accuracy when being trained in isolation \cite{DBLP:conf/iclr/FrankleC19}. Rather than extracting a more compressed model from an existing DNN, Neural Architecture Search (NAS) \cite{DBLP:conf/iclr/ZophL17} is proposed to gradually generate DNNs by searching the optimal building block candidates from a predefined search space via well-designed evaluation strategies. Although these algorithms are not limited to PWL-DNNs, they commonly adopt ReLU  to introduce nonlinearity.

 In fact, algorithms for the optimized network structures can also be found in the learning of different shallow PWLNNs, such as the $l_1$ norm sparsity regularization \cite{tao2018fast} and the backward pruning \cite{Xu2009AHH}. NAS is  an incremental design, but seeks for both width and depth in network architectures, instead of merely growing the width in shallow architectures. In NAS, the candidate building blocks are network cells, which are commonly chosen as PWL sub-networks and are much more complicated, compared to those simple formulations of PWL basis functions. Though shallow PWLNNs are limited in empirical performances and problem scales, their learnings are also strongly related to current deep learning.

The success of effective learning techniques in deep learning now fundamentally realizes the pervasive applications of PWLNNs.  In turn, the PWL-DNN itself also boosts thriving developments to generic DNNs, leading to a win-win situation between deep learning and PWLNN methods.

\subsection{  Analysis on Shallow PWLNNs}\label{sec:analysis:shallow}
In shallow  PWLNNs,  the ability to approximate arbitrary continuous functions (approximation ability) and the ability to represent arbitrary PWL functions (representation ability) have been the main focus in theoretical analysis.

\subsubsection{  Universal Approximation and Representation Abilities}
\label{sec:analysis:approximation:representation}
 {The approximation  and representation abilities are two different but important aspects in reflecting the properties of a class of PWLNNs  in the form of a specific representation model.}
 Table \ref{table:compare:appro:repre} summarizes these two properties. 

\begin{table}
\caption{Comparison on the universal approximation and representation abilities of different PWLNN representations.}\label{table:compare:appro:repre}
\scriptsize
\begin{center}
\begin{tabular}{lll}
\hline 
Representation Model &  Approximation &    Representation \\
\hline
Conventional Representation\cite{chien1977solving} &           \checkmark            &  \checkmark \\
CPLR \cite{chua1977section} &   \checkmark  &  only for $\pmb x \in \mathbb R$ \\
G-CPLR  \cite{Lin1994CS}& \checkmark   & \checkmark  \\
HL-CPLR \cite{Julian1999CS} &   \checkmark & {\tiny{\XSolidBrush}} \\
HH \cite{Breiman1993}&  \checkmark  & only for $\pmb x \in \mathbb R$  \\
GHH \cite{Wang2005GHH} &   \checkmark & \checkmark\\
AHH \cite{Xu2009AHH}& \checkmark    & {\tiny{\XSolidBrush}}  \\
SBF  \cite{yu2017incremental} & \checkmark   & only for $\pmb x \in \mathbb R$ \\
Lattice \cite{Terela1999Lattice}&  \checkmark  & \checkmark \\
\hline
\end{tabular}
\end{center}
\end{table}

Compared to the universal approximation,  representation ability is more difficult to attain for a shallow PWLNN representation. It has been proven that the aforementioned representation models of  shallow PWLNNs  have  universal approximation ability for continuous functions, but not necessarily have universal representation ability for continuous PWL functions, such as the counter example in equation \eqref{eqn:ex:3}.  {More specifically, given a continuous function and a specific PWLNN representation model having universal approximation ability, there always exist proper PWLNNs that can approximate this continuous function to arbitrary accuracy, while given a certain PWLNN as an approximator for this continuous function, such  PWLNN  might not be expressed in the form of this specific representation model.}  Hence, it is of great significance to select a representation model when approximating a nonlinear system with sampled data, since different selected representation models can lead to different PWLNNs with varied properties and difficulties in implementing the approximation \cite{Wang2004General}.

\subsubsection{  Model Properties and Their Connections}
As the pioneering compact representation,  the  existence conditions of CPLR are as follows. A PWL function $f(\pmb x)$ has a CPLR representation  of equation (\ref{eq:cplr:2}) if and only if it satisfies the  {consistent variation property\cite{chua1988} [G]}. A PWL function $f(\pmb x)$  possesses the  consistent variation property if and only if $f(\pmb x)$  is partitioned by a finite set of  hyperplanes 
$
\mathcal H_k=\{\pmb x\in \mathbb R^n: \pi_k(\pmb x)\coloneqq \alpha_k^T\pmb x - \beta_k=0 \},
$
 for $k = 1, \ldots, h$;  $\forall k=1, \ldots, h$ partitioning hyperplane  $\mathcal H_k=\{\pmb x\in \mathbb R^n: \pi_k(\pmb x)\coloneqq \alpha_k^T\pmb x - \beta_k=0 \}$, the dyadic product of any two local linear functions intersecting at the common boundary $\mathcal H_k$, meaning that $\triangle \pmb J^{(i,j)}  \doteq  \pmb J_i -  \pmb J_j = c^{(i,j)} \pmb \alpha$ and $
\triangle b^{(i,j)}  \doteq  b_i - b_i = c^{(i,j)} \beta, \ c^{(i,j)} \in \mathbb R$ remains consistent. For every pair of 
neighboring sub-regions separated by a common 
boundary, the intersection between  such two local linear functions must 
be a subset of an $(n - 1)$-dimensional hyperplane and 
cannot be covered by any hyperplane of lower dimensions.

When  constructing PWLNNs using the Lattice representation, some interesting and fundamental properties are  obtained. Firstly, the Lattice representation  (any PWL function) can be transformed into the difference of two convex PWL functions \cite{Wang2004General}. For example, for any positive integers $n$ and $d$, nonempty index sets $S_i \subseteq \mathbf{Z}(M), i\in \mathbf Z(M)$ and  real vectors $\pmb \theta(j)\in \mathbb R^{n+1}$, $j\in \mathbf Z(d)$, there always exist  $n+1$-dimensional real vectors $\pmb  \theta_+(k)$, $k\in \mathbf Z(m_+)$ and $\pmb \theta_-(k)$, $k\in \mathbf Z(m_-)$ such that 
\begin{equation}\label{eq:lattice:difference}
\begin{array}{ll}
\underset{i\in \mathbf Z(M)}{\min} \ \underset{j \in S_i}{\max} \{l(\pmb x|\pmb  \theta(j))\} \\\
=  \underset{k\in \mathbf Z(m_+)}{\max}l(\pmb x|\pmb  \theta_+(k))\} - \underset{k\in \mathbf Z(m_-)}{\max}l(\pmb x| \pmb \theta_-(k))\}, \quad \forall \pmb x \in \mathbb R^n,
\end{array}
\end{equation}
with  $l(\pmb x| \pmb  \theta) = \pmb J^T \pmb x + b$.

The above property indicates that any PWL function can be expressed as a difference of two convex PWL functions, each of which is formulated as the maximum of multiple linear functions \cite{Wang2004General}. On this basis, the proposal of GHH is promoted \cite{Wang2005GHH}, where the basis functions resemble the convex items (equation \eqref{eq:lattice:difference}).  Moreover, in the analysis of GHH, the upper bound on the
nesting number  of  G-CPLR for describing all PWL functions is significantly tightened.

The presented analysis reveals how these shallow PWLNN representations develop, how they are  correlated and  different to each other, and what important properties they possess, in terms of theoretical significance.

\subsection{  Analysis on PWL-DNNs}\label{sec:analysis:deep} 
In PWL-DNNs,  the approximation ability is reconsidered under the framework of deep learning and novel bounds are derived  concerning both network width and depth. Rather than the representation ability  for shallow PWLNNs, theoretical analysis for PWL-DNNs are mainly cast towards the learning process.

\subsubsection{  Approximation Ability Concerning Width and Depth}
{In early studies on  approximation, neural networks containing one hidden layer with sigmoidal activations were favored \cite{cybenko1989approximation}; this universal approximation ability  also holds for PWLNNs \cite{Breiman1993,Lin1995Canonical}.
Universal approximation ability has been proven for two-hidden-layered neural networks\cite{kuurkova1992kolmogorov}, and  an estimation of hidden neurons was determined, rather than assuming an unbounded number of neurons \cite{cybenko1989approximation,hornik1989multilayer}.  With  success of ReLU, such approximation ability was further elucidated and specified for PWL-DNNs \cite{DBLP:journals/nn/Yarotsky17}.}
The  fully connected PWL-DNNs with ReLU  are universal approximators \cite{Lu2017}, and one variant of ResNets has also been proven as  a universal approximator when the depth of the network approaches infinity \cite{Lin2018}. 

It is also of great importance to investigate the relationship between the approximation behavior and their  architectures, including width and  depth. {In  early studies,  the relationship between the approximation error and the number  of neurons in the single hidden layer with sigmoid activations was investigated \cite{Barron1993Universal}.}   Specific Convolutional DNNs (namely  ConvNet,  ConvNet with ReLU  and max pooling) are universal approximators, 
where the lower bound of the number of hidden channels is also provided  \cite{Cohen2016}. 
Further,  by jointly considering   depth and width of layers, the bounds on the number of layers and the number of neurons in hidden layers can be derived simultaneously\cite{DBLP:conf/iclr/AroraBMM18}. From the perspective of input space partitions,  fully connected PWL-DNNs with ReLU can be equivalently transformed into two-layer fully connected DNNs, and  the bounds on the width of the network to ensure the universal approximation ability can be determined\cite{Kumar2019a}. These aforementioned works  utilize the properties of PWL nonlinearity which allows outcomes such as partitioned linear sub-regions, activated linear outputs over layers, PWL finite element spaces, to name a few.

Current analysis on the approximation analysis in deep learning sheds more light on PWL-DNNs mostly with ReLU.  {They are  heavily used in numerical settings, and more amenable to analysis, providing novel perspectives to broaden the existing theoretical understandings on the empirical learning process of DNNs.  More rigorous discussions on the approximation analysis for PWLNNs are presented elsewhere\cite{devore2021neural}.}

\subsubsection{  Expressive Capacity Given with Samples}
The expressive capacity of DNNs commonly refers to the design of networks being able to realize arbitrary functions over a finite subset of the input space.    {In  early studies,  the expressive capacity of neural networks with one hidden layer and the sign activation  were investigated  \cite{Huang1991}, where the focus was  on the  injective functions \cite{Mirchandani1989}. A two-hidden-layer neural network with  sigmoid was proven to be capable of learning any $N$ samples with arbitrary accuracy only if there are at least $2 \sqrt{(p+2) N}$ neurons in the hidden layers, where $p$ is the output dimension \cite{Huang2003}. For PWL-DNNs, relevant  analysis is also conducted.}
In PWL-DNNs with ReLU, the expressive capacity can be guaranteed with the depth  $k$, width $\mathcal  O(N/k)$, and $\mathcal O(N + n)$ weights on a sample of size $N$ in $n$ dimensions \cite{Zhang2017}.
For arbitrary $N$ samples, any pair of which has a specified minimum Euclidean distance, there exists a multi-layer ResNet that can express the samples accurately with only convolutional layers and ReLU activation functions \cite{Hardt2018}. The expressive capability of the multi-layer ConvNet\cite{Nguyen2018}  and fully connected PWL-DNNs with ReLU\cite{Yun2019} is also known. 
Though the above  informative analysis  helps to understand PWL-DNNs and even generic DNNs, the existing results still require tremendous neurons,   particularly when large amounts of data are involved.

\subsubsection{  Analysis Specified with Localized Linearity over Domain Configurations}
In PWL-DNNs, the unique property of localized nonlinearity  makes it possible to quantitatively analyze the  capacity of DNNs regarding domain configurations; a larger number of  linear sub-regions indicates greater flexibility.
The network expressive capacity can be evaluated  by estimating  the maximum number of linear sub-regions of fully connected PWL-DNNs with ReLU \cite{Pascanu2013On}. The basic idea is to use the Zaslavsky's Theorem of hyperplanes arrangement \cite{zaslavsky1975facing} [G], which estimates  the maximal number of regions in $\mathbb R^n$ with an arrangement of $m$ hyperplanes. In PWL-DNNs, where the retained positive neuron output of the linear function in ReLU corresponds to these hyperplanes, this result can then be applied by recursively reusing the PWL fractures from the previous layers. By identifying distinct linear sub-regions,  bounds for  the maximal number of sub-regions can be  derived.
 Various theoretical results and empirical studies on estimating the number of linear sub-regions can be obtained.   For example, specific  neuron allocation in each layer with the lower bound\cite{DBLP:conf/iclr/AroraBMM18} or upper bound\cite{raghu2017expressive}  can  be derived analogously. These bounds can be further improved  with  mild assumptions, and give more accurate evaluation in exploring the capacity of DNNs in this regard \cite{serra2018bounding,hanin2019complexity}.  In deep architectures, fully connected PWL-DNNs can  be extended to ConvNet \cite{DBLP:conf/icml/XiongHY00020}, where ConvNet  with ReLU brings more linear sub-regions than that of  fully connected PWL-DNNs with asymptotically the same number of parameters, input dimension,  and number of layers \cite{DBLP:conf/icml/XiongHY00020}. Such analyses consider the particular measurement of counting linear sub-regions in PWL-DNNs, and cannot be performed in DNNs with other types of nonlinearity, demonstrating the unique merit of PWL nonlinearity and providing a  novel quantitative measurement of showing the power of going deep.

Although DNNs exhibit great flexibility, they have been found  to be vulnerable to adversarial samples, meaning that human-imperceptible perturbation imposed on an image can fool the DNN to make a wrong prediction \cite{goodfellow2014explaining}.  Recent work on robustness certification focuses on whether the prediction  of any input $\pmb x$ is verifiably constant within a neighboring set of $\pmb x$ \cite{katz2017reluplex,DBLP:conf/nips/BunelTTKM18,DBLP:conf/iclr/JiaCWG20}.  Most  robustness certification is conducted with ReLU and heavily relies on the PWL nonlinearity for analysis and algorithm designs, such as the mix-integer linear programming  \cite{tjeng2017evaluating,cheng2017maximum} and the convex outer approximation technique \cite{wong2018provable}, where the local information of linear boundaries and vertices is utilized. Hence, PWL nonlinearity also plays an essential role to advocate the robustness analysis in deep learning.

\section{ Applications}\label{sec:application}
In this section,  representative applications in different fields are introduced to  exemplify the practical values of PWLNNs.

\subsection{  Circuits analysis}
The pioneering work of shallow  PWLNNs, particularly the canonical family, stems from the field of circuit analysis. It was first proposed to use vertices of simplices for the interpolation analysis of nonlinear resistive networks in 1956 \cite{1956stern}. The  PWL technique has since attracted more attention and found successful applications in nonlinear circuit analysis involving uncoupled and monotonically increasing PWL resistors \cite{katzenelson1965algorithm}. 
This work has further been extended with rigorous improvements  in circuit analysis, including more general solutions  of equilibrium points \cite{1971DC,chua1976switching,chien1977piecewise,Yamamura1992An,pastore1993polyhedral,yamamura1998finding}, variously characterized electronic devices \cite{chua1972modeling,meijer1990fast,yamamura1992piecewise,suykens1997family} and circuits with complex dynamics \cite{Chua1986The,billings1987piecewise,sontag1995linear,Mestl1995Periodic,suykens2005cellular}. In recent years, efforts have been made to apply PWL techniques to the circuit systems containing memristors.  PWL window functions are very flexible and have been applied to model different types of ideal memristors \cite{Yu2013A,Mu2015Modeling,Yu2015Aa}. The input-output data are also adopted to construct complex memristor systems from the perspective of approximation, in which  PWLNNs in the form  of SBF  have been successfully applied with promising performance \cite{juntangyuphd}.
Many circuit components, such as resistors and memristors, are PWL-characterized, meaning that PWLNN relating methods can be  naturally applied and good performance can commonly be expected.

\subsection{  Control}
One of the most successful achievements of  PWLNNs is their application in control systems, particularly in dynamic system identification and  MPC. 
General PWLNN models naturally and widely exist in control systems, where hybrid phenomena can be handled \cite{bemporad2000optimization,bemporad2000observability,HEEMELS20011085,bemporad2021piecewise}. 
The systems here are not necessarily continuous, albeit they are PWL-characterized, the discontinuous cases are not discussed in this Primer. 

System identification consists of building a proper mathematical model to  describe the coherent relationship based on the given input-output data from a dynamic control system. Note that the basic idea in system identification is similar to that of general approximation problems and supervised learning. Here, we mainly discuss the PWLNNs applied in typical problems concerning control community, where the dynamic system is in relatively lower dimensions and  the dynamics of  control systems are addressed.  PWLNNs have shown great potentials in various dynamic systems\cite{Hush1998Efficient,2002Nonlinear,wangsun2007}. In fact, many  learning algorithms specified for shallow  PWLNNs initially originated from resolving the  system identification problems in  control  \cite{Xu2009AHH,2010Identification,DBLP:journals/ijon/HuangXW12,2012Continuoushuang,yu2017incremental}. Recently, a novel PWLNN for dynamic system identification was developed to provide an interpretable predictor facilitating variable selection and analysis\cite{EHH}, and  has  been  applied to  traffic flow prediction \cite{ehhtf}.

MPC has long been addressed  as an important topic for complex constrained multi-variable control problems, particularly in industrial processes \cite{pistikopoulos2002line}. 
The controller design in MPC is an open-loop optimization problem. The intensive online computation of the repetitive solutions to the optimization problem is a major obstacle hindering its wider use.  Explicit solutions to linear MPC have been proposed, in which the controller is formulated as a PWL function of state variables \cite{bemporad2000piecewise,bemporad2002model,bemporad2002explicit}. It  has become more prevalent to use PWL-characterized predictive models in MPC, where different PWLNNs have been applied. In one example, the power of PWLNNs in MPC was preliminarily verified by using hinging hyperplanes, but this method neither fully considered the superiority of PWLNN nor exerted a good combination with MPC \cite{chikkula1998dynamically}. The Lattice representation was then utilized to achieve analytical expression of the explicit MPC solution \cite{2009Analytical}. AHH  was also successfully applied to determine the necessary and sufficient conditions for local optimality\cite{Xu2009AHH}. Further, the minimal conjunctive normal expression based on the Lattice was achieved for MPC with the smallest number of parameters\cite{Xu2016Minimal}. It has also been shown that using convex projections of the Lattice representation can be another potential technique for solving explicit MPC \cite{DBLP:conf/cdc/XuW19}. 
Compared to other methods, PWLNNs have relatively simpler identification process and also show advantageous accuracy  with efficiency.  It inspires us that a good utilization of specific geometrical properties of  PWLNNs and a proper combination with  practical settings fundamentally help  approach more extensive applications.

\subsection{  Image processing}
Nowadays, PWL-DNNs can be designed over hundreds of layers to extract informative features for the  learning of  complicated tasks \cite{DBLP:conf/cvpr/HeZRS16}. PWL-DNNs have become one of the most popular choices in deep learning, where the state-of-the-art performance of many popular image datasets, such as ImageNet, has been constantly refreshed along with the proposal of various PWL activations or related techniques \cite{leakyrelu,Goodfellow2013MaxoutN}. In addition to image processing, acoustic and video processing can be improved based on PWL-DNNs \cite{yue2015beyond,purwins2019deep}.

In 2012, a PWL-DNN with ReLU was applied to greatly improve the classification accuracy on approximately 1.2 million images into 1000 classes \cite{krizhevsky2012imagenet}. By introducing ReLU to  DNNs,  the gradient vanishing could be greatly relieved with boosted performances, resulting in faster learning compared to Tanh units and others \cite{krizhevsky2012imagenet}. In fact, faster learning has a great influence on the performance of large models trained on large-scale data. PWL-DNNs have long been dominating the image processing tasks and constantly improving accuracy in various benchmarks \cite{he2015delving,xie2020self}. Practical real-world applications have  received great benefits from the commercial aspects of hardware and algorithmic implementations \cite{DBLP:journals/neco/RawatW17,qiao2017fpga}. PWL-DNNs have made and will continue making remarkable contributions to image processing and far beyond.

\section{    Reproducibility and data deposition}\label{sec:learning}
To apply PWLNNs in data science,  reproducibility and data deposition is an important element. For a specific task with given data, a general workflow to sequentially apply a PWLNN consists of selecting a PWLNN representation model, feeding data into the PWLNN, conducting the learning, evaluating the  performance and tuning until the desired conditions are satisfied (Figure \ref{fig:workflow}). In such a workflow, there are many factors affecting the reproducibility, which can be categorized into data influence and algorithmic influence.

\subsection{  Data influence} 
When a new algorithm or model is proposed and needs to be verified, the baselines of data sources are required. Thus, various carefully designed benchmark datasets are collected from natural environments and have been widely used for applying PWLNNs (including other methods for data) to validate their effectiveness with fair comparisons. When using benchmark datasets, reproducibility can be guaranteed. 
In practical scenarios, particularly involving industrial processes, data can be corrupted with noise, due to environmental influences or measurement errors. Thus, sometimes in the evaluation, the performances against noises need to be considered. In such cases, to ensure reproducibility, manually-corrupted noises by computer programming can be made identical by fixing the computer seed for generating noise in the programming of each run. In this way,   benchmark datasets together with identical noises can be evaluated by computer runs.
In deep learning, there are numerous benchmark datasets, such as UCI repository \cite{Dua:2019}, MNIST\cite{lecun-mnisthandwrittendigit-2010}, SVHN \cite{netzer2011reading}, NORB \cite{lecun2004learning}, CIFAR 10/100 \cite{cifar10}, COCO \cite{lin2014microsoft}, and ImageNet \cite{ILSVRC15}, Visual Genome \cite{krishnavisualgenome} meaning that solid experiments of empirical evaluations can be conducted with guaranteed reproducibility. 

\subsection*{  Algorithmic influence} 
Learning algorithms are  also significant for reproducibility since different sets of parameters can result in a completely different performance even with the same PWLNN structure and identical dataset. Random initialization can lead to different results in learning algorithms, as parameters get iteratively updated to different values. Therefore, the same random seed should be fixed for the initialization of learning algorithms. In particular, the learning algorithms for PWL-DNNs are based on  gradients, the computation of which can differ in concrete implementations, because the gradients are usually computed numerically due to the nondifferentiable PWL nonlinearity. Various computational platforms are constructed, where PWL-DNNs can be easily applied and well learned. Each platform has its own standards; popular platforms include Tensorflow \cite{tensorflow2015-whitepaper}, PyTorch \cite{paszke2019pytorch}, Keras \cite{chollet2015keras}, Caffe \cite{jia2014caffe}, MXNet \cite{chen2015mxnet}, Theano \cite{bergstra+al:2010-scipy}. Once a certain platform is selected, reproducibility can be guaranteed. Data decomposition can also be implemented identically if all randomness has been eliminated by the previously mentioned strategies.

\section{    Limitations and optimizations}\label{sec:discussion}
Despite the developments of PWLNNs achieved so far, there are still many challenges worthy  to be addressed. In shallow  PWLNNs, there are different ways of introducing PWL nonlinearity. For example, each PWLNN representation based on basis functions has distinctively different motivations and explanations for the varied PWL feature mapings. In contrast, the PWL nonlinearity in PWL-DNNs is achieved  by directly adopting some simple PWL functions as   activations. In existing PWL-DNNs, the basic building blocks of the network and  connections between neurons, both within and across layers, are inherited from generic  DNNs. They lack specifications by combining PWL nonlinearity on constructing novel  deep  architectures $\mathcal{N}$. The potential of extending  PWLNNs in the form of AHH  towards novel deep network architectures has been explored\cite{tao2021toward}, but it still remains difficult to tackle  complex tasks involving really deep architectures.

To apply  PWLNNs in data science, various learning algorithms have been proposed. It should be noted that although shallow PWLNNs are only applicable to a limited range of problems in low dimensions and with small scales, each  PWLNN representation model has its own learning algorithm, which is constructed by fully considering the specific characteristics of the model. When it comes to PWL-DNNs, we admit that the  combination of SGD and backpropagation in deep learning has truly realized the powerful flexibility of PWL-DNNs and promoted the applications, but the regular learning algorithms for  PWL-DNNs hold no difference from other generic DNNs. In fact, when it involves an optimization problem with PWL nonlinearity,  local  information  of vertices and  linearity have not been utilized,  and yet it shows to be quite useful and promising in solving the related optimization problems of shallow  PWLNNs for attaining higher accuracy,  efficient computation, and explicitness of learning process. We can rethink developing novel learning algorithms for PWL-DNNs with specifications on different PWL activation functions or specific network architectures, in which full information from PWL nonlinearity should be utilized and more potentials are promising to be further explored.

In the existing literature, theoretical analysis concerning  shallow  PWLNN representations has been discussed vigorously, such as  representation ability,  existence conditions,  and domain configurations. As pointed out in this Primer, strong relations exist between the shallow and deep PWLNNs. However,  only a few of the theoretical conclusions for the existing PWLNN representations  have been used in deep learning. For example, the universal representation ability of  GHH  in analyzing shallow  PWLNNs has been employed to understand  PWL-DNNs with ReLU \cite{DBLP:conf/iclr/AroraBMM18}. PWL-DNNs shall be further investigated by recalling vigorous theories in the previously existing PWLNNs to facilitate further understanding of PWL-DNNs and even generic DNNs.

\section{    Outlook}\label{sec:conclusion}
An implicit and practical requirement for conventional PWL functions is to be compact and interpretable with regards to  domain partitions and locally linear expressions, since complex  partitions and expressions might result in PWL functions that are hard to be understood with unpredictable behaviors. As a result, shallow  PWLNNs usually assume  locally-dominant features among  sub-regions and aim to achieve sufficiently sparse model structures. In contrast, PWL-DNNs abandon such an assumption and directly adopt simple PWL mapping functions to connect neurons in a certain way. This results in much more complicated domain partitions and locally linear expressions. Although numerical results demonstrate the superior performance of PWL-DNNs, there are still many open problems in this field. In order to understand the fundamentals of DNNs and further improve practical applications, certain questions need to be answered. For example, given the data, is there a PWLNN that simultaneously pertains simple partitions and/or locally-dominant features with considerably good performances? Can we find such a PWLNN by explicitly seeking a shallow PWLNN or implicitly regularizing the learning of a  PWL-DNN? What are the differences and relations between PWLNNs and other kinds of NNs that address locally-dominant features \cite{tang2021sparse}?

Despite  the superior performance of PWL-DNNs, the shallow architectures discussed in this Primer show their own merits, including simpler  structures, better interpretability, and the alleviated overfitting, particularly  in  lower dimensions  with smaller-scale problems.  Novel  formulations of shallow-architecture PWLNNs are still worth being  explored. For example, inserting more linear functions into the hinges for flexibility using GHH and utilizing max-min  composites of univariate hinges for interpretability with AHH.
The Lattice representation {has not yet been extended to deep architectures}, and can be a promising alternative to develop novel network architectures in deep learning. The Lattice representation  can be transformed with strong relations to GHH, which has been extended to PWL-DNNs with Maxout.  Though  the boolean theory was discussed to formulate a multi-level Lattice network \cite{2019Multilevel}, the transformation is neither  unique nor irredundant, and its composite of PWL mappings lacks flexibility, leaving  potential for further improvement, such as large-scale explicit MPC and its efficient hardware implementation.

PWL nonlinearity has significantly contributed to the design of deeper  architectures. Although there is an emerging line of empirical and theoretical works on PWL-DNNs, current theoretical analysis of PWL-DNNs is still far from sufficient, particularly for different variants of PWL mapping functions, compared to the existing theoretical understanding of their shallow counterparts, where specific analysis is cast for each PWLNN representation.  In fact, there are many informative conclusions in shallow PWLNN representations that are very helpful for deep learning\cite{Goodfellow2013MaxoutN,DBLP:conf/iclr/AroraBMM18}.
More rigorous theoretical analysis can further benefit the understanding of PWL-DNNs and generic DNNs. Novel network architectures in deep learning are also promising to be inspired, when rethinking the existing systematical  theories and experiences for general PWLNNs.

The learning of shallow PWLNNs is specified in each representation, while PWL-DNNs directly inherit the regular learning from generic DNNs, where PWL-specified techniques are missing. Thus, when developing novel PWL-DNNs,    {learning algorithms for PWL-DNNs should be taken into account by fully utilizing PWL characteristics specified on different types of PWL} mapping functions, so that novel architectures and effective learning algorithms can be mutually  developed and boosted.    In the learning of  DNNs, another challenge  is to study how the parameter learning  can converge to a local
or global optimum in such highly nonconvex  optimization problems. Optimality conditions have been proven for linear DNNs \cite{kawaguchi2016deep,yun2017global}, and nonlinear DNNs mostly with differentiable activations \cite{Nguyen2017,yun2018small}. Current  analytical results rely on simple DNNs  restricted to multiple assumptions. Considering the superior performances of nondifferentiable PWL-DNNs, related analysis of their optimization is worthy of further attention. It is necessary to analyze these optimality conditions  specified on different types of (PWL) activations and more complex network structures for specific objectives.

 In this Primer, we comprehensively introduced  PWLNNs and their developments since 1970s, from tracing to the pioneering PWLNNs in the form of canonical representations, subsequent studies on shallow architectures, and the recent developments  in  deep ones. In the upcoming years, PWLNNs are sure to be of great significance, and vigorous  developments should be expected, particularly in such an era with massive information in-and-out, which keeps benefiting our society.

\section{Glossary}\label{sec:g}

    \textbf{An induced conclusion by the Stone-Weierstrass approximation  theorem}: any continuous function can be approximated by a PWL function to arbitrary accuracy.
    
    \textbf{PWL function}: it is  a function that appears to be linear in sub-regions of the domain but is by essence nonlinear in the whole domain.
    
    \textbf{Canonical piecewise Linear Representation}: it is the pioneering compact expression by which a PWL function is constructed through  a linear combination of multiple absolute-value  basis functions.

    \textbf{Rectified linear units}: it is one of the most popular  activation functions in neural networks and is defined as the positive part of its arguments by $\max\{0, x\}$. 
    
    \textbf{Hinging hyperplanes}: a hinge function consists of two hyperplanes, namely hinging hyperplanes, continuously joining at the so-called hinge, and has greatly contributed to construct flexible  representation models for continuous PWL functions.
    
    \textbf{Backpropagation strategy}: it is  widely used to train feedforward neural networks and works by computing the gradients of weights of each layer in the network and iterating backward layerwisely for efficient calculation.

    \textbf{Stochastic gradient descent}: it is an iterative optimization algorithm, where the actual gradient is approximated or estimated commonly by  a randomly selected subset of  data. 
    
    \textbf{PWL memristors}: other than the resistor, the inductor, and the capacitor, it is considered as the fourth fundamental two-terminal circuit element including a memory of past voltages or currents; those memristors pertain  PWL-characterized dynamics are called PWL memristors.
    
    \textbf{Gradient vanishing problem}: in the iterative updates of training DNNs with gradient-based algorithms, the multiplying of small values of gradients   by backpropogation can lead to a very small value (approaching  zero) in computing the gradients of early layers, which makes the network hard to proceed the training.
    
    \textbf{Least squares method}: it is an approach to approximate the solutions of an  unknown system given with a set of input-output data points
    by minimizing the sum of the squares of the residuals between the observed output data and network's output. 
    
    \textbf{Gauss-Newton algorithm}: it is a modified Newton's method, which computes the second-order derivatives, to  minimize a sum of squared loss in solving non-linear least squares problems.
    
    \textbf{Multivariate adaptive regression splines}: it is a flexible regression model, consisting of  weighted basis functions, which are expressed in terms of the product of truncated power splines $[\pm (x_i - \beta)]_{+}^q$, and its training procedures can be interpreted as a generalized tree searching based on recursive domain partitions.
    
    \textbf{Consistent variation property}: given a continuous PWL function, it is the necessary and sufficient condition on whether such a function can be expressed by a CPLR  model, where the properties of domain partitions and intersections between partitioned sub-regions are discussed; its detailed descriptions are given in the subsequent context.

    
    \textbf{Zaslavsky's Theorem of hyperplanes arrangement}:  the maximal number of regions in $\mathbb R^d$ with an arrangement of $m$ hyperplanes is estimated by $\sum_{j=0}^n \binom{m}{j}$.

\section{Acknowledgements}
This work is jointly supported by ERC Advanced Grant E-DUALITY (787960), KU Leuven Grant CoE PFV/10/002, and Grant  FWO G0A4917N, EU H2020 ICT-48 Network TAILOR (Foundations of Trustworthy AI - Integrating Reasoning, Learning and Optimization), 
Leuven.AI Institute, National Key Research and Development Program under Grant 2021YFB2501200, and Shanghai Municipal
Science and Technology Major Project (2021SHZDZX0102).



\bibliography{pwlreview}

\begin{thebibliography}{100}
\urlstyle{rm}
\expandafter\ifx\csname url\endcsname\relax
  \def\url#1{\texttt{#1}}\fi
\expandafter\ifx\csname urlprefix\endcsname\relax\def\urlprefix{URL }\fi
\expandafter\ifx\csname doiprefix\endcsname\relax\def\doiprefix{DOI: }\fi
\providecommand{\bibinfo}[2]{#2}
\providecommand{\eprint}[2][]{\url{#2}}

\bibitem{leenaerts2013piecewise}
\bibinfo{author}{Leenaerts, D.} \& \bibinfo{author}{Van~Bokhoven, W.~M.}
\newblock \emph{\bibinfo{title}{Piecewise linear modeling and analysis}}
  (\bibinfo{publisher}{Springer Science \& Business Media},
  \bibinfo{year}{2013}).

\bibitem{1999Real}
\bibinfo{author}{Folland, G.~B.}
\newblock \emph{\bibinfo{title}{Real Analysis: Modern Techniques and Their
  Applications}} (\bibinfo{publisher}{Wiley Interscience},
  \bibinfo{year}{1999}).

\bibitem{chien1977solving}
\bibinfo{author}{Chien, M.-J.} \& \bibinfo{author}{Kuh, E.}
\newblock \bibinfo{journal}{\bibinfo{title}{Solving nonlinear resistive
  networks using piecewise-linear analysis and simplicial subdivision}}.
\newblock {\emph{\JournalTitle{IEEE Transactions on Circuits and Systems}}}
  \textbf{\bibinfo{volume}{24}}, \bibinfo{pages}{305--317}
  (\bibinfo{year}{1977}).

\bibitem{chua1988}
\bibinfo{author}{Chua, L.~O.} \& \bibinfo{author}{Deng, A.}
\newblock \bibinfo{journal}{\bibinfo{title}{Canonical piecewise-linear
  representation}}.
\newblock {\emph{\JournalTitle{IEEE Transactions on Circuits and Systems}}}
  \textbf{\bibinfo{volume}{35}}, \bibinfo{pages}{101--111}
  (\bibinfo{year}{1988}).
\newblock \bibinfo{note}{\textbf{The systematical analysis on CPLR is given in
  the paper, including some crucial properties of PWLNNs.}}

\bibitem{chua1977section}
\bibinfo{author}{Chua, L.~O.} \& \bibinfo{author}{Kang, S.}
\newblock \bibinfo{journal}{\bibinfo{title}{Section-wise piecewise-linear
  functions: Canonical representation, properties, and applications}}.
\newblock {\emph{\JournalTitle{Proceedings of the IEEE}}}
  \textbf{\bibinfo{volume}{65}}, \bibinfo{pages}{915--929}
  (\bibinfo{year}{1977}).
\newblock \bibinfo{note}{\textbf{The pioneering compact expression for PWL
  functions is proposed and formally introduced in this paper for circuit
  systems, and then the analytical analysis for PWL functions since becomes
  viable.}}

\bibitem{Nair2010Rectified}
\bibinfo{author}{Nair, V.} \& \bibinfo{author}{Hinton, G.}
\newblock \bibinfo{title}{Rectified linear units improve restricted boltzmann
  machines}.
\newblock In \emph{\bibinfo{booktitle}{Proceedings of the International
  Conference on International Conference on Machine Learning}},
  \bibinfo{pages}{807--814} (\bibinfo{year}{2010}).
\newblock \bibinfo{note}{\textbf{PWL-DNNs start being prevalent and showing
  state-of-the-art performance since this paper, where the most popular ReLU is
  established.}}

\bibitem{Kang1978Chua}
\bibinfo{author}{Kang, S.} \& \bibinfo{author}{Chua, L.~O.}
\newblock \bibinfo{journal}{\bibinfo{title}{A global representation of
  multidimensional piecewise-linear functions with linear partitions}}.
\newblock {\emph{\JournalTitle{IEEE Transactions on Circuits and Systems}}}
  \textbf{\bibinfo{volume}{25}}, \bibinfo{pages}{938--940}
  (\bibinfo{year}{1978}).

\bibitem{1993Canonical}
\bibinfo{author}{Lin, J.~N.} \& \bibinfo{author}{Unbehauen, R.}
\newblock \bibinfo{journal}{\bibinfo{title}{Canonical representation: from
  piecewise-linear function to piecewise-smooth functions}}.
\newblock {\emph{\JournalTitle{IEEE Transactions on Circuits and Systems I:
  Fundamental Theory and Applications}}} \textbf{\bibinfo{volume}{40}},
  \bibinfo{pages}{461--468} (\bibinfo{year}{1993}).

\bibitem{Breiman1993}
\bibinfo{author}{Breiman, L.}
\newblock \bibinfo{journal}{\bibinfo{title}{Hinging hyperplanes for regression,
  classification, and function approximation}}.
\newblock {\emph{\JournalTitle{IEEE Transactions on Information Theory}}}
  \textbf{\bibinfo{volume}{39}}, \bibinfo{pages}{999--1013}
  (\bibinfo{year}{1993}).
\newblock \bibinfo{note}{\textbf{HH representation model and its hinge-finding
  learning algorithm are introduced in this paper. The connection with ReLU in
  PWL-DNNs can be referred.}}

\bibitem{1995Explicitlin}
\bibinfo{author}{Lin, J.~N.} \& \bibinfo{author}{Unbehauen, R.}
\newblock \bibinfo{journal}{\bibinfo{title}{Explicit piecewise-linear models}}.
\newblock {\emph{\JournalTitle{IEEE Transactions on Circuits and Systems I:
  Fundamental Theory and Applications}}} \textbf{\bibinfo{volume}{41}},
  \bibinfo{pages}{931--933} (\bibinfo{year}{1995}).

\bibitem{Terela1999Lattice}
\bibinfo{author}{Tarela, J.} \& \bibinfo{author}{Mart\'{i}nez, M.}
\newblock \bibinfo{journal}{\bibinfo{title}{Region configurations for
  realizability of lattice piecewise-linear models}}.
\newblock {\emph{\JournalTitle{Mathematical and Computer Modelling}}}
  \textbf{\bibinfo{volume}{30}}, \bibinfo{pages}{17--27}
  (\bibinfo{year}{1999}).
\newblock \bibinfo{note}{\textbf{Formal proofs on the universal representation
  ability of the Lattice representation are given and different locally linear
  sub-region realizations are summarized.}}

\bibitem{Julin2003The}
\bibinfo{author}{Juli\'{a}n, P.}
\newblock \bibinfo{journal}{\bibinfo{title}{The complete canonical
  piecewise-linear representation: Functional form for minimal degenerate
  intersections}}.
\newblock {\emph{\JournalTitle{IEEE Transactions on Circuits and Systems I:
  Fundamental Theory and Applications}}} \textbf{\bibinfo{volume}{50}},
  \bibinfo{pages}{387--396} (\bibinfo{year}{2003}).

\bibitem{2005wen}
\bibinfo{author}{Wen, C.}, \bibinfo{author}{Wang, S.}, \bibinfo{author}{Li, F.}
  \& \bibinfo{author}{Khan, M.~J.}
\newblock \bibinfo{journal}{\bibinfo{title}{A compact f-f model of
  high-dimensional piecewise-linear function over a degenerate intersection}}.
\newblock {\emph{\JournalTitle{IEEE Transactions on Circuits and Systems I:
  Regular Papers}}} \textbf{\bibinfo{volume}{52}}, \bibinfo{pages}{815--821}
  (\bibinfo{year}{2005}).

\bibitem{Wang2005GHH}
\bibinfo{author}{Wang, S.} \& \bibinfo{author}{Sun, X.}
\newblock \bibinfo{journal}{\bibinfo{title}{Generalization of hinging
  hyperplanes}}.
\newblock {\emph{\JournalTitle{IEEE Transactions on Information Theory}}}
  \textbf{\bibinfo{volume}{51}}, \bibinfo{pages}{4425--4431}
  (\bibinfo{year}{2005}).
\newblock \bibinfo{note}{\textbf{The idea of inserting multiple linear
  functions to the hinge is given in this paper, where formal proofs are given.
  The connection with Maxout in PWL-DNNs can be referred.}}

\bibitem{DBLP:conf/isnn/SunW05}
\bibinfo{author}{Sun, X.} \& \bibinfo{author}{Wang, S.}
\newblock \bibinfo{journal}{\bibinfo{title}{A special kind of neural networks:
  Continuous piecewise linear functions}}.
\newblock {\emph{\JournalTitle{Lecture Notes in Computer Science}}}
  \textbf{\bibinfo{volume}{3496}}, \bibinfo{pages}{375--379}
  (\bibinfo{year}{2005}).

\bibitem{Xu2009AHH}
\bibinfo{author}{Xu, J.}, \bibinfo{author}{Huang, X.} \& \bibinfo{author}{Wang,
  S.}
\newblock \bibinfo{journal}{\bibinfo{title}{Adaptive hinging hyperplanes and
  its applications in dynamic system identification}}.
\newblock {\emph{\JournalTitle{Automatica}}} \textbf{\bibinfo{volume}{45}},
  \bibinfo{pages}{2325--2332} (\bibinfo{year}{2009}).

\bibitem{yu2017incremental}
\bibinfo{author}{Yu, J.}, \bibinfo{author}{Wang, S.} \& \bibinfo{author}{Li,
  L.}
\newblock \bibinfo{journal}{\bibinfo{title}{Incremental design of simplex basis
  function model for dynamic system identification}}.
\newblock {\emph{\JournalTitle{IEEE Transactions on Neural Networks and
  Learning Systems}}} \textbf{\bibinfo{volume}{29}},
  \bibinfo{pages}{4758--4768} (\bibinfo{year}{2017}).

\bibitem{Chua1985Canonical}
\bibinfo{author}{Chua}, \bibinfo{author}{O., L.} \& \bibinfo{author}{Deng,
  A.~C.}
\newblock \bibinfo{journal}{\bibinfo{title}{Canonical piecewise-linear analysis
  - {P}art {II}: Tracing driving-point and transfer characteristics}}.
\newblock {\emph{\JournalTitle{IEEE Transactions on Circuits and Systems}}}
  \textbf{\bibinfo{volume}{32}}, \bibinfo{pages}{417--444}
  (\bibinfo{year}{1985}).

\bibitem{Wang2004General}
\bibinfo{author}{Wang, S.}
\newblock \bibinfo{journal}{\bibinfo{title}{General constructive
  representations for continuous piecewise-linear functions}}.
\newblock {\emph{\JournalTitle{IEEE Transactions on Circuits and Systems I:
  Regular Papers}}} \textbf{\bibinfo{volume}{51}}, \bibinfo{pages}{1889--1896}
  (\bibinfo{year}{2004}).
\newblock \bibinfo{note}{\textbf{A general constructive method for representing
  an arbitrary PWL function is considered, in which significant differences and
  connections between different representation models are vigorously discussed.
  Many theoretical analysis on PWL-DNNs adopts the Theorems and Lemmas proposed
  in this paper.}}

\bibitem{2010AWangsmooth}
\bibinfo{author}{Wang, S.}, \bibinfo{author}{Huang, X.} \&
  \bibinfo{author}{Yam, Y.}
\newblock \bibinfo{journal}{\bibinfo{title}{A neural network of smooth hinge
  functions}}.
\newblock {\emph{\JournalTitle{IEEE Transactions on Neural Networks}}}
  \textbf{\bibinfo{volume}{21}}, \bibinfo{pages}{1381--1395}
  (\bibinfo{year}{2010}).

\bibitem{2010Stability}
\bibinfo{author}{Xu, J.}, \bibinfo{author}{Huang, X.} \& \bibinfo{author}{Wang,
  S.}
\newblock \bibinfo{title}{Stability analysis of planar continuous piecewise
  linear systems}.
\newblock In \emph{\bibinfo{booktitle}{Proceedings of the American Control
  Conference}}, \bibinfo{pages}{2505--2510} (\bibinfo{year}{2010}).

\bibitem{2011Dynamic}
\bibinfo{author}{Mu, X.}, \bibinfo{author}{Huang, X.} \& \bibinfo{author}{Wang,
  S.}
\newblock \bibinfo{journal}{\bibinfo{title}{Dynamic behavior of
  piecewise-linear approximations}}.
\newblock {\emph{\JournalTitle{Journal of Tsinghua University}}}
  \textbf{\bibinfo{volume}{51}}, \bibinfo{pages}{879--883}
  (\bibinfo{year}{2011}).

\bibitem{2012Exact}
\bibinfo{author}{Huang, X.}, \bibinfo{author}{Xu, J.} \& \bibinfo{author}{Wang,
  S.}
\newblock \bibinfo{journal}{\bibinfo{title}{Exact penalty and optimality
  condition for nonseparable continuous piecewise linear programming}}.
\newblock {\emph{\JournalTitle{Journal of Optimization Theory and
  Applications}}} \textbf{\bibinfo{volume}{155}}, \bibinfo{pages}{145--164}
  (\bibinfo{year}{2012}).

\bibitem{Xu2016Irredundant}
\bibinfo{author}{Xu, J.}, \bibinfo{author}{Boom, T.},
  \bibinfo{author}{Schutter, B.} \& \bibinfo{author}{Wang, S.}
\newblock \bibinfo{journal}{\bibinfo{title}{Irredundant lattice representations
  of continuous piecewise affine functions}}.
\newblock {\emph{\JournalTitle{Automatica}}} \textbf{\bibinfo{volume}{70}},
  \bibinfo{pages}{109--120} (\bibinfo{year}{2016}).

\bibitem{Xu2016Minimal}
\bibinfo{author}{Xu, J.}, \bibinfo{author}{Boom, T.},
  \bibinfo{author}{Schutter, B.} \& \bibinfo{author}{Luo, X.}
\newblock \bibinfo{journal}{\bibinfo{title}{Minimal conjunctive normal
  expression of continuous piecewise affine functions}}.
\newblock {\emph{\JournalTitle{IEEE Transactions on Automatic Control}}}
  \textbf{\bibinfo{volume}{61}}, \bibinfo{pages}{1340--1345}
  (\bibinfo{year}{2016}).

\bibitem{Pucar95smoothhinging}
\bibinfo{author}{Pucar, P.} \& \bibinfo{author}{Millnert, M.}
\newblock \bibinfo{title}{Smooth hinging hyperplanes - an alternative to neural
  nets}.
\newblock In \emph{\bibinfo{booktitle}{Proceedings of the 3rd European Control
  Conference}}, \bibinfo{pages}{1173--1178} (\bibinfo{year}{1995}).

\bibitem{Hush1998Efficient}
\bibinfo{author}{Hush, D.} \& \bibinfo{author}{Horne, B.}
\newblock \bibinfo{journal}{\bibinfo{title}{Efficient algorithms for function
  approximation with piecewise linear sigmoidal networks}}.
\newblock {\emph{\JournalTitle{IEEE Transactions on Neural Networks}}}
  \textbf{\bibinfo{volume}{9}}, \bibinfo{pages}{1129--1141}
  (\bibinfo{year}{1998}).

\bibitem{2002Nonlinear}
\bibinfo{author}{Wang, S.} \& \bibinfo{author}{Narendra, K.~S.}
\newblock \bibinfo{title}{Nonlinear system identification with lattice
  piecewise-linear functions}.
\newblock In \emph{\bibinfo{booktitle}{Proceedings of the American Control
  Conference}}, \bibinfo{pages}{388--393} (\bibinfo{year}{2002}).

\bibitem{wangsun2007}
\bibinfo{author}{Wen, C.}, \bibinfo{author}{Wang, S.}, \bibinfo{author}{Jin,
  X.} \& \bibinfo{author}{Ma, X.}
\newblock \bibinfo{journal}{\bibinfo{title}{Identification of dynamic systems
  using piecewise-affine basis function models}}.
\newblock {\emph{\JournalTitle{Automatica}}} \textbf{\bibinfo{volume}{43}},
  \bibinfo{pages}{1824--1831} (\bibinfo{year}{2007}).

\bibitem{2008Configuration}
\bibinfo{author}{Wang, S.}, \bibinfo{author}{Huang, X.} \&
  \bibinfo{author}{Khan~Junaid, K.~M.}
\newblock \bibinfo{journal}{\bibinfo{title}{Configuration of continuous
  piecewise-linear neural networks}}.
\newblock {\emph{\JournalTitle{IEEE Transactions on Neural Networks}}}
  \textbf{\bibinfo{volume}{19}}, \bibinfo{pages}{1431--45}
  (\bibinfo{year}{2008}).

\bibitem{2010Identification}
\bibinfo{author}{Huang, X.}, \bibinfo{author}{Xu, J.} \& \bibinfo{author}{Wang,
  S.}
\newblock \bibinfo{title}{Identification algorithm for standard continuous
  piecewise linear neural network}.
\newblock In \emph{\bibinfo{booktitle}{Proceedings of the American Control
  Conference}}, \bibinfo{pages}{4431--4936} (\bibinfo{year}{2010}).
\newblock \bibinfo{note}{\textbf{A gradient descent learning algorithm is
  proposed for PWLNNs, where domain partitions and parameter optimizations are
  both elucidated.}}

\bibitem{krizhevsky2012imagenet}
\bibinfo{author}{Krizhevsky, A.}, \bibinfo{author}{Sutskever, I.} \&
  \bibinfo{author}{Hinton, G.~E.}
\newblock \bibinfo{title}{Imagenet classification with deep convolutional
  neural networks}.
\newblock In \emph{\bibinfo{booktitle}{Advances in Neural Information
  Processing Systems}}, \bibinfo{pages}{1097--1105} (\bibinfo{year}{2012}).

\bibitem{DBLP:conf/cvpr/HeZRS16}
\bibinfo{author}{He, K.}, \bibinfo{author}{Zhang, X.}, \bibinfo{author}{Ren,
  S.} \& \bibinfo{author}{Sun, J.}
\newblock \bibinfo{title}{Deep residual learning for image recognition}.
\newblock In \emph{\bibinfo{booktitle}{Proceedings of the {IEEE} Conference on
  Computer Vision and Pattern Recognition}}, \bibinfo{pages}{770--778}
  (\bibinfo{year}{2016}).

\bibitem{DBLP:conf/cvpr/HuangLMW17}
\bibinfo{author}{Huang, G.}, \bibinfo{author}{Liu, Z.},
  \bibinfo{author}{van~der Maaten, L.} \& \bibinfo{author}{Weinberger, K.~Q.}
\newblock \bibinfo{title}{Densely connected convolutional networks}.
\newblock In \emph{\bibinfo{booktitle}{Proceedings of the {IEEE} Conference on
  Computer Vision and Pattern Recognition}}, \bibinfo{pages}{2261--2269}
  (\bibinfo{year}{2017}).

\bibitem{DBLP:conf/iclr/AroraBMM18}
\bibinfo{author}{Arora, R.}, \bibinfo{author}{Basu, A.},
  \bibinfo{author}{Mianjy, P.} \& \bibinfo{author}{Mukherjee, A.}
\newblock \bibinfo{title}{Understanding deep neural networks with rectified
  linear units}.
\newblock In \emph{\bibinfo{booktitle}{Proceedings of the International
  Conference on Learning Representations}} (\bibinfo{year}{2018}).

\bibitem{paszke2019pytorch}
\bibinfo{author}{Paszke, A.} \emph{et~al.}
\newblock \bibinfo{title}{Pytorch: An imperative style, high-performance deep
  learning library}.
\newblock In \emph{\bibinfo{booktitle}{Advances in Neural Information
  Processing Systems}}, \bibinfo{pages}{8024--8035} (\bibinfo{year}{2019}).

\bibitem{Julian1999phd}
\bibinfo{author}{Juli\'{a}n, P.}
\newblock \emph{\bibinfo{title}{A high level canonical piecewise linear
  representation: theory and applications}}.
\newblock Ph.D. thesis, \bibinfo{school}{Universidad Nacional del Sur
  (Argentina)} (\bibinfo{year}{1999}).
\newblock \bibinfo{note}{\textbf{This dissertation gives a very good view on
  the PWL functions and their applications mainly in circuits systems developed
  before the 2000s.}}

\bibitem{Ohnishi1994A}
\bibinfo{author}{Ohnishi, M.} \& \bibinfo{author}{Inaba, N.}
\newblock \bibinfo{journal}{\bibinfo{title}{A singular bifurcation into instant
  chaos in a piecewise-linear circuit}}.
\newblock {\emph{\JournalTitle{IEEE Transactions on Circuits and Systems I:
  Fundamental Theory and Applications}}} \textbf{\bibinfo{volume}{41}},
  \bibinfo{pages}{433--442} (\bibinfo{year}{1994}).

\bibitem{itoh2008memristor}
\bibinfo{author}{Itoh, M.} \& \bibinfo{author}{Chua, L.~O.}
\newblock \bibinfo{journal}{\bibinfo{title}{Memristor oscillators}}.
\newblock {\emph{\JournalTitle{International Journal of Bifurcation and
  Chaos}}} \textbf{\bibinfo{volume}{18}}, \bibinfo{pages}{3183--3206}
  (\bibinfo{year}{2008}).

\bibitem{Bradley1997Clustering}
\bibinfo{author}{Bradley, P.~S.}, \bibinfo{author}{Mangasarian, O.~L.} \&
  \bibinfo{author}{Street, W.~N.}
\newblock \bibinfo{title}{Clustering via concave minimization}.
\newblock In \emph{\bibinfo{booktitle}{Advances in Neural Information
  Processing Systems}}, \bibinfo{pages}{368--374} (\bibinfo{year}{1996}).

\bibitem{Kim2000A}
\bibinfo{author}{Kim, D.} \& \bibinfo{author}{Pardalos, P.~M.}
\newblock \bibinfo{journal}{\bibinfo{title}{A dynamic domain contraction
  algorithm for nonconvex piecewise linear network flow problems}}.
\newblock {\emph{\JournalTitle{Journal of Global Optimization}}}
  \textbf{\bibinfo{volume}{17}}, \bibinfo{pages}{225--234}
  (\bibinfo{year}{2000}).

\bibitem{2010Acost}
\bibinfo{author}{Balakrishnan, A.} \& \bibinfo{author}{Graves, S.~C.}
\newblock \bibinfo{journal}{\bibinfo{title}{A composite algorithm for a
  concave-cost network flow problem}}.
\newblock {\emph{\JournalTitle{Networks}}} \textbf{\bibinfo{volume}{19}},
  \bibinfo{pages}{175--202} (\bibinfo{year}{2010}).

\bibitem{Liu2016Sparse}
\bibinfo{author}{Liu, K.}, \bibinfo{author}{Xu, Z.}, \bibinfo{author}{Xi, X.}
  \& \bibinfo{author}{Wang, S.}
\newblock \bibinfo{journal}{\bibinfo{title}{Sparse signal reconstruction via
  concave continuous piecewise linear programming}}.
\newblock {\emph{\JournalTitle{Digital Signal Processing}}}
  \textbf{\bibinfo{volume}{54}}, \bibinfo{pages}{12--26}
  (\bibinfo{year}{2016}).

\bibitem{2017Aliu}
\bibinfo{author}{Liu, K.}, \bibinfo{author}{Xi, X.}, \bibinfo{author}{Xu, Z.}
  \& \bibinfo{author}{Wang, S.}
\newblock \bibinfo{journal}{\bibinfo{title}{A piecewise linear programming
  algorithm for sparse signal reconstruction}}.
\newblock {\emph{\JournalTitle{Tsinghua Science and Technology}}}
  \textbf{\bibinfo{volume}{22}}, \bibinfo{pages}{29--41}
  (\bibinfo{year}{2017}).

\bibitem{Zhang2006Separable}
\bibinfo{author}{Zhang, H.} \& \bibinfo{author}{Wang, S.}
\newblock \bibinfo{journal}{\bibinfo{title}{Global optimization of separable
  objective functions on convex polyhedra via piecewise-linear approximation}}.
\newblock {\emph{\JournalTitle{Journal of Computational and Applied
  Mathematics}}} \textbf{\bibinfo{volume}{197}}, \bibinfo{pages}{212--217}
  (\bibinfo{year}{2006}).

\bibitem{Zhang2008Nonseparable}
\bibinfo{author}{Zhang, H.} \& \bibinfo{author}{Wang, S.}
\newblock \bibinfo{journal}{\bibinfo{title}{Linearly constrained global
  optimization via piecewise-linear approximation}}.
\newblock {\emph{\JournalTitle{Journal of Computational and Applied
  Mathematics}}} \textbf{\bibinfo{volume}{214}}, \bibinfo{pages}{111--120}
  (\bibinfo{year}{2008}).

\bibitem{Guisewite1991Minimum}
\bibinfo{author}{Guisewite, G.~M.} \& \bibinfo{author}{Pardalos, P.~M.}
\newblock \bibinfo{journal}{\bibinfo{title}{Minimum concave-cost network flow
  problems: Applications, complexity, and algorithms}}.
\newblock {\emph{\JournalTitle{Annals of Operations Research}}}
  \textbf{\bibinfo{volume}{25}}, \bibinfo{pages}{75--99}
  (\bibinfo{year}{1991}).

\bibitem{2001Linearconace}
\bibinfo{author}{Burkard, R.~E.}, \bibinfo{author}{Dollani, H.} \&
  \bibinfo{author}{Thach, P.~T.}
\newblock \bibinfo{journal}{\bibinfo{title}{Linear approximations in a dynamic
  programming approach for the uncapacitated single-source minimum concave cost
  network flow problem in acyclic networks}}.
\newblock {\emph{\JournalTitle{Journal of Global Optimization}}}
  \textbf{\bibinfo{volume}{19}}, \bibinfo{pages}{121--139}
  (\bibinfo{year}{2001}).

\bibitem{DBLP:journals/npl/XiHSW16}
\bibinfo{author}{Xi, X.}, \bibinfo{author}{Huang, X.},
  \bibinfo{author}{Suykens, J. A.~K.} \& \bibinfo{author}{Wang, S.}
\newblock \bibinfo{journal}{\bibinfo{title}{Coordinate descent algorithm for
  ramp loss linear programming support vector machines}}.
\newblock {\emph{\JournalTitle{Neural Processing Letter}}}
  \textbf{\bibinfo{volume}{43}}, \bibinfo{pages}{887--903}
  (\bibinfo{year}{2016}).

\bibitem{xu2015tunneling}
\bibinfo{author}{Xu, Z.}, \bibinfo{author}{Liu, K.}, \bibinfo{author}{Xi, X.}
  \& \bibinfo{author}{Wang, S.}
\newblock \bibinfo{title}{Method of hill tunneling via simplex centroid for
  continuous piecewise linear programming}.
\newblock In \emph{\bibinfo{booktitle}{Proceedings of the IEEE Conference on
  Decision and Control}}, \bibinfo{pages}{6609--6616} (\bibinfo{year}{2015}).

\bibitem{Goodfellow2013MaxoutN}
\bibinfo{author}{Goodfellow, I.}, \bibinfo{author}{Warde-Farley, D.},
  \bibinfo{author}{Mirza, M.}, \bibinfo{author}{Courville, A.} \&
  \bibinfo{author}{Bengio, Y.}
\newblock \bibinfo{title}{Maxout networks}.
\newblock In \emph{\bibinfo{booktitle}{Proceedings of the International
  Conference on Machine Learning}}, \bibinfo{pages}{1319--1327}
  (\bibinfo{year}{2013}).
\newblock \bibinfo{note}{\textbf{A flexible PWL activation function is proposed
  for PWL-DNNs, and ReLU can be regarded as its special case, where analysis on
  the universal approximation ability and the relations to the
  shallow-architectured PWLNNs are given.}}

\bibitem{hopfield1982neural}
\bibinfo{author}{Hopfield, J.~J.}
\newblock \bibinfo{journal}{\bibinfo{title}{Neural networks and physical
  systems with emergent collective computational abilities}}.
\newblock {\emph{\JournalTitle{Proceedings of the national academy of
  sciences}}} \textbf{\bibinfo{volume}{79}}, \bibinfo{pages}{2554--2558}
  (\bibinfo{year}{1982}).

\bibitem{Kahlert1990CS}
\bibinfo{author}{Kahlert, C.} \& \bibinfo{author}{Chua, L.~O.}
\newblock \bibinfo{journal}{\bibinfo{title}{A generalized canonical
  piecewise-linear representation}}.
\newblock {\emph{\JournalTitle{IEEE Transactions on Circuits and Systems}}}
  \textbf{\bibinfo{volume}{37}}, \bibinfo{pages}{373--383}
  (\bibinfo{year}{1990}).

\bibitem{Lin1994CS}
\bibinfo{author}{Lin, J.}, \bibinfo{author}{Xu, H.-Q.} \&
  \bibinfo{author}{Unbehauen, R.}
\newblock \bibinfo{journal}{\bibinfo{title}{A generalization of canonical
  piecewise-linear functions}}.
\newblock {\emph{\JournalTitle{IEEE Transactions on Circuits and Systems I:
  Fundamental Theory and Applications}}} \textbf{\bibinfo{volume}{41}},
  \bibinfo{pages}{345--347} (\bibinfo{year}{1994}).

\bibitem{1998hingefinding}
\bibinfo{author}{Ernst, S.}
\newblock \bibinfo{title}{Hinging hyperplane trees for approximation and
  identification}.
\newblock In \emph{\bibinfo{booktitle}{Proceedings of the IEEE Conference on
  Decision and Control}}, vol.~\bibinfo{volume}{2}, \bibinfo{pages}{1266--1271}
  (\bibinfo{year}{1998}).

\bibitem{Pucar1998}
\bibinfo{author}{Pucar, P.} \& \bibinfo{author}{Sj{\"o}berg, J.}
\newblock \bibinfo{journal}{\bibinfo{title}{On the hinge-finding algorithm for
  hinging hyperplanes}}.
\newblock {\emph{\JournalTitle{IEEE Transactions on Information Theory}}}
  \textbf{\bibinfo{volume}{44}}, \bibinfo{pages}{3310--3319}
  (\bibinfo{year}{1998}).

\bibitem{Ramirez2004Implementation}
\bibinfo{author}{Ramirez, D.~R.}, \bibinfo{author}{Camacho, E.~F.} \&
  \bibinfo{author}{Arahal, M.~R.}
\newblock \bibinfo{journal}{\bibinfo{title}{Implementation of min-max {MPC}
  using hinging hyperplanes. application to a heat exchanger}}.
\newblock {\emph{\JournalTitle{Control Engineering Practice}}}
  \textbf{\bibinfo{volume}{12}}, \bibinfo{pages}{1197--1205}
  (\bibinfo{year}{2004}).

\bibitem{huang2013hinging}
\bibinfo{author}{Huang, X.}, \bibinfo{author}{Matija{\v{s}}, M.} \&
  \bibinfo{author}{Suykens, J.~A.}
\newblock \bibinfo{journal}{\bibinfo{title}{Hinging hyperplanes for time-series
  segmentation}}.
\newblock {\emph{\JournalTitle{IEEE Transactions on Neural Networks and
  Learning Systems}}} \textbf{\bibinfo{volume}{24}},
  \bibinfo{pages}{1279--1291} (\bibinfo{year}{2013}).

\bibitem{DBLP:conf/smc/HuangXW10}
\bibinfo{author}{Huang, X.}, \bibinfo{author}{Xu, J.} \& \bibinfo{author}{Wang,
  S.}
\newblock \bibinfo{title}{Operation optimization for centrifugal chiller plants
  using continuous piecewise linear programming}.
\newblock In \emph{\bibinfo{booktitle}{Proceedings of the {IEEE} International
  Conference on Systems, Man and Cybernetics}}, \bibinfo{pages}{1121--1126}
  (\bibinfo{year}{2010}).

\bibitem{Julian1999CS}
\bibinfo{author}{Juli\'{a}n, P.}, \bibinfo{author}{Desages, A.} \&
  \bibinfo{author}{Agamennoni, O.}
\newblock \bibinfo{journal}{\bibinfo{title}{High-level canonical piecewise
  linear representation using a simplicial partition}}.
\newblock {\emph{\JournalTitle{IEEE Transactions on Circuits and Systems I:
  Fundamental Theory and Applications}}} \textbf{\bibinfo{volume}{46}},
  \bibinfo{pages}{463--480} (\bibinfo{year}{1999}).

\bibitem{padberg2000approximating}
\bibinfo{author}{Padberg, M.}
\newblock \bibinfo{journal}{\bibinfo{title}{Approximating separable nonlinear
  functions via mixed zero-one programs}}.
\newblock {\emph{\JournalTitle{Operations Research Letters}}}
  \textbf{\bibinfo{volume}{27}}, \bibinfo{pages}{1--5} (\bibinfo{year}{2000}).

\bibitem{croxton2003comparison}
\bibinfo{author}{Croxton, K.~L.}, \bibinfo{author}{Gendron, B.} \&
  \bibinfo{author}{Magnanti, T.~L.}
\newblock \bibinfo{journal}{\bibinfo{title}{A comparison of mixed-integer
  programming models for nonconvex piecewise linear cost minimization
  problems}}.
\newblock {\emph{\JournalTitle{Management Science}}}
  \textbf{\bibinfo{volume}{49}}, \bibinfo{pages}{1268--1273}
  (\bibinfo{year}{2003}).

\bibitem{Keha2006branch}
\bibinfo{author}{Keha, A.~B.}, \bibinfo{author}{de~Farias, I.~R.} \&
  \bibinfo{author}{Nemhauser, G.~L.}
\newblock \bibinfo{journal}{\bibinfo{title}{A branch-and-cut algorithm without
  binary variables for nonconvex piecewise linear optimization}}.
\newblock {\emph{\JournalTitle{Operations Research}}}
  \textbf{\bibinfo{volume}{54}}, \bibinfo{pages}{847--858}
  (\bibinfo{year}{2006}).

\bibitem{vielma2010mixed}
\bibinfo{author}{Vielma, J.~P.}, \bibinfo{author}{Ahmed, S.} \&
  \bibinfo{author}{Nemhauser, G.}
\newblock \bibinfo{journal}{\bibinfo{title}{Mixed-integer models for
  nonseparable piecewise-linear optimization: Unifying framework and
  extensions}}.
\newblock {\emph{\JournalTitle{Operations research}}}
  \textbf{\bibinfo{volume}{58}}, \bibinfo{pages}{303--315}
  (\bibinfo{year}{2010}).

\bibitem{lattice1963}
\bibinfo{author}{Wilkinson, R.}
\newblock \bibinfo{journal}{\bibinfo{title}{A method of generating functions of
  several variables using analog diode logic}}.
\newblock {\emph{\JournalTitle{IEEE Transactions on Electronic Computers}}}
  \textbf{\bibinfo{volume}{12}}, \bibinfo{pages}{112--129}
  (\bibinfo{year}{1963}).

\bibitem{Birkhoff1958Lattice}
\bibinfo{author}{Birkhoff} \& \bibinfo{author}{Garrett}.
\newblock \bibinfo{journal}{\bibinfo{title}{Lattice theory}}.
\newblock {\emph{\JournalTitle{Bulletin of the American Mathematical Society}}}
  \textbf{\bibinfo{volume}{64}}, \bibinfo{pages}{50--57}
  (\bibinfo{year}{1958}).

\bibitem{streubel2013representation}
\bibinfo{author}{Streubel, T.}, \bibinfo{author}{Griewank, A.},
  \bibinfo{author}{Radons, M.} \& \bibinfo{author}{Bernt, J.-U.}
\newblock \bibinfo{title}{Representation and analysis of piecewise linear
  functions in abs-normal form}.
\newblock In \emph{\bibinfo{booktitle}{Proceedings of IFIP Conference on System
  Modeling and Optimization}}, \bibinfo{pages}{327--336}
  (\bibinfo{year}{2013}).

\bibitem{DBLP:journals/oms/Griewank13}
\bibinfo{author}{Griewank, A.}
\newblock \bibinfo{journal}{\bibinfo{title}{On stable piecewise linearization
  and generalized algorithmic differentiation}}.
\newblock {\emph{\JournalTitle{Optimization Methods and Software}}}
  \textbf{\bibinfo{volume}{28}}, \bibinfo{pages}{1139--1178}
  (\bibinfo{year}{2013}).

\bibitem{fiege2019algorithm}
\bibinfo{author}{Fiege, S.}, \bibinfo{author}{Walther, A.} \&
  \bibinfo{author}{Griewank, A.}
\newblock \bibinfo{journal}{\bibinfo{title}{An algorithm for nonsmooth
  optimization by successive piecewise linearization}}.
\newblock {\emph{\JournalTitle{Mathematical Programming}}}
  \textbf{\bibinfo{volume}{177}}, \bibinfo{pages}{343--370}
  (\bibinfo{year}{2019}).

\bibitem{griewank2020polyhedral}
\bibinfo{author}{Griewank, A.} \& \bibinfo{author}{Walther, A.}
\newblock \bibinfo{journal}{\bibinfo{title}{Polyhedral {DC} decomposition and
  {DCA} optimization of piecewise linear functions}}.
\newblock {\emph{\JournalTitle{Algorithms}}} \textbf{\bibinfo{volume}{13}},
  \bibinfo{pages}{166} (\bibinfo{year}{2020}).

\bibitem{glorot2011deep}
\bibinfo{author}{Glorot, X.}, \bibinfo{author}{Bordes, A.} \&
  \bibinfo{author}{Bengio, Y.}
\newblock \bibinfo{title}{Deep sparse rectifier neural networks}.
\newblock In \emph{\bibinfo{booktitle}{Proceedings of the International
  Conference on Artificial Intelligence and Statistics}},
  \bibinfo{pages}{315--323} (\bibinfo{year}{2011}).

\bibitem{mcculloch1943logical}
\bibinfo{author}{McCulloch, W.~S.} \& \bibinfo{author}{Pitts, W.}
\newblock \bibinfo{journal}{\bibinfo{title}{A logical calculus of the ideas
  immanent in nervous activity}}.
\newblock {\emph{\JournalTitle{The bulletin of mathematical biophysics}}}
  \textbf{\bibinfo{volume}{5}}, \bibinfo{pages}{115--133}
  (\bibinfo{year}{1943}).

\bibitem{batruni1991multilayer}
\bibinfo{author}{Batruni, R.}
\newblock \bibinfo{journal}{\bibinfo{title}{A multilayer neural network with
  piecewise-linear structure and back-propagation learning}}.
\newblock {\emph{\JournalTitle{IEEE Transactions on Neural Networks}}}
  \textbf{\bibinfo{volume}{2}}, \bibinfo{pages}{395--403}
  (\bibinfo{year}{1991}).

\bibitem{Lin1995Canonical}
\bibinfo{author}{Lin, J.~N.} \& \bibinfo{author}{Unbehauen, R.}
\newblock \bibinfo{journal}{\bibinfo{title}{Canonical piecewise-linear
  networks}}.
\newblock {\emph{\JournalTitle{IEEE Transactions on Neural Networks}}}
  \textbf{\bibinfo{volume}{6}}, \bibinfo{pages}{43--50} (\bibinfo{year}{1995}).
\newblock \bibinfo{note}{\textbf{The network topology for G-CPLR is depicted,
  and the idea of introducing general PWL activation functions for PWL-DNNs is
  also discussed in this paper, yet without numerical evaluations.}}

\bibitem{DBLP:journals/neco/RawatW17}
\bibinfo{author}{Rawat, W.} \& \bibinfo{author}{Wang, Z.}
\newblock \bibinfo{journal}{\bibinfo{title}{Deep convolutional neural networks
  for image classification: {A} comprehensive review}}.
\newblock {\emph{\JournalTitle{Neural Computation}}}
  \textbf{\bibinfo{volume}{29}}, \bibinfo{pages}{2352--2449}
  (\bibinfo{year}{2017}).

\bibitem{leakyrelu}
\bibinfo{author}{Maas, A.}, \bibinfo{author}{Hannun, A.~Y.} \&
  \bibinfo{author}{Ng., A.~Y.}
\newblock \bibinfo{title}{Rectifier nonlinearities improve neural network
  acoustic models}.
\newblock In \emph{\bibinfo{booktitle}{Proceedings of the International
  Conference Machine Learning}}, \bibinfo{pages}{1--8} (\bibinfo{year}{2013}).

\bibitem{he2015delving}
\bibinfo{author}{He, K.}, \bibinfo{author}{Zhang, X.}, \bibinfo{author}{Ren,
  S.} \& \bibinfo{author}{Sun, J.}
\newblock \bibinfo{title}{Delving deep into rectifiers: Surpassing human-level
  performance on imagenet classification}.
\newblock In \emph{\bibinfo{booktitle}{Proceedings of the IEEE International
  Conference on Computer Vision}}, \bibinfo{pages}{1026--1034}
  (\bibinfo{year}{2015}).
\newblock \bibinfo{note}{\textbf{Modifications of optimization strategies on
  the PWL-DNNs and a novel PWL activation function are given in this paper,
  where PWL-DNNs can be delved into fairly deep.}}

\bibitem{DBLP:journals/corr/XuWCL15}
\bibinfo{author}{Xu, B.}, \bibinfo{author}{Wang, N.}, \bibinfo{author}{Chen,
  T.} \& \bibinfo{author}{Li, M.}
\newblock \bibinfo{title}{Empirical evaluation of rectified activations in
  convolutional network}.
\newblock \bibinfo{howpublished}{Preprint at
  \url{https://arxiv.org/abs/1505.00853}} (\bibinfo{year}{2015}).

\bibitem{DBLP:journals/ijon/LiangX21}
\bibinfo{author}{Liang, X.} \& \bibinfo{author}{Xu, J.}
\newblock \bibinfo{journal}{\bibinfo{title}{Biased {R}e{LU} neural networks}}.
\newblock {\emph{\JournalTitle{Neurocomputing}}}
  \textbf{\bibinfo{volume}{423}}, \bibinfo{pages}{71--79}
  (\bibinfo{year}{2021}).

\bibitem{DBLP:conf/icml/ShangSAL16}
\bibinfo{author}{Shang, W.}, \bibinfo{author}{Sohn, K.},
  \bibinfo{author}{Almeida, D.} \& \bibinfo{author}{Lee, H.}
\newblock \bibinfo{title}{Understanding and improving convolutional neural
  networks via concatenated rectified linear units}.
\newblock In \emph{\bibinfo{booktitle}{Proceedings of the International
  Conference on Machine Learning}}, vol.~\bibinfo{volume}{48},
  \bibinfo{pages}{2217--2225} (\bibinfo{year}{2016}).

\bibitem{2015Deep}
\bibinfo{author}{Jin, X.} \emph{et~al.}
\newblock \bibinfo{title}{Deep learning with s-shaped rectified linear
  activation units}.
\newblock In \emph{\bibinfo{booktitle}{Proceedings of the {AAAI} Conference on
  Artificial Intelligence}}, \bibinfo{pages}{1737--1743}
  (\bibinfo{year}{2016}).

\bibitem{DBLP:conf/icpr/QiuXC18}
\bibinfo{author}{Qiu, S.}, \bibinfo{author}{Xu, X.} \& \bibinfo{author}{Cai,
  B.}
\newblock \bibinfo{title}{{FR}e{LU}: Flexible rectified linear units for
  improving convolutional neural networks}.
\newblock In \emph{\bibinfo{booktitle}{Proceedings of the International
  Conference on Pattern Recognition}}, \bibinfo{pages}{1223--1228}
  (\bibinfo{year}{2018}).

\bibitem{DBLP:journals/corr/AgostinelliHSB14}
\bibinfo{author}{Agostinelli, F.}, \bibinfo{author}{Hoffman, M.~D.},
  \bibinfo{author}{Sadowski, P.~J.} \& \bibinfo{author}{Baldi, P.}
\newblock \bibinfo{title}{Learning activation functions to improve deep neural
  networks}.
\newblock In \emph{\bibinfo{booktitle}{Workshop Track Proceedings of the
  International Conference on Learning Representations}}
  (\bibinfo{year}{2015}).

\bibitem{DBLP:conf/dcsmart/BodyanskiyDPS19}
\bibinfo{author}{Bodyanskiy, Y.}, \bibinfo{author}{Deineko, A.},
  \bibinfo{author}{Pliss, I.} \& \bibinfo{author}{Slepanska, V.}
\newblock \bibinfo{title}{Formal neuron based on adaptive parametric rectified
  linear activation function and its learning}.
\newblock In \emph{\bibinfo{booktitle}{Proceedings of the International
  Workshop on Digital Content {\&} Smart Multimedia}}, vol.
  \bibinfo{volume}{2533}, \bibinfo{pages}{14--22} (\bibinfo{year}{2019}).

\bibitem{EHH}
\bibinfo{author}{Xu, J.} \emph{et~al.}
\newblock \bibinfo{journal}{\bibinfo{title}{Efficient hinging hyperplanes
  neural network and its application in nonlinear system identification}}.
\newblock {\emph{\JournalTitle{Automatica}}} \textbf{\bibinfo{volume}{116}},
  \bibinfo{pages}{108906} (\bibinfo{year}{2020}).

\bibitem{suykens1997family}
\bibinfo{author}{Suykens, J.~A.}, \bibinfo{author}{Huang, A.} \&
  \bibinfo{author}{Chua, L.~O.}
\newblock \bibinfo{journal}{\bibinfo{title}{A family of n-scroll attractors
  from a generalized chua's circuit}}.
\newblock {\emph{\JournalTitle{Archiv fur Elektronik und Ubertragungstechnik
  (International Journal of Electronics and Communications)}}}
  \textbf{\bibinfo{volume}{51}}, \bibinfo{pages}{131--137}
  (\bibinfo{year}{1997}).

\bibitem{friedman1991multivariate}
\bibinfo{author}{Friedman, J.~H.} \emph{et~al.}
\newblock \bibinfo{journal}{\bibinfo{title}{Multivariate adaptive regression
  splines}}.
\newblock {\emph{\JournalTitle{The Annals of Statistics}}}
  \textbf{\bibinfo{volume}{19}}, \bibinfo{pages}{1--67} (\bibinfo{year}{1991}).

\bibitem{wang1996induction}
\bibinfo{author}{Wang, Y.} \& \bibinfo{author}{Witten, I.~H.}
\newblock \bibinfo{title}{Induction of model trees for predicting continuous
  classes}.
\newblock In \emph{\bibinfo{booktitle}{Poster papers of the 9th European
  Conference on Machine Learning}} (\bibinfo{year}{1997}).

\bibitem{A2020Learningtao}
\bibinfo{author}{Tao, Q.} \emph{et~al.}
\newblock \bibinfo{journal}{\bibinfo{title}{Learning with continuous piecewise
  linear decision trees}}.
\newblock {\emph{\JournalTitle{Expert Systems with Applications}}}
  \textbf{\bibinfo{volume}{168}}, \bibinfo{pages}{114--214}
  (\bibinfo{year}{2020}).

\bibitem{ferrari2003clustering}
\bibinfo{author}{Ferrari-Trecate, G.}, \bibinfo{author}{Muselli, M.},
  \bibinfo{author}{Liberati, D.} \& \bibinfo{author}{Morari, M.}
\newblock \bibinfo{journal}{\bibinfo{title}{A clustering technique for the
  identification of piecewise affine systems}}.
\newblock {\emph{\JournalTitle{Automatica}}} \textbf{\bibinfo{volume}{39}},
  \bibinfo{pages}{205--217} (\bibinfo{year}{2003}).

\bibitem{nakada2005identification}
\bibinfo{author}{Nakada, H.}, \bibinfo{author}{Takaba, K.} \&
  \bibinfo{author}{Katayama, T.}
\newblock \bibinfo{journal}{\bibinfo{title}{Identification of piecewise affine
  systems based on statistical clustering technique}}.
\newblock {\emph{\JournalTitle{Automatica}}} \textbf{\bibinfo{volume}{41}},
  \bibinfo{pages}{905--913} (\bibinfo{year}{2005}).

\bibitem{bottou1991stochastic}
\bibinfo{author}{Bottou, L.}
\newblock \bibinfo{journal}{\bibinfo{title}{Stochastic gradient learning in
  neural networks}}.
\newblock {\emph{\JournalTitle{Proceedings of Neuro-Nimes}}}
  \textbf{\bibinfo{volume}{91}}, \bibinfo{pages}{12} (\bibinfo{year}{1991}).

\bibitem{jin2021nonconvex}
\bibinfo{author}{Jin, C.}, \bibinfo{author}{Netrapalli, P.},
  \bibinfo{author}{Ge, R.}, \bibinfo{author}{Kakade, S.~M.} \&
  \bibinfo{author}{Jordan, M.~I.}
\newblock \bibinfo{journal}{\bibinfo{title}{On nonconvex optimization for
  machine learning: Gradients, stochasticity, and saddle points}}.
\newblock {\emph{\JournalTitle{Journal of the ACM}}}
  \textbf{\bibinfo{volume}{68}}, \bibinfo{pages}{1--29} (\bibinfo{year}{2021}).

\bibitem{duchi2011adaptive}
\bibinfo{author}{Duchi, J.}, \bibinfo{author}{Hazan, E.} \&
  \bibinfo{author}{Singer, Y.}
\newblock \bibinfo{journal}{\bibinfo{title}{Adaptive subgradient methods for
  online learning and stochastic optimization.}}
\newblock {\emph{\JournalTitle{Journal of Machine Learning Research}}}
  \textbf{\bibinfo{volume}{12}}, \bibinfo{pages}{2121--2159}
  (\bibinfo{year}{2011}).

\bibitem{DBLP:journals/corr/KingmaB14}
\bibinfo{author}{Kingma, D.~P.} \& \bibinfo{author}{Ba, J.}
\newblock \bibinfo{title}{Adam: {A} method for stochastic optimization}.
\newblock In \emph{\bibinfo{booktitle}{Proceedings of the International
  Conference on Learning Representations}} (\bibinfo{year}{2015}).

\bibitem{gupta2018shampoo}
\bibinfo{author}{Gupta, V.}, \bibinfo{author}{Koren, T.} \&
  \bibinfo{author}{Singer, Y.}
\newblock \bibinfo{title}{Shampoo: Preconditioned stochastic tensor
  optimization}.
\newblock In \emph{\bibinfo{booktitle}{Proceedings of the International
  Conference on Machine Learning}}, \bibinfo{pages}{1842--1850}
  (\bibinfo{year}{2018}).

\bibitem{anil2020scalable}
\bibinfo{author}{Anil, R.}, \bibinfo{author}{Gupta, V.},
  \bibinfo{author}{Koren, T.}, \bibinfo{author}{Regan, K.} \&
  \bibinfo{author}{Singer, Y.}
\newblock \bibinfo{title}{Scalable second order optimization for deep
  learning}.
\newblock \bibinfo{howpublished}{Preprint at
  \url{https://arxiv.org/abs/2002.09018}} (\bibinfo{year}{2020}).

\bibitem{srivastava2014dropout}
\bibinfo{author}{Srivastava, N.}, \bibinfo{author}{Hinton, G.},
  \bibinfo{author}{Krizhevsky, A.}, \bibinfo{author}{Sutskever, I.} \&
  \bibinfo{author}{Salakhutdinov, R.}
\newblock \bibinfo{journal}{\bibinfo{title}{Dropout: a simple way to prevent
  neural networks from overfitting}}.
\newblock {\emph{\JournalTitle{Journal of Machine Learning Research}}}
  \textbf{\bibinfo{volume}{15}}, \bibinfo{pages}{1929--1958}
  (\bibinfo{year}{2014}).

\bibitem{ioffe2015batch}
\bibinfo{author}{Ioffe, S.} \& \bibinfo{author}{Szegedy, C.}
\newblock \bibinfo{title}{Batch normalization: Accelerating deep network
  training by reducing internal covariate shift}.
\newblock In \emph{\bibinfo{booktitle}{Proceedings of the International
  Conference on Machine Learning}}, \bibinfo{pages}{448--456}
  (\bibinfo{year}{2015}).

\bibitem{shorten2019survey}
\bibinfo{author}{Shorten, C.} \& \bibinfo{author}{Khoshgoftaar, T.~M.}
\newblock \bibinfo{journal}{\bibinfo{title}{A survey on image data augmentation
  for deep learning}}.
\newblock {\emph{\JournalTitle{Journal of Big Data}}}
  \textbf{\bibinfo{volume}{6}}, \bibinfo{pages}{1--48} (\bibinfo{year}{2019}).

\bibitem{erhan2010does}
\bibinfo{author}{Erhan, D.}, \bibinfo{author}{Courville, A.},
  \bibinfo{author}{Bengio, Y.} \& \bibinfo{author}{Vincent, P.}
\newblock \bibinfo{title}{Why does unsupervised pre-training help deep
  learning?}
\newblock In \emph{\bibinfo{booktitle}{Proceedings of the International
  Conference on Artificial Intelligence and Statistics}},
  \bibinfo{pages}{201--208} (\bibinfo{year}{2010}).

\bibitem{neyshabur2016path}
\bibinfo{author}{Neyshabur, B.}, \bibinfo{author}{Wu, Y.},
  \bibinfo{author}{Salakhutdinov, R.} \& \bibinfo{author}{Srebro, N.}
\newblock \bibinfo{title}{Path-normalized optimization of recurrent neural
  networks with {R}e{LU} activations}.
\newblock In \emph{\bibinfo{booktitle}{Advances in Neural Information
  Processing Systems}}, \bibinfo{pages}{3477--3485} (\bibinfo{year}{2016}).

\bibitem{DBLP:conf/iclr/MengZZCYMYL19}
\bibinfo{author}{Meng, Q.} \emph{et~al.}
\newblock \bibinfo{title}{{G-SGD:} optimizing relu neural networks in its
  positively scale-invariant space}.
\newblock In \emph{\bibinfo{booktitle}{Proceedings of the International
  Conference on Learning Representations}} (\bibinfo{year}{2019}).

\bibitem{wang2019learning}
\bibinfo{author}{Wang, G.}, \bibinfo{author}{Giannakis, G.~B.} \&
  \bibinfo{author}{Chen, J.}
\newblock \bibinfo{journal}{\bibinfo{title}{Learning relu networks on linearly
  separable data: Algorithm, optimality, and generalization}}.
\newblock {\emph{\JournalTitle{IEEE Transactions on Signal Processing}}}
  \textbf{\bibinfo{volume}{67}}, \bibinfo{pages}{2357--2370}
  (\bibinfo{year}{2019}).

\bibitem{tsay2021partition}
\bibinfo{author}{Tsay, C.}, \bibinfo{author}{Kronqvist, J.},
  \bibinfo{author}{Thebelt, A.} \& \bibinfo{author}{Misener, R.}
\newblock \bibinfo{title}{Partition-based formulations for mixed-integer
  optimization of trained {R}e{LU} neural networks}.
\newblock In \emph{\bibinfo{booktitle}{Advances in Neural Information
  Processing Systems}}, vol.~\bibinfo{volume}{34}, \bibinfo{pages}{2993--3003}
  (\bibinfo{year}{2021}).

\bibitem{ergen2021global}
\bibinfo{author}{Ergen, T.} \& \bibinfo{author}{Pilanci, M.}
\newblock \bibinfo{title}{Global optimality beyond two layers: Training deep
  relu networks via convex programs}.
\newblock In \emph{\bibinfo{booktitle}{International Conference on Machine
  Learning}}, \bibinfo{pages}{2993--3003} (\bibinfo{year}{2021}).

\bibitem{wen2016learning}
\bibinfo{author}{Wen, W.}, \bibinfo{author}{Wu, C.}, \bibinfo{author}{Wang,
  Y.}, \bibinfo{author}{Chen, Y.} \& \bibinfo{author}{Li, H.}
\newblock \bibinfo{title}{Learning structured sparsity in deep neural
  networks}.
\newblock In \emph{\bibinfo{booktitle}{Advances in neural information
  processing systems}}, \bibinfo{pages}{2074--2082} (\bibinfo{year}{2016}).

\bibitem{han2015learning}
\bibinfo{author}{Han, S.}, \bibinfo{author}{Pool, J.}, \bibinfo{author}{Tran,
  J.} \& \bibinfo{author}{Dally, W.}
\newblock \bibinfo{title}{Learning both weights and connections for efficient
  neural network}.
\newblock In \emph{\bibinfo{booktitle}{Advances in neural information
  processing systems}}, \bibinfo{pages}{1135--1143} (\bibinfo{year}{2015}).

\bibitem{denton2014exploiting}
\bibinfo{author}{Denton, E.~L.}, \bibinfo{author}{Zaremba, W.},
  \bibinfo{author}{Bruna, J.}, \bibinfo{author}{LeCun, Y.} \&
  \bibinfo{author}{Fergus, R.}
\newblock \bibinfo{title}{Exploiting linear structure within convolutional
  networks for efficient evaluation}.
\newblock \bibinfo{pages}{1269--1277} (\bibinfo{year}{2014}).

\bibitem{DBLP:conf/iclr/FrankleC19}
\bibinfo{author}{Frankle, J.} \& \bibinfo{author}{Carbin, M.}
\newblock \bibinfo{title}{The lottery ticket hypothesis: Finding sparse,
  trainable neural networks}.
\newblock In \emph{\bibinfo{booktitle}{Proceedings of the International
  Conference on Learning Representations}}, \bibinfo{pages}{6336--6347}
  (\bibinfo{year}{2019}).

\bibitem{DBLP:conf/iclr/ZophL17}
\bibinfo{author}{Zoph, B.} \& \bibinfo{author}{Le, Q.~V.}
\newblock \bibinfo{title}{Neural architecture search with reinforcement
  learning}.
\newblock In \emph{\bibinfo{booktitle}{Proceedings of the International
  Conference on Learning Representations}} (\bibinfo{year}{2017}).

\bibitem{tao2018fast}
\bibinfo{author}{Tao, Q.}, \bibinfo{author}{Xu, J.}, \bibinfo{author}{Suykens,
  J. A.~K.} \& \bibinfo{author}{Wang, S.}
\newblock \bibinfo{title}{Fast adaptive hinging hyperplanes}.
\newblock In \emph{\bibinfo{booktitle}{Proceedings of the IEEE Conference on
  Decision and Control}}, \bibinfo{pages}{1482--1487} (\bibinfo{year}{2018}).

\bibitem{cybenko1989approximation}
\bibinfo{author}{Cybenko, G.}
\newblock \bibinfo{journal}{\bibinfo{title}{Approximation by superpositions of
  a sigmoidal function}}.
\newblock {\emph{\JournalTitle{Mathematics of Control, Signals and Systems}}}
  \textbf{\bibinfo{volume}{2}}, \bibinfo{pages}{303--314}
  (\bibinfo{year}{1989}).

\bibitem{kuurkova1992kolmogorov}
\bibinfo{author}{Kurkov{\'{a}}, V.}
\newblock \bibinfo{journal}{\bibinfo{title}{Kolmogorov's theorem and multilayer
  neural networks}}.
\newblock {\emph{\JournalTitle{Neural networks}}} \textbf{\bibinfo{volume}{5}},
  \bibinfo{pages}{501--506} (\bibinfo{year}{1992}).

\bibitem{hornik1989multilayer}
\bibinfo{author}{Hornik, K.}, \bibinfo{author}{Stinchcombe, M.} \&
  \bibinfo{author}{White, H.}
\newblock \bibinfo{journal}{\bibinfo{title}{Multilayer feedforward networks are
  universal approximators}}.
\newblock {\emph{\JournalTitle{Neural networks}}} \textbf{\bibinfo{volume}{2}},
  \bibinfo{pages}{359--366} (\bibinfo{year}{1989}).

\bibitem{DBLP:journals/nn/Yarotsky17}
\bibinfo{author}{Yarotsky, D.}
\newblock \bibinfo{journal}{\bibinfo{title}{Error bounds for approximations
  with deep {R}e{LU} networks}}.
\newblock {\emph{\JournalTitle{Neural Networks}}}
  \textbf{\bibinfo{volume}{94}}, \bibinfo{pages}{103--114}
  (\bibinfo{year}{2017}).

\bibitem{Lu2017}
\bibinfo{author}{Lu, Z.}, \bibinfo{author}{Pu, H.}, \bibinfo{author}{Wang, F.},
  \bibinfo{author}{Hu, Z.} \& \bibinfo{author}{Wang, L.}
\newblock \bibinfo{title}{The expressive power of neural networks: {A} view
  from the width}.
\newblock In \emph{\bibinfo{booktitle}{Advances in Neural Information
  Processing Systems}}, \bibinfo{pages}{6231--6239} (\bibinfo{year}{2017}).

\bibitem{Lin2018}
\bibinfo{author}{Lin, H.} \& \bibinfo{author}{Jegelka, S.}
\newblock \bibinfo{title}{{ResNet} with one-neuron hidden layers is a universal
  approximator}.
\newblock In \emph{\bibinfo{booktitle}{Advances in Neural Information
  Processing Systems}}, vol.~\bibinfo{volume}{31}, \bibinfo{pages}{1--10}
  (\bibinfo{year}{2018}).

\bibitem{Barron1993Universal}
\bibinfo{author}{Barron, A.~R.}
\newblock \bibinfo{journal}{\bibinfo{title}{Universal approximation bounds for
  superpositions of a sigmoidal function}}.
\newblock {\emph{\JournalTitle{IEEE Transactions on Information Theory}}}
  \textbf{\bibinfo{volume}{39}}, \bibinfo{pages}{930--945}
  (\bibinfo{year}{1993}).

\bibitem{Cohen2016}
\bibinfo{author}{Cohen, N.} \& \bibinfo{author}{Shashua, A.}
\newblock \bibinfo{title}{Convolutional rectifier networks as generalized
  tensor decompositions}.
\newblock In \emph{\bibinfo{booktitle}{Proceedings of the International
  Conference on International Conference on Machine Learning}},
  \bibinfo{pages}{955--963} (\bibinfo{year}{2016}).

\bibitem{Kumar2019a}
\bibinfo{author}{Kumar, A.}, \bibinfo{author}{Serra, T.} \&
  \bibinfo{author}{Ramalingam, S.}
\newblock \bibinfo{title}{Equivalent and approximate transformations of deep
  neural networks}.
\newblock \bibinfo{howpublished}{Preprint at
  \url{http://arxiv.org/abs/1905.11428}} (\bibinfo{year}{2019}).

\bibitem{devore2021neural}
\bibinfo{author}{DeVore, R.}, \bibinfo{author}{Hanin, B.} \&
  \bibinfo{author}{Petrova, G.}
\newblock \bibinfo{journal}{\bibinfo{title}{Neural network approximation}}.
\newblock {\emph{\JournalTitle{Acta Numerica}}} \textbf{\bibinfo{volume}{30}},
  \bibinfo{pages}{327--444} (\bibinfo{year}{2021}).
\newblock \bibinfo{note}{\textbf{The approximation properties of NNs are
  described as they are presently understood and their performance with other
  methods of approximation is also discussed, where ReLU is centered in the
  analysis involving univariate and multivariate forms with both shallow and
  deep architectures.}}

\bibitem{Huang1991}
\bibinfo{author}{Huang, S.-C.} \& \bibinfo{author}{Huang, Y.-F.}
\newblock \bibinfo{journal}{\bibinfo{title}{Bounds on the number of hidden
  neurons in multilayer perceptrons}}.
\newblock {\emph{\JournalTitle{IEEE Transactions on Neural Networks}}}
  \textbf{\bibinfo{volume}{2}}, \bibinfo{pages}{47--55} (\bibinfo{year}{1991}).

\bibitem{Mirchandani1989}
\bibinfo{author}{Mirchandani, G.} \& \bibinfo{author}{Cao, W.}
\newblock \bibinfo{journal}{\bibinfo{title}{On hidden nodes for neural nets}}.
\newblock {\emph{\JournalTitle{IEEE Transactions on Circuits and Systems}}}
  \textbf{\bibinfo{volume}{36}}, \bibinfo{pages}{661--664}
  (\bibinfo{year}{1989}).

\bibitem{Huang2003}
\bibinfo{author}{Huang, G.-B.}
\newblock \bibinfo{journal}{\bibinfo{title}{Learning capability and storage
  capacity of two-hidden-layer feedforward networks}}.
\newblock {\emph{\JournalTitle{IEEE Transactions on Neural Networks}}}
  \textbf{\bibinfo{volume}{14}}, \bibinfo{pages}{274--281}
  (\bibinfo{year}{2003}).

\bibitem{Zhang2017}
\bibinfo{author}{Zhang, C.}, \bibinfo{author}{Bengio, S.},
  \bibinfo{author}{Hardt, M.}, \bibinfo{author}{Recht, B.} \&
  \bibinfo{author}{Vinyals, O.}
\newblock \bibinfo{title}{Understanding deep learning requires rethinking
  generalization}.
\newblock In \emph{\bibinfo{booktitle}{Proceedings of the International
  Conference on Learning Representations}}, \bibinfo{pages}{1--15}
  (\bibinfo{year}{2017}).

\bibitem{Hardt2018}
\bibinfo{author}{Hardt, M.} \& \bibinfo{author}{Ma, T.}
\newblock \bibinfo{title}{Identity matters in deep learning}.
\newblock \bibinfo{howpublished}{Preprint at
  \url{https://arxiv.org/abs/1611.04231}} (\bibinfo{year}{2016}).

\bibitem{Nguyen2018}
\bibinfo{author}{Nguyen, Q.} \& \bibinfo{author}{Hein, M.}
\newblock \bibinfo{title}{Optimization landscape and expressivity of deep
  {CNNs}}.
\newblock In \emph{\bibinfo{booktitle}{Proceedings of the International
  Conference on Machine Learning}}, vol.~\bibinfo{volume}{80},
  \bibinfo{pages}{3730--3739} (\bibinfo{year}{2018}).

\bibitem{Yun2019}
\bibinfo{author}{Yun, C.}, \bibinfo{author}{Sra, S.} \&
  \bibinfo{author}{Jadbabaie, A.}
\newblock \bibinfo{title}{Generalization bounds and consistency for latent
  structural probit and ramp loss.}
\newblock In \emph{\bibinfo{booktitle}{Advances in Neural Information
  Processing Systems}}, vol.~\bibinfo{volume}{32} (\bibinfo{year}{2019}).

\bibitem{Pascanu2013On}
\bibinfo{author}{Pascanu, R.}, \bibinfo{author}{Montufar, G.} \&
  \bibinfo{author}{Bengio, Y.}
\newblock \bibinfo{title}{On the number of response regions of deep feed
  forward networks with piece-wise linear activations}.
\newblock \bibinfo{howpublished}{Preprint at
  \url{https://arxiv.org/abs/1312.6098}} (\bibinfo{year}{2013}).

\bibitem{zaslavsky1975facing}
\bibinfo{author}{Zaslavsky, T.}
\newblock \emph{\bibinfo{title}{Facing up to arrangements: Face-count formulas
  for partitions of space by hyperplanes: Face-count formulas for partitions of
  space by hyperplanes}}, vol. \bibinfo{volume}{154}
  (\bibinfo{publisher}{American Mathematical Sociesty}, \bibinfo{year}{1975}).

\bibitem{raghu2017expressive}
\bibinfo{author}{Raghu, M.}, \bibinfo{author}{Poole, B.},
  \bibinfo{author}{Kleinberg, J.}, \bibinfo{author}{Ganguli, S.} \&
  \bibinfo{author}{Sohl-Dickstein, J.}
\newblock \bibinfo{title}{On the expressive power of deep neural networks}.
\newblock In \emph{\bibinfo{booktitle}{Proceedings of the International
  Conference on Machine Learning}}, \bibinfo{pages}{2847--2854}
  (\bibinfo{year}{2017}).

\bibitem{serra2018bounding}
\bibinfo{author}{Serra, T.}, \bibinfo{author}{Tjandraatmadja, C.} \&
  \bibinfo{author}{Ramalingam, S.}
\newblock \bibinfo{title}{Bounding and counting linear regions of deep neural
  networks}.
\newblock In \emph{\bibinfo{booktitle}{Proceedings of the International
  Conference on Machine Learning}}, \bibinfo{pages}{4558--4566}
  (\bibinfo{year}{2018}).

\bibitem{hanin2019complexity}
\bibinfo{author}{Hanin, B.} \& \bibinfo{author}{Rolnick, D.}
\newblock \bibinfo{title}{Complexity of linear regions in deep networks}.
\newblock In \emph{\bibinfo{booktitle}{Proceedings of the International
  Conference on Machine Learning}}, \bibinfo{pages}{2596--2604}
  (\bibinfo{year}{2019}).

\bibitem{DBLP:conf/icml/XiongHY00020}
\bibinfo{author}{Xiong, H.} \emph{et~al.}
\newblock \bibinfo{title}{On the number of linear regions of convolutional
  neural networks}.
\newblock In \emph{\bibinfo{booktitle}{Proceedings of the International
  Conference on Machine Learning}}, vol. \bibinfo{volume}{119},
  \bibinfo{pages}{10514--10523} (\bibinfo{year}{2020}).

\bibitem{goodfellow2014explaining}
\bibinfo{author}{Goodfellow, I.~J.}, \bibinfo{author}{Shlens, J.} \&
  \bibinfo{author}{Szegedy, C.}
\newblock \bibinfo{title}{Explaining and harnessing adversarial examples}.
\newblock In \emph{\bibinfo{booktitle}{Proceedings of the International
  Conference on Learning Representations}} (\bibinfo{year}{2015}).

\bibitem{katz2017reluplex}
\bibinfo{author}{Katz, G.}, \bibinfo{author}{Barrett, C.},
  \bibinfo{author}{Dill, D.~L.}, \bibinfo{author}{Julian, K.} \&
  \bibinfo{author}{Kochenderfer, M.~J.}
\newblock \bibinfo{title}{{R}e{LU}plex: {A}n efficient {SMT} solver for
  verifying deep neural networks}.
\newblock In \emph{\bibinfo{booktitle}{Proceedings of the International
  Conference on Computer Aided Verification}}, \bibinfo{pages}{97--117}
  (\bibinfo{year}{2017}).

\bibitem{DBLP:conf/nips/BunelTTKM18}
\bibinfo{author}{Bunel, R.}, \bibinfo{author}{Turkaslan, I.},
  \bibinfo{author}{Torr, P. H.~S.}, \bibinfo{author}{Kohli, P.} \&
  \bibinfo{author}{Mudigonda, P.~K.}
\newblock \bibinfo{title}{A unified view of piecewise linear neural network
  verification}.
\newblock In \emph{\bibinfo{booktitle}{Advances in Neural Information
  Processing Systems}}, \bibinfo{pages}{4795--4804} (\bibinfo{year}{2018}).

\bibitem{DBLP:conf/iclr/JiaCWG20}
\bibinfo{author}{Jia, J.}, \bibinfo{author}{Cao, X.}, \bibinfo{author}{Wang,
  B.} \& \bibinfo{author}{Gong, N.~Z.}
\newblock \bibinfo{title}{Certified robustness for top-k predictions against
  adversarial perturbations via randomized smoothing}.
\newblock In \emph{\bibinfo{booktitle}{Proceedings of the International
  Conference on Learning Representations}} (\bibinfo{year}{2020}).

\bibitem{tjeng2017evaluating}
\bibinfo{author}{Tjeng, V.}, \bibinfo{author}{Xiao, K.~Y.} \&
  \bibinfo{author}{Tedrake, R.}
\newblock \bibinfo{title}{Evaluating robustness of neural networks with mixed
  integer programming}.
\newblock In \emph{\bibinfo{booktitle}{Proceedings of the International
  Conference on Learning}} (\bibinfo{year}{2019}).

\bibitem{cheng2017maximum}
\bibinfo{author}{Cheng, C.-H.}, \bibinfo{author}{N{\"u}hrenberg, G.} \&
  \bibinfo{author}{Ruess, H.}
\newblock \bibinfo{title}{Maximum resilience of artificial neural networks}.
\newblock In \emph{\bibinfo{booktitle}{International Symposium on Automated
  Technology for Verification and Analysis}}, \bibinfo{pages}{251--268}
  (\bibinfo{year}{2017}).

\bibitem{wong2018provable}
\bibinfo{author}{Wong, E.} \& \bibinfo{author}{Kolter, Z.}
\newblock \bibinfo{title}{Provable defenses against adversarial examples via
  the convex outer adversarial polytope}.
\newblock In \emph{\bibinfo{booktitle}{Proceedings of the International
  Conference on Machine Learning}}, \bibinfo{pages}{5286--5295}
  (\bibinfo{year}{2018}).

\bibitem{1956stern}
\bibinfo{author}{Stern, T.~E.}
\newblock \emph{\bibinfo{title}{Piecewise-linear Network Theory}}
  (\bibinfo{publisher}{MIT Tech. Rep.}, \bibinfo{year}{1956}).

\bibitem{katzenelson1965algorithm}
\bibinfo{author}{Katzenelson, J.}
\newblock \bibinfo{journal}{\bibinfo{title}{An algorithm for solving nonlinear
  resistor networks}}.
\newblock {\emph{\JournalTitle{The Bell System Technical Journal}}}
  \textbf{\bibinfo{volume}{44}}, \bibinfo{pages}{1605--1620}
  (\bibinfo{year}{1965}).

\bibitem{1971DC}
\bibinfo{author}{Ohtsuki, T.} \& \bibinfo{author}{Yoshida, N.}
\newblock \bibinfo{journal}{\bibinfo{title}{Dc analysis of nonlinear networks
  based on generalized piecewise-linear characterization}}.
\newblock {\emph{\JournalTitle{IEEE Transactions on Circuit Theory}}}
  \textbf{\bibinfo{volume}{CT-18}}, \bibinfo{pages}{146--152}
  (\bibinfo{year}{1971}).

\bibitem{chua1976switching}
\bibinfo{author}{Chua, L.~O.} \& \bibinfo{author}{Ushida, A.}
\newblock \bibinfo{journal}{\bibinfo{title}{A switching-parameter algorithm for
  finding multiple solutions of nonlinear resistive circuits}}.
\newblock {\emph{\JournalTitle{International Journal of Circuit Theory and
  Applications}}} \textbf{\bibinfo{volume}{4}}, \bibinfo{pages}{215--239}
  (\bibinfo{year}{1976}).

\bibitem{chien1977piecewise}
\bibinfo{author}{Chien, M.-J.}
\newblock \bibinfo{journal}{\bibinfo{title}{Piecewise-linear theory and
  computation of solutions of homeomorphic resistive networks}}.
\newblock {\emph{\JournalTitle{IEEE Transactions on Circuits and Systems}}}
  \textbf{\bibinfo{volume}{24}}, \bibinfo{pages}{118--127}
  (\bibinfo{year}{1977}).

\bibitem{Yamamura1992An}
\bibinfo{author}{Yamamura, K.} \& \bibinfo{author}{Ochiai, M.}
\newblock \bibinfo{journal}{\bibinfo{title}{An efficient algorithm for finding
  all solutions of piecewise-linear resistive circuits}}.
\newblock {\emph{\JournalTitle{IEEE Transactions on Circuits and Systems}}}
  \textbf{\bibinfo{volume}{39}}, \bibinfo{pages}{P.213--221}
  (\bibinfo{year}{1992}).

\bibitem{pastore1993polyhedral}
\bibinfo{author}{Pastore, S.} \& \bibinfo{author}{Premoli, A.}
\newblock \bibinfo{journal}{\bibinfo{title}{Polyhedral elements: A new
  algorithm for capturing all the equilibrium points of piecewise-linear
  circuits}}.
\newblock {\emph{\JournalTitle{IEEE Transactions on Circuits and Systems I:
  Fundamental Theory and Applications}}} \textbf{\bibinfo{volume}{40}},
  \bibinfo{pages}{124--132} (\bibinfo{year}{1993}).

\bibitem{yamamura1998finding}
\bibinfo{author}{Yamamura, K.} \& \bibinfo{author}{Ohshima, T.}
\newblock \bibinfo{journal}{\bibinfo{title}{Finding all solutions of
  piecewise-linear resistive circuits using linear programming}}.
\newblock {\emph{\JournalTitle{IEEE Transactions on Circuits and Systems I:
  Fundamental Theory and Applications}}} \textbf{\bibinfo{volume}{45}},
  \bibinfo{pages}{434--445} (\bibinfo{year}{1998}).

\bibitem{chua1972modeling}
\bibinfo{author}{Chua, L.~O.}
\newblock \bibinfo{journal}{\bibinfo{title}{Modeling of three terminal devices:
  A black box approach}}.
\newblock {\emph{\JournalTitle{IEEE Transactions on Circuit Theory}}}
  \textbf{\bibinfo{volume}{19}}, \bibinfo{pages}{555--562}
  (\bibinfo{year}{1972}).

\bibitem{meijer1990fast}
\bibinfo{author}{Meijer, P.~B.}
\newblock \bibinfo{journal}{\bibinfo{title}{Fast and smooth highly nonlinear
  multidimensional table models for device modeling}}.
\newblock {\emph{\JournalTitle{IEEE Transactions on Circuits and Systems}}}
  \textbf{\bibinfo{volume}{37}}, \bibinfo{pages}{335--346}
  (\bibinfo{year}{1990}).

\bibitem{yamamura1992piecewise}
\bibinfo{author}{Yamamura, K.}
\newblock \bibinfo{journal}{\bibinfo{title}{On piecewise-linear approximation
  of nonlinear mappings containing gummel-poon models or schichman-hodges
  models}}.
\newblock {\emph{\JournalTitle{IEEE Transactions on Circuits and Systems I:
  Fundamental Theory and Applications}}} \textbf{\bibinfo{volume}{39}},
  \bibinfo{pages}{694--697} (\bibinfo{year}{1992}).

\bibitem{Chua1986The}
\bibinfo{author}{Chua, L.~O.}, \bibinfo{author}{Komuro, M.} \&
  \bibinfo{author}{Matsumoto, T.}
\newblock \bibinfo{journal}{\bibinfo{title}{The double scroll family}}.
\newblock {\emph{\JournalTitle{IEEE Transactions on Circuits and Systems}}}
  \textbf{\bibinfo{volume}{33}}, \bibinfo{pages}{1072--1118}
  (\bibinfo{year}{1986}).

\bibitem{billings1987piecewise}
\bibinfo{author}{Billings, S.} \& \bibinfo{author}{Voon, W.}
\newblock \bibinfo{journal}{\bibinfo{title}{Piecewise linear identification of
  non-linear systems}}.
\newblock {\emph{\JournalTitle{International Journal of Control}}}
  \textbf{\bibinfo{volume}{46}}, \bibinfo{pages}{215--235}
  (\bibinfo{year}{1987}).

\bibitem{sontag1995linear}
\bibinfo{author}{Sontag, E.}
\newblock \bibinfo{title}{From linear to nonlinear: some complexity
  comparisons}.
\newblock In \emph{\bibinfo{booktitle}{Proceedings of the IEEE Conference on
  Decision and Control}}, vol.~\bibinfo{volume}{3}, \bibinfo{pages}{2916--2920}
  (\bibinfo{year}{1995}).

\bibitem{Mestl1995Periodic}
\bibinfo{author}{Mestl, T.}, \bibinfo{author}{Plahte, E.} \&
  \bibinfo{author}{Omholt, S.~W.}
\newblock \bibinfo{journal}{\bibinfo{title}{Periodic solutions in systems of
  piecewise- linear differential equations}}.
\newblock {\emph{\JournalTitle{Dynamics \& Stability of Systems}}}
  \textbf{\bibinfo{volume}{10}}, \bibinfo{pages}{179--193}
  (\bibinfo{year}{1995}).

\bibitem{suykens2005cellular}
\bibinfo{author}{Yalcin, M.}, \bibinfo{author}{Suykens, J.~A.} \&
  \bibinfo{author}{Vandewalle, J.}
\newblock \emph{\bibinfo{title}{Cellular neural networks, multi-scroll chaos
  and synchronization}}, vol.~\bibinfo{volume}{50} (\bibinfo{publisher}{World
  Scientific}, \bibinfo{year}{2005}).

\bibitem{Yu2013A}
\bibinfo{author}{Yu, J.}, \bibinfo{author}{Mu, X.}, \bibinfo{author}{Xi, X.} \&
  \bibinfo{author}{Wang, S.}
\newblock \bibinfo{journal}{\bibinfo{title}{A memristor model with piecewise
  window function}}.
\newblock {\emph{\JournalTitle{Radioengineering}}}
  \textbf{\bibinfo{volume}{22}}, \bibinfo{pages}{969--974}
  (\bibinfo{year}{2013}).

\bibitem{Mu2015Modeling}
\bibinfo{author}{Mu, X.}, \bibinfo{author}{Yu, J.} \& \bibinfo{author}{Wang,
  S.}
\newblock \bibinfo{journal}{\bibinfo{title}{Modeling the memristor with
  piecewise linear function}}.
\newblock {\emph{\JournalTitle{International Journal of Numerical Modelling
  Electronic Networks Devices \& Fields}}} \textbf{\bibinfo{volume}{28}},
  \bibinfo{pages}{96--106} (\bibinfo{year}{2015}).

\bibitem{Yu2015Aa}
\bibinfo{author}{Yu, Y., Juntang~Li}, \bibinfo{author}{Mu, X.},
  \bibinfo{author}{Zhang, J.}, \bibinfo{author}{Miao, X.} \&
  \bibinfo{author}{Wang, S.}
\newblock \bibinfo{journal}{\bibinfo{title}{Modeling the {A}gin{S}b{T}e
  memristor}}.
\newblock {\emph{\JournalTitle{Radioengineering}}}
  \textbf{\bibinfo{volume}{24}}, \bibinfo{pages}{808--813}
  (\bibinfo{year}{2015}).

\bibitem{juntangyuphd}
\bibinfo{author}{Yu, J.}
\newblock \emph{\bibinfo{title}{Memristor model with window function and its
  applications}}.
\newblock Ph.D. thesis, \bibinfo{school}{Tsinghua University}
  (\bibinfo{year}{2016}).

\bibitem{bemporad2000optimization}
\bibinfo{author}{Bemporad, A.}, \bibinfo{author}{Torrisi, F.~D.} \&
  \bibinfo{author}{Morari, M.}
\newblock \bibinfo{title}{Optimization-based verification and stability
  characterization of piecewise affine and hybrid systems}.
\newblock In \emph{\bibinfo{booktitle}{International Workshop on Hybrid
  Systems: Computation and Control}}, \bibinfo{pages}{45--58}
  (\bibinfo{year}{2000}).

\bibitem{bemporad2000observability}
\bibinfo{author}{Bemporad, A.}, \bibinfo{author}{Ferrari-Trecate, G.} \&
  \bibinfo{author}{Morari, M.}
\newblock \bibinfo{journal}{\bibinfo{title}{Observability and controllability
  of piecewise affine and hybrid systems}}.
\newblock {\emph{\JournalTitle{IEEE transactions on automatic control}}}
  \textbf{\bibinfo{volume}{45}}, \bibinfo{pages}{1864--1876}
  (\bibinfo{year}{2000}).

\bibitem{HEEMELS20011085}
\bibinfo{author}{Heemels, W.}, \bibinfo{author}{{De Schutter}, B.} \&
  \bibinfo{author}{Bemporad, A.}
\newblock \bibinfo{journal}{\bibinfo{title}{Equivalence of hybrid dynamical
  models}}.
\newblock {\emph{\JournalTitle{Automatica}}} \textbf{\bibinfo{volume}{37}},
  \bibinfo{pages}{1085--1091} (\bibinfo{year}{2001}).

\bibitem{bemporad2021piecewise}
\bibinfo{author}{Bemporad, A.}
\newblock \bibinfo{title}{Piecewise linear regression and classification}.
\newblock \bibinfo{howpublished}{Preprint at
  \url{https://arxiv.org/abs/2103.06189}} (\bibinfo{year}{2021}).

\bibitem{DBLP:journals/ijon/HuangXW12}
\bibinfo{author}{Huang, X.}, \bibinfo{author}{Xu, J.} \& \bibinfo{author}{Wang,
  S.}
\newblock \bibinfo{journal}{\bibinfo{title}{Nonlinear system identification
  with continuous piecewise linear neural network}}.
\newblock {\emph{\JournalTitle{Neurocomputing}}} \textbf{\bibinfo{volume}{77}},
  \bibinfo{pages}{167--177} (\bibinfo{year}{2012}).

\bibitem{2012Continuoushuang}
\bibinfo{author}{Huang, X.}, \bibinfo{author}{Mu, X.} \& \bibinfo{author}{Wang,
  S.}
\newblock \bibinfo{title}{Continuous piecewise linear identification with
  moderate number of subregions}.
\newblock In \emph{\bibinfo{booktitle}{the 16th {IFAC} Symposium on System
  Identification}}, \bibinfo{pages}{535--540} (\bibinfo{year}{2012}).

\bibitem{ehhtf}
\bibinfo{author}{Tao, Q.} \emph{et~al.}
\newblock \bibinfo{journal}{\bibinfo{title}{Short-term traffic flow prediction
  based on the efficient hinging hyperplanes neural network}}.
\newblock {\emph{\JournalTitle{IEEE Transactions on Intelligent Transportation
  Systems}}} \bibinfo{pages}{1--13} (\bibinfo{year}{2022}).

\bibitem{pistikopoulos2002line}
\bibinfo{author}{Pistikopoulos, E.~N.}, \bibinfo{author}{Dua, V.},
  \bibinfo{author}{Bozinis, N.~A.}, \bibinfo{author}{Bemporad, A.} \&
  \bibinfo{author}{Morari, M.}
\newblock \bibinfo{journal}{\bibinfo{title}{On-line optimization via off-line
  parametric optimization tools}}.
\newblock {\emph{\JournalTitle{Computers \& Chemical Engineering}}}
  \textbf{\bibinfo{volume}{26}}, \bibinfo{pages}{175--185}
  (\bibinfo{year}{2002}).

\bibitem{bemporad2000piecewise}
\bibinfo{author}{Bemporad, A.}, \bibinfo{author}{Borrelli, F.} \&
  \bibinfo{author}{Morari, M.}
\newblock \bibinfo{title}{Piecewise linear optimal controllers for hybrid
  systems}.
\newblock In \emph{\bibinfo{booktitle}{Proceedings of the American Control
  Conference}}, vol.~\bibinfo{volume}{2}, \bibinfo{pages}{1190--1194}
  (\bibinfo{year}{2000}).
\newblock \bibinfo{note}{\textbf{The characteristic of PWL in control systems
  and the applications of PWL nonlinearity are introduced.}}

\bibitem{bemporad2002model}
\bibinfo{author}{Bemporad, A.}, \bibinfo{author}{Borrelli, F.} \&
  \bibinfo{author}{Morari, M.}
\newblock \bibinfo{journal}{\bibinfo{title}{Model predictive control based on
  linear programming - the explicit solution}}.
\newblock {\emph{\JournalTitle{IEEE Transactions on Automatic Control}}}
  \textbf{\bibinfo{volume}{47}}, \bibinfo{pages}{1974--1985}
  (\bibinfo{year}{2002}).

\bibitem{bemporad2002explicit}
\bibinfo{author}{Bemporad, A.}, \bibinfo{author}{Morari, M.},
  \bibinfo{author}{Dua, V.} \& \bibinfo{author}{Pistikopoulos, E.~N.}
\newblock \bibinfo{journal}{\bibinfo{title}{The explicit linear quadratic
  regulator for constrained systems}}.
\newblock {\emph{\JournalTitle{Automatica}}} \textbf{\bibinfo{volume}{38}},
  \bibinfo{pages}{3--20} (\bibinfo{year}{2002}).

\bibitem{chikkula1998dynamically}
\bibinfo{author}{Chikkula, Y.}, \bibinfo{author}{Lee, J.} \&
  \bibinfo{author}{Okunnaike, B.}
\newblock \bibinfo{journal}{\bibinfo{title}{Dynamically scheduled model
  predictive control using hinging hyperplane models}}.
\newblock {\emph{\JournalTitle{{AIC}h{E} Journal}}}
  \textbf{\bibinfo{volume}{44}}, \bibinfo{pages}{2658--2674}
  (\bibinfo{year}{1998}).

\bibitem{2009Analytical}
\bibinfo{author}{Wen, C.}, \bibinfo{author}{Ma, X.} \& \bibinfo{author}{Ydstie,
  B.~E.}
\newblock \bibinfo{journal}{\bibinfo{title}{Analytical expression of explicit
  mpc solution via lattice piecewise-affine function}}.
\newblock {\emph{\JournalTitle{Automatica}}} \textbf{\bibinfo{volume}{45}},
  \bibinfo{pages}{910--917} (\bibinfo{year}{2009}).

\bibitem{DBLP:conf/cdc/XuW19}
\bibinfo{author}{Xu, J.} \& \bibinfo{author}{Wang, S.}
\newblock \bibinfo{title}{Lattice piecewise affine representations on convex
  projection regions}.
\newblock In \emph{\bibinfo{booktitle}{Proceedings of the {IEEE} Conference on
  Decision and Control}}, \bibinfo{pages}{7240--7245} (\bibinfo{year}{2019}).

\bibitem{yue2015beyond}
\bibinfo{author}{Yue-Hei~Ng, J.} \emph{et~al.}
\newblock \bibinfo{title}{Beyond short snippets: Deep networks for video
  classification}.
\newblock In \emph{\bibinfo{booktitle}{Proceedings of the IEEE Conference on
  Computer Vision and Pattern Recognition}}, \bibinfo{pages}{4694--4702}
  (\bibinfo{year}{2015}).

\bibitem{purwins2019deep}
\bibinfo{author}{Purwins, H.} \emph{et~al.}
\newblock \bibinfo{journal}{\bibinfo{title}{Deep learning for audio signal
  processing}}.
\newblock {\emph{\JournalTitle{IEEE Journal of Selected Topics in Signal
  Processing}}} \textbf{\bibinfo{volume}{13}}, \bibinfo{pages}{206--219}
  (\bibinfo{year}{2019}).

\bibitem{xie2020self}
\bibinfo{author}{Xie, Q.}, \bibinfo{author}{Luong, M.-T.},
  \bibinfo{author}{Hovy, E.} \& \bibinfo{author}{Le, Q.~V.}
\newblock \bibinfo{title}{Self-training with noisy student improves imagenet
  classification}.
\newblock In \emph{\bibinfo{booktitle}{Proceedings of the IEEE/CVF Conference
  on Computer Vision and Pattern Recognition}}, \bibinfo{pages}{10687--10698}
  (\bibinfo{year}{2020}).

\bibitem{qiao2017fpga}
\bibinfo{author}{Qiao, Y.} \emph{et~al.}
\newblock \bibinfo{journal}{\bibinfo{title}{Fpga-accelerated deep convolutional
  neural networks for high throughput and energy efficiency}}.
\newblock {\emph{\JournalTitle{Concurrency and Computation: Practice and
  Experience}}} \textbf{\bibinfo{volume}{29}}, \bibinfo{pages}{e3850}
  (\bibinfo{year}{2017}).

\bibitem{Dua:2019}
\bibinfo{author}{Dua, D.} \& \bibinfo{author}{Graff, C.}
\newblock \bibinfo{title}{{UCI} machine learning repository}.
\newblock \bibinfo{howpublished}{\url{http://archive.ics.uci.edu/ml}}
  (\bibinfo{year}{2017}).

\bibitem{lecun-mnisthandwrittendigit-2010}
\bibinfo{author}{LeCun, Y.}, \bibinfo{author}{Bottou, L.},
  \bibinfo{author}{Bengio, Y.}, \bibinfo{author}{Haffner, P.} \emph{et~al.}
\newblock \bibinfo{journal}{\bibinfo{title}{Gradient-based learning applied to
  document recognition}}.
\newblock {\emph{\JournalTitle{Proceedings of the IEEE}}}
  \textbf{\bibinfo{volume}{86}}, \bibinfo{pages}{2278--2324}
  (\bibinfo{year}{1998}).
\newblock \bibinfo{note}{\textbf{The basic learning framework for generic DNNs
  including PWL-DNNs is formally introduced in this work.}}

\bibitem{netzer2011reading}
\bibinfo{author}{Netzer, Y.} \emph{et~al.}
\newblock \bibinfo{title}{Reading digits in natural images with unsupervised
  feature learning}.
\newblock In \emph{\bibinfo{booktitle}{NIPS Workshop on Deep Learning and
  Unsupervised Feature Learning 2011}} (\bibinfo{year}{2011}).

\bibitem{lecun2004learning}
\bibinfo{author}{LeCun, Y.}, \bibinfo{author}{Huang, F.~J.} \&
  \bibinfo{author}{Bottou, L.}
\newblock \bibinfo{title}{Learning methods for generic object recognition with
  invariance to pose and lighting}.
\newblock In \emph{\bibinfo{booktitle}{Proceedings of the IEEE Computer Society
  Conference on Computer Vision and Pattern Recognition}},
  vol.~\bibinfo{volume}{2}, \bibinfo{pages}{II--104} (\bibinfo{year}{2004}).

\bibitem{cifar10}
\bibinfo{author}{Krizhevsky, A.} \& \bibinfo{author}{Hinton, G.}
\newblock \emph{\bibinfo{title}{Learning multiple layers of features from tiny
  images}} (\bibinfo{publisher}{Technical report, University of Toronto},
  \bibinfo{year}{2009}).

\bibitem{lin2014microsoft}
\bibinfo{author}{Lin, T.-Y.} \emph{et~al.}
\newblock \bibinfo{title}{Microsoft {COCO}: Common objects in context}.
\newblock In \emph{\bibinfo{booktitle}{Proceedings of the European Conference
  on Computer Vision}}, \bibinfo{pages}{740--755} (\bibinfo{year}{2014}).

\bibitem{ILSVRC15}
\bibinfo{author}{Russakovsky, O.} \emph{et~al.}
\newblock \bibinfo{journal}{\bibinfo{title}{{ImageNet Large Scale Visual
  Recognition Challenge}}}.
\newblock {\emph{\JournalTitle{International Journal of Computer Vision}}}
  \textbf{\bibinfo{volume}{115}}, \bibinfo{pages}{211--252}
  (\bibinfo{year}{2015}).

\bibitem{krishnavisualgenome}
\bibinfo{author}{Krishna, R.} \emph{et~al.}
\newblock \bibinfo{journal}{\bibinfo{title}{Visual genome: Connecting language
  and vision using crowdsourced dense image annotations}}.
\newblock {\emph{\JournalTitle{International Journal of Computer Vision}}}
  \textbf{\bibinfo{volume}{123}}, \bibinfo{pages}{32--73}
  (\bibinfo{year}{2017}).

\bibitem{tensorflow2015-whitepaper}
\bibinfo{author}{Abadi, M.} \emph{et~al.}
\newblock \bibinfo{title}{{TensorFlow}: Large-scale machine learning on
  heterogeneous systems}.
\newblock \bibinfo{howpublished}{\url{https://www.tensorflow.org/}}
  (\bibinfo{year}{2015}).

\bibitem{chollet2015keras}
\bibinfo{author}{Chollet, F.}
\newblock \bibinfo{title}{Keras}.
\newblock \bibinfo{howpublished}{\url{https://github.com/fchollet/keras}}
  (\bibinfo{year}{2015}).

\bibitem{jia2014caffe}
\bibinfo{author}{Jia, Y.} \emph{et~al.}
\newblock \bibinfo{title}{Caffe: Convolutional architecture for fast feature
  embedding}.
\newblock In \emph{\bibinfo{booktitle}{Proceedings of the ACM international
  conference on Multimedia}}, \bibinfo{pages}{675--678} (\bibinfo{year}{2014}).

\bibitem{chen2015mxnet}
\bibinfo{author}{Chen, T.} \emph{et~al.}
\newblock \bibinfo{title}{{MXN}et: A flexible and efficient machine learning
  library for heterogeneous distributed systems}.
\newblock \bibinfo{howpublished}{Preprint at
  \url{https://arxiv.org/abs/1512.01274}} (\bibinfo{year}{2015}).

\bibitem{bergstra+al:2010-scipy}
\bibinfo{author}{Bergstra, J.} \emph{et~al.}
\newblock \bibinfo{title}{Theano: a {CPU} and {GPU} math expression compiler}.
\newblock In \emph{\bibinfo{booktitle}{Proceedings of the Python for Scientific
  Computing Conference}} (\bibinfo{year}{2010}).

\bibitem{tao2021toward}
\bibinfo{author}{Tao, Q.} \emph{et~al.}
\newblock \bibinfo{journal}{\bibinfo{title}{Toward deep adaptive hinging
  hyperplanes}}.
\newblock {\emph{\JournalTitle{IEEE Transactions on Neural Networks and
  Learning Systems}}}  (\bibinfo{year}{2021}).

\bibitem{tang2021sparse}
\bibinfo{author}{Tang, C.} \emph{et~al.}
\newblock \bibinfo{title}{Sparse {MLP} for image recognition: Is self-attention
  really necessary?}
\newblock \bibinfo{howpublished}{Preprint at
  \url{https://arxiv.org/abs/2109.05422}} (\bibinfo{year}{2021}).

\bibitem{2019Multilevel}
\bibinfo{author}{Wang, Y.}, \bibinfo{author}{Li, Z.}, \bibinfo{author}{Xu, J.}
  \& \bibinfo{author}{Li, J.}
\newblock \bibinfo{title}{Multilevel lattice piecewise linear representation
  and its application in explicit predictive control}.
\newblock In \emph{\bibinfo{booktitle}{Proceedings of the Asian Control
  Conference}}, \bibinfo{pages}{1066--1071} (\bibinfo{year}{2019}).

\bibitem{kawaguchi2016deep}
\bibinfo{author}{Kawaguchi, K.}
\newblock \bibinfo{title}{Deep learning without poor local minima}.
\newblock In \emph{\bibinfo{booktitle}{Advances in Neural Information
  Processing Systems}}, \bibinfo{pages}{586--594} (\bibinfo{year}{2016}).

\bibitem{yun2017global}
\bibinfo{author}{Yun, C.}, \bibinfo{author}{Sra, S.} \&
  \bibinfo{author}{Jadbabaie, A.}
\newblock \bibinfo{title}{Global optimality conditions for deep neural
  networks}.
\newblock \bibinfo{howpublished}{Preprint at
  \url{https://arxiv.org/abs/1707.02444}} (\bibinfo{year}{2017}).

\bibitem{Nguyen2017}
\bibinfo{author}{Nguyen, Q.} \& \bibinfo{author}{Hein, M.}
\newblock \bibinfo{title}{The loss surface of deep and wide neural networks}.
\newblock In \emph{\bibinfo{booktitle}{Proceedings of the International
  Conference on Machine Learning}}, vol.~\bibinfo{volume}{70},
  \bibinfo{pages}{2603--2612} (\bibinfo{year}{2017}).

\bibitem{yun2018small}
\bibinfo{author}{Yun, C.}, \bibinfo{author}{Sra, S.} \&
  \bibinfo{author}{Jadbabaie, A.}
\newblock \bibinfo{title}{Small nonlinearities in activation functions create
  bad local minima in neural networks}.
\newblock In \emph{\bibinfo{booktitle}{Proceedings of the International
  Conference on Learning Representations}} (\bibinfo{year}{2019}).

\end{thebibliography}


\end{document}